%% file: conflict_generalisation.tex
\documentclass{new_tlp}
\usepackage{mathptmx}

\usepackage{csquotes}
\usepackage{listings}
\usepackage{amsmath}
\usepackage{amssymb}
\usepackage{bm}
\usepackage{url}
\usepackage{algorithm}
\usepackage{algpseudocode}
\usepackage{cancel}
\usepackage[inline]{enumitem}

\usepackage[capitalise]{cleveref}

\input{macros}

  \title[Conflict Generalisation in ASP]
        {Conflict Generalisation in ASP: Learning Correct and Effective Non-Ground Constraints}

  \author[R.\ Taupe, A.\ Weinzierl, G.\ Friedrich]
{
	RICHARD TAUPE\textsuperscript{1,2}, ANTONIUS WEINZIERL\textsuperscript{3}, and GERHARD FRIEDRICH\textsuperscript{2}\\
	\textsuperscript{1} Siemens AG Österreich,\\
	\email{richard.taupe@siemens.com}\\
	\textsuperscript{2} Alpen-Adria-Universität, Klagenfurt, Austria,\\
	\email{gerhard.friedrich@aau.at}\\
	\textsuperscript{3} TU Wien (Vienna University of Technology), Austria,\\
	\email{antonius.weinzierl@kr.tuwien.ac.at}
}

\jdate{TBD}
\pubyear{TBD}
\pagerange{\pageref{firstpage}--\pageref{lastpage}}
\doi{TBD}

\submitted{TBD}
\revised{TBD}
\accepted{TBD}

\begin{document}
% PAGE LIMIT FOR ICLP 2020: 14 pages including bibliography

% globally reduce space before aligns:
\setlength{\abovedisplayskip}{0.5em}

\label{firstpage}

\maketitle

  \begin{abstract}
  	Generalising and re-using knowledge learned while solving one problem instance has been neglected by state-of-the-art answer set solvers.
	We suggest a new approach that generalises learned nogoods for re-use to speed-up the solving of future problem instances.
	Our solution combines well-known ASP solving techniques with deductive logic-based machine learning.
	Solving performance can be improved by adding learned non-ground constraints to the original program.
	We demonstrate the effects of our method by means
	of realistic examples, showing that our approach requires low computational cost to learn constraints that yield
	significant performance benefits in our test cases.
	These benefits can be seen with ground-and-solve systems as well as lazy-grounding systems.
	However, ground-and-solve systems suffer from additional grounding overheads, induced by the additional constraints in some cases.
	By means of conflict minimization, non-minimal learned constraints can be reduced.
	This can result in significant reductions of grounding and solving efforts, as our experiments show.
	(Under consideration for acceptance in TPLP.)
  \end{abstract}


\section{Introduction}
\label{sec:intro}

Conflict-Driven Nogood Learning (CDNL) is a major success factor for high-performance state-of-the-art ASP systems \cite{cdnl}.
When a conflict occurs, new propositional nogoods are learned that prevent the same conflict from re-occurring, which improves search performance.

Previous work has failed to address the question whether learned nogoods can be generalised and re-used to speed up the solving of different instances of the same problem. 
This paper aims to fill this gap.
We present an extension of CDNL that learns non-ground constraints.
The idea is that whole parts of the search tree can be pruned when these learned constraints are added to the original program.
We presume the common distinction between an unvarying problem encoding and separate inputs consisting only of facts.
A problem instance is specified by the problem encoding and a set of input facts.
We aim to derive non-ground constraints from the problem encoding that are valid for all possible inputs and which can be employed to speed up solving new instances.
In practice, instances are often not random but share some similarities that are reflected by nogoods learned during solving.
This led us to assume the existence of non-ground nogoods capable of significantly speeding up solving of practical problem instances.

CDNL and Explanation-Based Learning (EBL) \cite{DBLP:journals/ml/MitchellKK86,DBLP:journals/ml/DejongM86,DBLP:conf/ijcai/Hirsh87} are our starting point.
EBL is a well-known logic-based machine learning technique which learns first-order rules that are entailed by the background knowledge (in our case, the problem encoding).
We combine CDNL with EBL to learn non-ground nogoods while solving prior problem instances. 
Since the number of generalised nogoods can be overwhelming, choosing those that will actually pay off is particularly challenging.
Our basic idea is to generalise those non-ground conflicts that occur most often, i.e., we generalise propositional nogoods learned from frequently violated nogoods.
The underlying assumption is that nogoods learned from frequent conflicts will also be able to prevent many conflicts.
%In particular, our basic idea is to detect nogoods which were frequently exploited for solving instances of a specific problem. (???? What are frequently used nogoods, distinction between propositional and non-ground nogoods????) 

A realistic hardware configuration example and a graph colouring problem are used for demonstration and experimentation purposes.
Results show that both lazy-grounding and ground-and-solve systems benefit from our approach.
By adding learned constraints to the problem encodings, up to 64\% more configuration instances can be solved and graph colouring instances can be solved much faster.
Learning itself requires low computational resources.

We believe that we have developed an innovative tool supporting the design of efficient answer-set programs.
This view is supported by encouraging experimental results.
Our main contributions can be summarized as follows:
\begin{itemize}[label=--,leftmargin=3mm]
	\item We present a motivating practical example and describe how some redundant constraints can be derived from a given encoding.
			We also sketch how non-minimal nogoods can be reduced.
	\item Next, we introduce a novel approach combining CDNL and EBL to automatically learn correct non-ground nogoods while solving an answer-set program.
			A pseudo-code algorithm is given.
	\item We suggest techniques to make learned nogoods use only predicates from the input program, the means to choose from the wide range of possible generalisations, and methods to quickly conduct learning under limited resources.
	\item Finally, we experimentally compare the effects of various learned constraints on two domains.
			Results demonstrate the practical benefits of our approach.
\end{itemize}

\section{Preliminaries}
\label{sec:preliminaries}

\subsection{Answer Set Programming}
\label{sec:preliminaries:asp}

An answer-set program \prog\ is a finite set of non-ground rules of the form
%\begin{equation}
%\label{eq:rule}
%h \leftarrow b_1,~\ldots~,b_m,~\naf{b_{m+1}},~\ldots,~\naf{b_n}, c_1,~\ldots~,c_o,~\naf{c_{o+1}},~\ldots,~\naf{c_p}.
%\end{equation}
%where $h$ and $b_1,\dots,b_n$ are classical atoms and $c_1,\dots,c_p$ are fixed-interpretation atoms, such as external atoms or atoms using built-in comparison predicates like $<$, $>$, or $=$.
\begin{equation}
\label{eq:rule}
h \leftarrow b_1,~\ldots,~b_m,~\naf{b_{m+1}},~\ldots,~\naf{b_n}.
\end{equation}
where $h$ and $b_1,\dots,b_n$ are atoms.
%An atom is either a classical atom, a cardinality atom, or an aggregate atom.
A classical atom is an expression $p(t_1,\dots,t_n)$ where $p$ is an $n$-ary predicate and $t_1,\dots,t_n$ are terms.
A term is a variable, a constant, or a complex term involving arithmetics.
A NaF-literal is either an atom $a$ or its negation $\naf{a}$.
Negation as failure (NaF) refers to the absence of information, i.e., an atom is assumed to be false as long as it is not proven to be true.
If $l$ is a literal, then $\overline{l}$ denotes its complement (i.e., $\overline{\naf{\atom}} = \atom$ and $\overline{\atom} = \naf{\atom}$).

Given a rule $r$, $\head(r)=\{h\}$ is called the \textit{head} of $r$, and
$\body(r) = \{ b_1,~\ldots,~b_m,~\naf{b_{m+1}},~\ldots$, $\naf{b_n} \}$ is called the \textit{body} of $r$.
%The positive body of $r$ is denoted by $\bodyp(r) = \{ b_1,~\ldots~,b_m \}$ and its negative body by $\bodyn(r) = \{ b_{m+1},\dots,b_n\}$.
A rule $r$ where the head is absent ($\head(r) = \{\}$), e.g., $\leftarrow b.$, is called \textit{constraint}.
A rule $r$ where the body is absent ($\body(r) = \{\}$), e.g., $h \leftarrow.$, is called \textit{fact}.
Given an answer-set program $\prog$, the \textit{universe} of $\prog$, denoted by $\universe_\prog$, is the set of constants appearing in $\prog$.
%The \textit{base} of $\prog$ is the set of classical atoms constructible from predicates of $\prog$ with constants in $\universe_\prog$.
%It is denoted by $\base_\prog$ or, if $\prog$ need not be specified, simply by $\atoms$ (set of all atoms).
By $\atoms$ we denote the set of classical atoms constructible from predicates of $\prog$ with constants in $\universe_\prog$.
The set of all literals is denoted by $\literals = \atoms \cup \{ \naf{a} \mid a \in \atoms \}$.
The $\facts$ function maps a program $\prog$ to all ground atoms defined by facts in $\prog$, i.e., $\facts(\prog) = \{ \head(r) \mid r \in \prog \text{ and } \body(r) = \emptyset \}$.

The $\vars$ function maps any structure containing variables to the set of variables it contains, e.g., %$\vars(\mathrm{a}(X)) = \{ X \}$ and
for a rule $r_1\colon \mathrm{a}(X) \leftarrow \mathrm{b}(X,Y).$, $\vars(r_1) = \{ X, Y \}$.
The set $\allvars \supset \vars(\prog)$ includes all variables from a program and also variables that can additionally be used by a solver, it is usually infinite.
A literal $l$ or rule $r$ is ground if $\vars(l) = \emptyset$ or $\vars(r) = \emptyset$, respectively.
A program $\prog$ is ground if all its rules $r \in \prog$ are.
A \textit{substitution} $\substitution\colon \allvars \rightarrow \allvars \cup \universe_\prog$ is a mapping from variables to variables or constants.
A substitution is called \textit{grounding} for a set of variables $V \subseteq \allvars$ if for every $v \in V$ there exists a constant $c \in \universe_\prog$ s.t.\ $\varsigma(v) = c$.
The function $\assignedtruth_\assignment\colon \atoms \to \{ \sigT, \sigM, \sigF, \sigU \}$ for a (partial) assignment $\assignment$ maps an atom to the truth value the atom is currently assigned in the given assignment ($\sigM$ stands for must-be-true and $\sigU$ for currently unassigned).

A nogood is a set of Boolean signed literals $\{ \sigT\; \atom_1, \dots, \sigT\; \atom_n, \dots, \sigF\; \atom_{n+1}, \dots, \sigF\; \atom_m \}$ which means that an answer set may not satisfy all the literals in the nogood.
A literal in a nogood differs from a NaF-literal in a rule semantically:
In a nogood, negation as failure has no role.
A nogood is violated by an assignment $\assignment$ if $\assignedtruth_\assignment(\atom_i) \in \{ \sigT, \sigM \}$ for all $i \in \{1, \dots, n \}$ and $\assignedtruth_\assignment(\atom_j) = \sigF$ for all $j \in \{ n + 1, \dots, m \}$.
Sometimes, we will represent nogoods not by sets but by fixed-order multisets, because we need the possibility of duplicates and of fixed positions to be able to map literals between ground and non-ground nogoods.
When this is the case, $\omega[i]$ denotes the literal at position $i$ in nogood $\omega$.

%An interpretation $I \subseteq \atoms$
%satisfies a ground rule $r$, denoted $I \models r$, if
%$\bodyp(r) \subseteq I \land \bodyn(r) \cap I = \emptyset$ implies
%$\head(r) \subseteq I$ and $\head(r) \neq \emptyset$.
%$I$ is an \emph{answer set} of a ground program $\prog$ if $I$ is the subset-minimal model of $\prog^I$, where
%$\prog^I = \{ r \in \prog \mid \bodyp(r) \subseteq I \land \bodyn(r) \cap I = \emptyset \}$
%is the so-called FLP reduct, the set of rules whose body is satisfied by $I$ \cite{DBLP:journals/ai/FaberPL11}.
For ASP semantics, we refer to \citeN{DBLP:journals/ai/FaberPL11}.

Aggregates and choice rules are common extensions to the input language of ASP  \cite{aspcore2}.
Aggregate atoms are used to express arithmetic constraints on sets of atoms (e.g., cardinality constraints).
%Choice rules are another common extension to the input language of ASP that are supported by many solvers.
A choice rule expresses that a subset of a set of atoms will be included in an answer set.
This does not clash with the subset-minimality of answer sets because choice rules are translated to normal rules involving additional atoms whose predicates we call \textit{choice predicates}.
For simplicity, we only consider choice rules without bounds (which are the only ones currently supported by \alphaslv).
A choice rule is of the form \labelcref{eq:rule}, but the head $h$ is not a classical atom but a choice atom of the form $\{ C_1; \dots; C_n \}$, where for $n > 0$ and $0 \leq i \leq n$ each $C_i$ is a \textit{choice element}.
A choice element has the form $a \colon l_1, \dots, l_k$, where $a$ is a classical atom and $l_1, \dots, l_k$ are literals for $k \geq 0$.
If the body of a choice rule is satisfied, it means that for each $C_i$ where the literals $l_1, \dots, l_k$ are satisfied, the atom $a$ may or may not be true in an answer set.
A choice rule $r$ can be translated to $2n$ rules, two for each choice element $a \colon l_1, \dots, l_k$:
%\begin{align*}
%	a		&\leftarrow \body(r), l_1, \dots, l_k, \naf{\widehat{a}}.\\
%	\widehat{a}	&\leftarrow \body(r), l_1, \dots, l_k, \naf{a}.
%\end{align*}
\begingroup
\setlength{\abovedisplayskip}{0.1em}
\setlength{\belowdisplayskip}{0.1em}
\[
a		\leftarrow \body(r), l_1, \dots, l_k, \naf{\widehat{a}}. \hspace{3em} \widehat{a}	\leftarrow \body(r), l_1, \dots, l_k, \naf{a}.
\]
\endgroup
For both aggregates and choice rules, see \citeN{aspcore2} for details.

%\subsection{Theoretical Basis}

The goal of our method is to generate non-ground nogoods which can be re-used for solving further problem instances.
More formally, let $\encoding$ be an ASP problem encoding. $\InPred$ and $\OutPred$ are sets of predicate symbols with defined arities. The input for $\encoding$ is specified by a set of ground facts $\InA$ containing only predicate symbols from $\InPred$.  The output of a problem instance $\encoding \cup \InA$  is the set of atoms of a stable model whose predicate symbols are in $\OutPred$. 

A conflict for an ASP program $P$ is a constraint $C$ s.t.\ $P \modelsSkeptically C$ where $\modelsSkeptically$ refers to the skeptical ASP semantic, i.e., $C$ must be true in all stable models of $P$. 
\begin{definition} A generalized conflict of $(\encoding, \InPred)$ is a constraint $\GC$ s.t. $\encoding \cup \InA \modelsSkeptically \GC$ for all finite sets of ground facts $\InA$ whose predicate symbols are in  $\InPred$.
\end{definition}

Let  $\modelsFOL$ be the implication relation based on standard FOL semantic. Since every stable model of an ASP program $P$ is also a model for $P$ if $P$ is interpreted as an FOL program, it follows that if $P \modelsFOL C$ then $P \modelsSkeptically C$. 
It is well known that every stable-model of $P$ is also a model of $P \cup \comp (P)$, i.e., the extension of $P$ by the axioms of the Clark completion. Note that for problem instances  $\encoding \cup \InA$ the Clark completion changes depending on $\InA$. Since we are interested in conflicts which are independent of the input, we consider only the completion of the head literals which appear in $\encoding$ but not in $\InPred$. We denote this restricted Clark completion for a program $P$ and input predicates $\InPred$ as $\compR (P, \InPred)$. It holds that  $\compR (P, \InPred) \subseteq \comp (P) $ and $\encoding \cup \compR (\encoding \cup \InA, \InPred )$ is the same for all problem instances $\encoding \cup \InA$.
\begin{corollary} Let a problem be defined by $(\encoding, \InPred)$ and $C$ is a conflict. If $\encoding \cup \compR (P, \InPred) \modelsFOL C$ then $C$ is a generalized conflict of $(\encoding, \InPred)$. 
\end{corollary} 

\paragraph{Proof sketch:} Since $\compR (P, \InPred)$ depends only on $P$, it holds that if  $\encoding \cup \compR (P, \InPred)$ $\modelsFOL C$ then $\encoding \cup \compR (P, \InPred) \cup \InA \modelsFOL C$ where $\InA$ is a finite set of ground facts whose predicate symbols are in  $\InPred$. Because every stable model of $\encoding \cup \InA$ is also a model for $\encoding \cup \compR (P, \InPred) \cup \InA$ under FOL semantic and $C$ is true in these models it follows that $\encoding \cup \InA \modelsSkeptically C$. \quad\proofbox

Consequently, adding a generalized conflict $\GC$ to $\encoding$ does not change the set of outputs of $(\encoding, \InPred)$ for any input. 

\subsection{Conflict-Driven Nogood Learning (CDNL)}
\label{sec:preliminaries:cdnl}

Conflict-Driven Clause Learning (CDCL) \cite{cdcl,cdcl2} is a SAT solving technique that extends DPLL by conflict analysis and enables the solver to learn new clauses and to do non-chronological backtracking (\enquote{backjumping}).
While the original definitions of CDCL are based on clauses, i.e., disjunctions of literals, we use the notion of nogoods %A nogood is a set of literals that cannot be true all together.
and speak of Conflict-Driven Nogood Learning (CDNL) \cite{cdnl}.
The two variants are equivalent since the conjunction of classically negated nogoods can be directly transformed to an equivalent conjunction of clauses by applying De Morgan's laws.

CDCL- and CDNL-based solvers % suspended hyphens are used also in English: https://en.wikipedia.org/wiki/Hyphen#Suspended_hyphens
usually characterize each atom $\atom \in \atoms$ by the following properties:
its \emph{truth value}, its \emph{antecedent clause}, and its \textit{decision level}, denoted respectively by $\assignedtruth(\atom) \in \{ \sigT, \sigF, \sigM, \sigU \}$, %\footnote{Some solvers do not use the truth value $\sigM$ (must-be-true).}
$\antecedent(\atom) \in 2^\literals \cup \{ \nil \}$, and $\dl(\atom) \in \{ -1, 0, 1, \dots, |\atoms| \}$.
The truth value of an atom is the value currently assigned to it.
By the \emph{antecedent clause} of an atom we mean the cause of its current truth value.
The antecedent clause of an atom whose current truth value has been implied is the nogood that was unit at the time the implication happened and thus forced the atom to assume its current value.
The value $\nil$ is used instead if the current truth value of an atom results from a heuristic decision.
The decision level of an atom denotes the depth of the decision tree at which it has been assigned, and $-1$ if the atom is unassigned.
%The notation $\atom = v@d$ is used to denote that $\assignedtruth(\atom) = v$ and $\dl(\atom) = d$.
The decision level of a literal is the same as the literal's atom's decision level ($\dl(\naf{\atom}) = \dl(\atom)$),
and the antecedent of a literal is the one of its atom ($\antecedent(\naf{\atom}) = \antecedent(\atom)$).
%The decision level of an assignment $\dl(\assignment)$ is the decision level of the atom most recently assigned in $\assignment$ \cite{cdcl,cdcl2}.
An assignment's decision level is the decision level of the atom most recently assigned \cite{cdcl,cdcl2}.

When a nogood is violated during solving (i.e., all literals in the nogood are satisfied), the conflict is analysed and a new nogood may be learned.
Learning starts with the violated nogood and resolves one literal assigned at the current (most recent) decision level with its antecedent.
This is repeated with the nogood resulting from resolution (the \enquote{resolvent}) as long as it is possible. Resolution stops when the only literal in the resolvent that has been assigned at the current decision level is a decision literal, i.e., its antecedent is \nil.
The resulting nogood is then \enquote{learned}, i.e., added to the set of known nogoods, and the solver executes a backjump \cite{cdcl}.

\emph{First UIP clause learning} is a modification of this algorithm that already stops resolution when only one literal from the current decision level remains in the nogood, even if its antecedent is not \nil.
This is correct because this literal is a unique implication point (UIP), a node in the implication graph that lies on every path from the last decision to the conflict.
Each UIP could be used to learn a new nogood, but modern SAT solvers stop already at the first UIP, i.e., the one nearest to the conflict \cite{cdcl}.

\subsection{Explanation-Based Learning}
\label{sec:preliminaries:ebl}

Explanation-Based Learning (EBL), a.k.a.\ Explanation-Based Generalization (EBG), is a logic-based learning technique that, in contrast to Inductive Logic Programming (ILP), learns only general rules that are entailed by background knowledge alone (and not by background knowledge together with some new hypotheses).
Therefore, in EBL nothing factually new is learned from the example.
In EBL, two proof trees are created simultaneously (one for the concrete example and one with variables instead of constants).
Then, new rules can be obtained from the non-ground proof tree.
The \textit{operationality criterion} is a restriction which predicates can be used to express the learned rules.
The purpose of operationality is to use only predicates that are easy to solve.
A trade-off between operationality and generality is usually an issue in EBL.
By choosing adequate general rules, EBL makes a knowledge base more efficient for the kind of problems that one would reasonably anticipate
\cite{DBLP:journals/ml/MitchellKK86,DBLP:journals/ml/DejongM86,DBLP:conf/ijcai/Hirsh87,DBLP:books/lib/Mitchell97,DBLP:books/daglib/0023820}.

%\citeN[ch.\ 19]{DBLP:books/daglib/0023820} give a good overview of Explanation-Based Learning (EBL) and compare it to other logic-based learning techniques, such as inductive logic programming (ILP).
%%An example is provided, which consists of a logic program that simplifies arithmetic expressions.
%The authors emphasise the importance of finding a good balance between generality and operationality of learned rules: \enquote{Empirical analysis of efficiency is actually at the heart of EBL. [\dots] \emph{By generalizing from past example problems, EBL makes the knowledge base more efficient for the kind of problems that it is reasonable to expect.}}

\citeN{DBLP:journals/ai/HarmelenB88} show that EBL/EBG and Partial Evaluation (PE) are equivalent to each other in the context of logic programming.
PE is a program optimisation method that reformulates an input program in an equivalent but more efficient way.
%A classical example from the machine learning literature, \enquote{Stacking Example}, is presented.
%This seems to be unsuited as an example for conflict generalisation in ASP, though, since the goal concept, \texttt{safe\_to\_stack/2}, is stratified and can therefore be fully evaluated by a grounder.
%\rttodo{Look at original sources of EBG and PE in logic programming mentioned in this paper.}
%\rttodo{Look at the simultaneous and independent similar result mentioned in the final note.}
%
%\citeN{Friedrich.1990} use EBL / PE to make general diagnostic knowledge explicit that is already implicitly contained in a Horn knowledge base.
%They use the \enquote{D74 Circuit Example} to illustrate their approach.
\citeN{Weinzierl.2013} employs PE (and terms it \enquote{unfolding} of rules) in lazy-grounding answer set solving to learn new rules during solving.
Learning is triggered by conflicts to prevent future occurrences of similar conflicts.
Learned rules are constructed from constraints that are almost violated and in which atoms have been replaced by their definitions: 
The single literal in the constraint's body that is not yet satisfied must not be satisfied.
Learned rules are created and used online during solving.%, but \citeANP{Weinzierl.2013} also mentions that learned rules may be reused when solving future problem instances.

\section{A Motivating Example}
\label{sec:example}

The House Configuration Problem (HCP) serves as a motivating example.
It is an abstraction of real-world hardware configuration problems and defined as follows:\footnote{See \url{https://sites.google.com/view/aspcomp2019/problem-domains}, \citeN{DBLP:conf/confws/FriedrichRFHSS11}, and \citeN{Ryabokon:Dissertation} for more complete descriptions of the problem.}
Given an association of things to the persons they belong to, a domain of cabinet IDs, and a domain of room IDs, the task is to assign things to cabinets and cabinets to rooms such that there are at most five things in each cabinet and at most four cabinets in each room, and each room only contains things that belong to the same person.
The problem encoding stays the same for all problem instances.
Input facts of predicates $\mr{personTOthing}/2$, $\mr{cabinetDomain}/1$, and $\mr{roomDomain}/1$ vary from instance to instance. %These facts describe the input for generating a solution.
Each answer set specifies one valid configuration. 

We are guided by the original definition of HCP by \citeN{DBLP:conf/confws/FriedrichRFHSS11} and limit ourselves to the configuration  aspects of the problem, neglecting reconfiguration for presentation purposes. 
The complete encoding is given in \cref{lst:house}.\footnote{Compared to \citeN{Ryabokon:Dissertation}, variables have been renamed for this paper %, the $\neq$ operator in constraints has been replaced by $<$ to break symmetries, and some simplifications have been made (e.g., function terms have been eliminated without changing the semantics of the program).}
and some simplifications have been made.}

\begin{lstlisting}[float,label={lst:house},caption={An encoding for the House Configuration Problem}]
{ cabinet(C) } :- cabinetDomain(C).
{ room(R) } :- roomDomain(R).
room(R1) :- roomDomain(R1), roomDomain(R2), room(R2), R1 < R2.
cabinet(C1) :- cabinetDomain(C1), cabinetDomain(C2), cabinet(C2), C1 < C2.

{ cabinetTOthing(C,T) } :- cabinetDomain(C), thing(T).
thingHasCabinet(T) :- cabinetTOthing(C,T).
:- thing(T), not thingHasCabinet(T).
:- thing(T), cabinetDomain(C1), cabinetTOthing(C1,T), cabinetDomain(C2), cabinetTOthing(C2,T), C1 < C2.
:- 6 <= #count { T : cabinetTOthing(C,T), thing(T) }, cabinet(C).

{ roomTOcabinet(R,C) } :- roomDomain(R), cabinet(C).
cabinetHasRoom(C) :- roomTOcabinet(R,C).
:- cabinet(C), not cabinetHasRoom(C). %|\label{lst:house:constraint_cabinet_without_room}%|
:- cabinet(C), roomDomain(R1), roomTOcabinet(R1,C), roomDomain(R2), roomTOcabinet(R2,C), R1 < R2.
:- 5 <= #count { C : roomTOcabinet(R,C), cabinetDomain(C) }, room(R).

personTOroom(P,R) :- personTOthing(P,T), cabinetTOthing(C,T), roomTOcabinet(R,C). %|\label{lst:house:rule_personTOroom}%|
:- personTOroom(P1,R), personTOroom(P2,R), P1 < P2. %|\label{lst:house:most_violated_constraint}%|

room(R) :- roomTOcabinet(R,C).
room(R) :- personTOroom(P,R).
cabinet(C) :- cabinetTOthing(C,T). %|\label{lst:house:rule_cabinet}%|
cabinet(C) :- roomTOcabinet(R,C).
thing(T) :- personTOthing(P,T).
\end{lstlisting}

\citeN{Ryabokon:Dissertation} added the following redundant constraint to the problem:
\begin{align}
\label{eq:redundant_constraint}
\leftarrow\;
& \mr{cabinetTOthing}(C, \mi{T1}), \mr{personTOthing}(\mi{P1}, \mi{T1}), \nonumber \\
& \mr{cabinetTOthing}(C, \mi{T2}), \mr{personTOthing}(\mi{P2}, \mi{T2}), 
  \mi{P1} < \mi{P2}.
\end{align}

However, this constraint can be learned automatically. %starting from the nogood which is most frequently violated. 
Using \alphaslv\ \cite{DBLP:conf/lpnmr/Weinzierl17} to solve the smallest instance, the nogood that is violated most often and thus leads to the highest number of conflicts is $\{ \sigT~\mr{personTOroom}(\mi{P1}, R),$ $\sigT~\mr{personTOroom}(\mi{P2}, R),$ $\mi{P1} < \mi{P2} \}$ which corresponds to the constraint in \cref{lst:house:most_violated_constraint} of \cref{lst:house}. This constraint forbids things of two different persons $\mi{P1}$ and $\mi{P2}$ from ending up in the same room $R$.
Using our method, the following non-ground constraint can be learned at the first UIP of the implication graphs of all these conflicts:
\begin{align}
\label{eq:1uip_constraint}
\leftarrow\; \mr{roomTOcabinet}(R, C),
&\; \mr{cabinetTOthing}(C, \mi{T1}), \mr{personTOthing}(\mi{P1}, \mi{T1}), \nonumber \\
&\; \mr{cabinetTOthing}(C, \mi{T2}), \mr{personTOthing}(\mi{P2}, \mi{T2}),
  \mi{P1} < \mi{P2}.
\end{align}

This constraint can also be easily generated by applying the rule in \cref{lst:house:rule_personTOroom} to the constraint in \cref{lst:house:most_violated_constraint} of \cref{lst:house} and by factorizing different literals of predicate $\mr{roomTOcabinet}/2$. The automatically learned constraint \labelcref{eq:1uip_constraint} is almost identical to the human-created one \labelcref{eq:redundant_constraint}. However, it includes one additional literal $\mr{roomTOcabinet}(R, C)$.

Learned non-ground constraints can be minimized by employing axioms of the Clark completion \cite{DBLP:journals/jlp/Shepherdson84}. The Clark completion of a logic program is computed by completing the head literals of the clauses of the program, including the completion of facts. 
Every stable model of a program $P$ must also satisfy the Clark completion $\comp(P)$. For example, the completion of $\mr{cabinetHasRoom(\mi{C)}}$ is $ \exists R : \mr{roomTOcabinet(\mi{R,C})} \leftarrow \mr{cabinetHasRoom(\mi{C)}}$. To minimize the learned nogood by resolution we substitute the existential quantifier by a Skolem function resulting in the rule $  \mr{roomTOcabinet(\mi{r(C),C})} \leftarrow \mr{cabinetHasRoom(\mi{C)}}$.

To minimize the conflict, we have to prove that constraint \labelcref{eq:redundant_constraint} is entailed.
This is achieved by showing that the program in \cref{lst:house} becomes unsatisfiable when the negation of constraint \labelcref{eq:redundant_constraint} is added to this program. 
Consequently, it is sufficient to show that the program becomes unsatisfiable when the facts $ \mr{cabinetTOthing}(\mi{c}, \mi{t1}),$ $ \mr{personTOthing}(\mi{p1}, \mi{t1}), $ $\mr{cabinetTOthing}(\mi{c}, \mi{t2}), $ $\mr{personTOthing}(\mi{p2}, \mi{t2}), $ $\mi{p1} < \mi{p2}$ are added, where $\mi{c}$, $\mi{t1}$, $\mi{t2}$, $\mi{p1}$, and $\mi{p2}$ are Skolem constants.

Every stable model of an ASP program is also a model of the program interpreted under the standard first-order logic (FOL) semantic. Consequently, if a program is unsatisfiable under FOL semantics, it is also unsatisfiable skeptical ASP semantics.
From $\mr{cabinetTOthing}(\mi{c}, \mi{t1})$ and the rule in \cref{lst:house:rule_cabinet} we can deduce  $\mr{cabinet(\mi{c})}$.
From the constraint in \cref{lst:house:constraint_cabinet_without_room} of \cref{lst:house}, we can deduce $ \mr{cabinetHasRoom(\mi{c)}}$ interpreting the sentences in \cref{lst:house} as FOL clauses. 
Exploiting the rule provided by the Clark completion of $\mr{cabinetHasRoom(\mi{C)}}$ as shown above, we can deduce $ \mr{roomTOcabinet(\mi{r(c),c})}$.
Finally, using all the deduced and given facts, constraint \labelcref{eq:1uip_constraint} is violated.  Since this constraint is implied by the program and we have shown that it is violated if the negation of constraint \labelcref{eq:redundant_constraint} is added, constraint \labelcref{eq:redundant_constraint} is implied. As a result, we have shown that indeed the learned non-ground constraint can be reduced by dropping $\mr{roomTOcabinet}(R, C)$.

\section{Conflict Generalisation}
\label{sec:conflict-generalisation}

We propose to combine CDNL with EBL to facilitate the learning of general (i.e., non-ground) constraints.
This can be done online (during solving), or offline.
In the latter case, learned constraints are computed and recorded.
A selection can then be made, for example either automatically based on metrics or by a human, and useful constraints can be added to the original program to improve solving performance in the future.
In this section, we describe how offline conflict generalisation can be implemented in an ASP solver and exemplify this by means of a prototypical implementation in the lazy-grounding system \alphaslv\ \cite{DBLP:conf/lpnmr/Weinzierl17}.
The constraints learned by this method can be used by any ASP system, not only by the system employed for the learning task, because they are stated in pure ASP-Core-2 \cite{aspcore2}, using only predicates from the original program.

%\subsection{Ground and non-ground nogoods.}
%\label{sec:conflict-generalisation:nogoods}

Parallelly conducting CDNL on the ground level and on the non-ground level when a conflict is encountered
is the basic idea of combining CDNL with EBL to generalise learned nogoods.
A key requirement for this is that for each ground nogood, a corresponding non-ground nogood is known that can be used for non-ground resolution.
We will now describe how non-ground nogoods are maintained in \alphaslv. This can be implemented similarly in any other ASP system.

In \alphaslv, there are five types of nogoods:
\begin{itemize}
	\item \emph{static} nogoods that represent ground rules and are generated by the grounder;
	\item \emph{support} nogoods that encode the situation that the body-representing atom of a rule must be true if the head of the rule is true; %(nogoods of this kind are generated only in very special cases by \alphaslv);
	\item \emph{learned} nogoods originating from CDNL;
	\item \emph{justification} nogoods learned by justification analysis \cite{DBLP:conf/ijcai/BogaertsW18};
	\item and \emph{internal} nogoods containing solver-internal atoms.
\end{itemize}
Non-ground nogoods for \textit{static} and \textit{support} nogoods are produced by the grounder as described in the following paragraphs.
Non-ground nogoods for \textit{learned} nogoods are produced by the conflict generalisation procedure that is the main contribution of this paper.
Non-ground nogoods are not maintained for \textit{justification} nogoods (because these nogoods depend too heavily on the specific problem instance) and for \textit{internal} nogoods (because these are irrelevant to CDNL).

We now describe how non-ground nogoods for static and support nogoods are produced by the grounder.
Let $r$ be a rule of the form \labelcref{eq:rule} and $\substitution$ be a substitution that is grounding for $\vars(r)$.
The grounder produces a ground rule $r\substitution$ from $r$ if no fact and no fixed-interpretation literal makes the body of $r\substitution$ false.
Because of this, facts and fixed-interpretation literals are actually omitted from the ground rule produced by the grounder.\footnote{
A more sophisticated grounder would even use the set of atoms derived by a stratified component of the program depending only on facts instead of $\facts(\prog)$ to make generated rules more compact.}
For simplicity of presentation, however, we assume here that the grounder does not eliminate any atoms known to be true from generated nogoods.
This ensures a one-to-one relationship between literals in ground nogoods and literals in non-ground nogoods.
Additional effort is necessary to map from ground literals to non-ground literals if true literals are eliminated from ground nogoods, but this is purely a matter of implementation.
Not eliminating any literals from non-ground nogoods, however, is crucial.

A body-representing atom $\choiceatom{r}{\substitution}$ is created for every ground rule $r\substitution$.
Similarly, $\nongroundchoiceatom{r}$ is a fictitious atom representing the body of a non-ground rule $r$.
The latter atom contains a term that lists all the variables that occur in the rule $r$, so that they can affect unification when a body-representing literal is used for resolution.
This will become crucial in \cref{sec:conflict-generalisation:cdnl}.

The following static nogoods are produced by the grounder from a rule $r$ and a grounding substitution $\substitution$ \cite{DBLP:conf/inap/LeutgebW17}:
%\begin{align}
%	\label{eq:nogood-beta-to-full-body} &\{ \underline{\sigF~\choiceatom{r}{\substitution}}, \sigT~b_1 \substitution, \dots, \sigT~b_m \substitution, \sigF~b_{m+1} \substitution, \dots, \sigF~b_n \substitution \}\\
%	&\{ \underline{\sigF~h\substitution}, \sigT~\choiceatom{r}{\substitution}\}\\
%	&\{ \sigT~\choiceatom{r}{\substitution}, \sigF~b_1 \substitution \}, \dots, \{ \sigT~\choiceatom{r}{\substitution}, \sigF~b_m \substitution \}\\
%	&\{ \sigT~\choiceatom{r}{\substitution}, \sigT~b_{m+1} \substitution \}, \dots, \{ \sigT~\choiceatom{r}{\substitution}, \sigT~b_n \substitution \}
%\end{align}
% note: do not align heads of nogoods, because heads are not used in this paper and also not explained:
\begin{align}
	\label{eq:nogood-beta-to-full-body} &\{ \sigF~\choiceatom{r}{\substitution}, \sigT~b_1 \substitution, \dots, \sigT~b_m \substitution, \sigF~b_{m+1} \substitution, \dots, \sigF~b_n \substitution \}\\
	\label{eq:nogood-neg-head-to-beta} &\{ \sigF~h\substitution, \sigT~\choiceatom{r}{\substitution}\}\\
	\label{eq:nogood-beta-to-posbodylit} &\{ \sigT~\choiceatom{r}{\substitution}, \sigF~b_1 \substitution \}, \dots, \{ \sigT~\choiceatom{r}{\substitution}, \sigF~b_m \substitution \}\\
	\label{eq:nogood-beta-to-negbodylit} &\{ \sigT~\choiceatom{r}{\substitution}, \sigT~b_{m+1} \substitution \}, \dots, \{ \sigT~\choiceatom{r}{\substitution}, \sigT~b_n \substitution \}\\
	\label{eq:support-nogood} &\{ \sigT~h\substitution, \sigF~\choiceatom{r}{\substitution}\}
\end{align}
Nogood \labelcref{eq:support-nogood} is a so-called \textit{support nogood} and is produced by \alphaslv\ currently only when an atom occurs in the head of just a single rule.
If $r$ is a constraint, %(i.e., $\head(r) = \emptyset$),
only one nogood is created, which consists of the whole body of $r\substitution$.

As a prerequisite for conflict generalisation, a solver must associate each ground nogood with a non-ground nogood.
This non-ground nogood is obtained from a non-ground rule the same way as a ground nogood is obtained from a ground rule.
This means that the non-ground nogoods are exactly the same ones as given above, except the substitution $\substitution$ does not appear anywhere:
The atom $\choiceatom{r}{\substitution}$ becomes $\nongroundchoiceatom{r}$, and any other ground atom $a\substitution$ becomes just $a$.

\subsection{Non-ground CDNL}
\label{sec:conflict-generalisation:cdnl}

In this section, we describe how we extend CDNL to learn non-ground nogoods.
The non-ground nogoods learned that way can then just be used as constraints and be added to the original program.
When learning only constraints and no other kinds of rules, $\sigF$ in nogoods can just be replaced by negation as failure.

\begin{algorithm}[tp]
	\textbf{Input:} $\omega$: the violated antecedent, $\assignment$: current assignment\\
	\textbf{Output:} a list of learned ground nogoods and a list of learned non-ground nogoods, or UNSAT
	\begin{algorithmic}[1]
			\If{$\dl(\assignment) = 0$}
				\State \Return{UNSAT} \label{alg:conflict-generalisation:AnalyzeConflict:return-UNSAT}
			\EndIf
			\State $\mi{LearnedNoGoods} \leftarrow$ empty list
			\State $\mi{LearnedNonGroundNoGoods} \leftarrow$ empty list
			\State $\Omega \leftarrow$ non-ground nogood associated with $\omega$
			\While{$(l \leftarrow \Call{FindNextLiteralForResolution}{\omega, \assignment}) \neq \nil$} \label{alg:conflict-generalisation:AnalyzeConflict:while}
				\State $\omega,\Omega \leftarrow \Call{Resolve}{\omega,\Omega,l}$ \label{alg:conflict-generalisation:AnalyzeConflict:Resolve}
				\If{$\exists!\; l' \in \omega\colon \dl(\omega) = \dl(\assignment)$} \label{alg:conflict-generalisation:AnalyzeConflict:UIP}
					\State{append $\omega$ to $\mi{LearnedNoGoods}$}
					\State{append $\Omega$ to $\mi{LearnedNonGroundNoGoods}$}
				\EndIf
			\EndWhile
			\State \Return{$\mi{LearnedNoGoods},\mi{LearnedNonGroundNoGoods}$} \label{alg:conflict-generalisation:AnalyzeConflict:return-lists}
		
		\Statex
		\Procedure{FindNextLiteralForResolution}{$\omega,\assignment$}
			\State \Return{the literal $l$ most recently assigned in $\assignment$ that appears in $\omega$}
			\StatexIndent[3] and for which it holds that $\dl(l) = \dl(\assignment)$ and $\antecedent(l) \neq \nil$,
			\StatexIndent[3] or $\nil$ if no such literal exists
		\EndProcedure
		
		\Statex
		\Procedure{Resolve}{$\omega, \Omega, l$}
%			\textbf{Input:} $\omega$: a ground nogood, $\Omega$: a non-ground nogood, $l$: literal\\
%			\textbf{Output:} the ground resolvent and the non-ground resolvent
			\State $\omega' \leftarrow (\omega \setminus \{ l \}) \cup (\antecedent(l) \setminus \{\overline{l}\})$ \label{alg:resolution:ground-resolution}
			\State $L \leftarrow$ non-ground literal s.t.\ $\omega[i] = l$ and $\Omega[i] = L$ for some $i$
			\State $\bar{L} \leftarrow$ non-ground literal s.t.\ $\antecedent(l)[j] = \bar{l}$ and $\Omega[j] = \bar{L}$ for some $j$
			\State $\Omega' \leftarrow$ the non-ground nogood associated with $\antecedent(l)$, standardised apart from $\Omega$ \label{alg:resolution:identify-ngng}
			\State $\substitution \leftarrow \mathrm{unify}(L,\bar{L})$ \label{alg:resolution:unify-resolution-literal}
			\State $\substitution \leftarrow \substitution \circ \Call{UnifyDuplicateLiterals}{\Omega \substitution, \Omega' \substitution,\omega,\antecedent(l)}$ \label{alg:resolution:unify-duplicate-literals}
			\State $\Omega' \leftarrow (\Omega \setminus \{L \}) \substitution \cup (\Omega' \setminus \{\overline{L}\}) \substitution$ \label{alg:resolution:nonground-resolution}
			\State \Return{$\omega',\Omega'$}
		\EndProcedure
		
		\Statex
		\Procedure{UnifyDuplicateLiterals}{$\Omega, \Omega', \omega, \omega'$}
%			\textbf{Input:} $\Omega,\Omega'$: standardised-apart non-ground nogoods; $\omega,\omega'$: corresponding ground nogoods\\
%			\textbf{Output:} a unifier
			\State $\gamma \leftarrow \omega \cap \omega'$
			\State $\substitution \leftarrow$ empty unifier
			\For{$l' \in \gamma$}
				\State $L \leftarrow$ non-ground literal s.t.\ $\omega[i] = l'$ and $\Omega[i] = L$ for some $i$
				\State $L' \leftarrow$ non-ground literal s.t.\ $\omega'[j] = l'$ and $\Omega'[j] = L'$ for some $j$
				\State $\substitution \leftarrow \substitution \circ \mathrm{unify}(L \substitution, L' \substitution)$
			\EndFor
			\State \Return{$\substitution$}
		\EndProcedure
	\end{algorithmic}
	\caption{CDNL and conflict generalisation}
	\label{alg:AnalyzeConflict}
\end{algorithm}

Our main conflict generalisation algorithm is shown in \cref{alg:AnalyzeConflict}.
We represent nogoods by fixed-order multisets to be able to map between literals in ground and non-ground nogoods.
The algorithm takes as input a violated ground nogood $\omega$ %(in CDCL terminology: the antecedent of the empty clause $\emptyclause$)
and the current assignment $\assignment$.
If the conflict occurred at decision level 0, UNSAT is returned in \cref{alg:conflict-generalisation:AnalyzeConflict:return-UNSAT}, otherwise two lists of learned nogoods are returned in \cref{alg:conflict-generalisation:AnalyzeConflict:return-lists}.
The first list contains ground nogoods and the second list contains non-ground nogoods.
Each element of both lists corresponds to one UIP and both lists are ordered by distance from the conflict, i.e., the first element of both lists is a nogood learned at the first UIP and the last element of both lists is a nogood learned at the last UIP.

As long as the current nogood contains a literal from the current decision level\footnote{It would also be correct to choose a literal from a lower decision level for resolution, even though this is not done in CDNL.
	We will use such an extension of \cref{alg:AnalyzeConflict} for parts of our experiments reported on in \cref{sec:experiments}.} with non-\nil\ antecedent (\cref{alg:conflict-generalisation:AnalyzeConflict:while}),
it is resolved with the antecedent of such a literal to produce a new current nogood $\omega$ (the resolvent, \cref{alg:conflict-generalisation:AnalyzeConflict:Resolve}).
If $\omega$ contains exactly one literal on the current decision level (\cref{alg:conflict-generalisation:AnalyzeConflict:UIP}), we have found a UIP and remember the current nogood (both ground and non-ground).

Resolution of ground nogoods is straightforward and well-known (\cref{alg:resolution:ground-resolution}):
The resolvent is the union of two input nogoods ($\omega$ and the antecedent of $l$, $\antecedent(l)$), minus $l$ that occurs in the first input nogood and its complement $\bar{l}$ that occurs in the second input nogood.

Following the resolution on the non-ground level is more complex, however, and several special cases have to be considered.
First, the non-ground nogood corresponding to $\antecedent(l)$ is identified and standardised apart from the current nogood (i.e., variables are renamed to avoid overlaps; \cref{alg:resolution:identify-ngng}).\footnote{If no non-ground nogood is available for $\antecedent(l)$, conflict generalisation can continue with another literal from the current decision level or just learn the current resolvent.}
Then, two steps of unification are necessary:
First, in \cref{alg:resolution:unify-resolution-literal}, the two complementary resolution literals are unified s.t.\ variable occurrences are correctly updated in the resolvent.
Then, an additional step is necessary only on the non-ground level:
If the two ground input nogoods share some literals, duplicates are just removed during resolution, because a nogood is a \emph{set} of literals.
On the non-ground level, however,
for each ground literal that occurs in both input nogoods, the corresponding non-ground literals must be unified before one of them can be removed (\cref{alg:resolution:unify-duplicate-literals}).
This is a restricted form of factoring that is guided by ground resolution.
Note that in the context of CDNL, it is not possible for two input nogoods to contain complementary literals apart from the resolution literal, because all antecedents must be unit to entail a literal.
Finally, after applying the computed unifier to both non-ground nogoods, the resolution step is the same as on the ground level (\cref{alg:resolution:nonground-resolution}).

The connection between our suggestion of non-ground CDNL and EBL becomes apparent when viewing the violated nogood as the training instance.
The implication graph utilised by CDNL constitutes a proof that the nogood is violated by giving a derivation for each literal in the nogood.
Each arc in the implication graph that originates from unit propagation can be seen as a definite Horn clause that derives the literal assigned by the propagation step.
Operationality in our setting is specified by demanding that nogoods are only learned at UIPs.
The set of predicates allowed in learned nogoods is not restricted, because in ASP there are no predicates that are \enquote{easier to solve} than others.
In lazy grounding, rules are even grounded only when (part of) their body is already satisfied \cite{DBLP:conf/lpnmr/TaupeWF19}, which alleviates the overhead of additional rules.
Currently, facts from the ASP program are not regarded by our approach because they are not included in nogoods generated by state-of-the-art answer-set solvers.
This ensures that learned constraints depend only on the rules from the problem encoding.

\Cref{alg:AnalyzeConflict} is correct because it derives constraints that are implied when the input program $\prog$ is interpreted as an FOL program, and because such constraints are also true in all stable models of $\prog$, as has been discussed in \cref{sec:preliminaries:asp}.

\paragraph{Replacing internal atoms in learned constraints.}
%\label{sec:conflict-generalisation:replacing-internals}
Learned constraints may contain literals of solver-internal predicates.
Literals of two kinds of such predicates, namely body-representing and choice predicates, can easily be replaced by equivalent (sets of) literals of predicates from the problem encoding.
Only then can resulting constraints be correctly added to the input program.

A body-representing atom such as $\nongroundchoiceatom{r}$ represents the body of non-ground rule $r$.
If a nogood contains a positive literal of such an atom, by definition this literal can just be replaced by the body of $r$.
This is equivalent to resolving with nogood \labelcref{eq:nogood-beta-to-full-body} above.
If the head of $r$ is the head of no other rule, $\nongroundchoiceatom{r}$ can also be replaced by $\head(r)$ by resolving with the support nogood \labelcref{eq:support-nogood}.
If a nogood contains a negative literal of a body-representing atom, it can be replaced by the negated head of $r$ (which is equivalent to resolving with nogood \labelcref{eq:nogood-neg-head-to-beta} above) or by an arbitrary negated body literal of $r$ (equivalent to resolving with any of the nogoods in \labelcref{eq:nogood-beta-to-posbodylit} or \labelcref{eq:nogood-beta-to-negbodylit}).

For choice rules, internal literals of a different kind are created, as has been described in \cref{sec:preliminaries:asp}.
Because of this, atoms $\widehat{a}$ may occur in learned nogoods.
Due to the way choice rules are translated to normal rules, the literal $\sigF~\widehat{a}$ is equivalent to $\sigT~a$ and can simply be replaced.
Similarly, $\sigT~\widehat{a}$ is equivalent to $\sigF~a$.

\paragraph{Choosing effective constraints.}
%\label{sec:conflict-generalisation:choosing}

Since the number of learnt constraints might be overwhelming, strategies to choose effective ones are of vital importance.
Our approach is simple but effective:
We count how many conflicts could be avoided if the additional non-ground constraints were already included in the input program.

Each ground nogood that is violated during a run of CDNL belongs to a class of nogoods that share the same non-ground nogood.
When running \cref{alg:AnalyzeConflict} upon such a nogood violation, the lists of learned non-ground nogoods are associated with this class.
When conflict generalisation terminates, the learned nogoods are printed together with the number of violations of their associated non-ground nogood, s.t.\ a (human) user can then select the most useful constraints.
Since learned constraints can of course be equivalent to each other when variables are renamed, only one unique representation of each nogood is remembered.
For example, the class of ground nogoods violated most often while \alphaslv\ solves the HCP (\cref{sec:example}) is identified by the non-ground nogood $\{ \sigT~\mr{personTOroom}(\mi{P1}, R),$ $\sigT~\mr{personTOroom}(\mi{P2}, R),$ $\mi{P1} < \mi{P2} \}$.

Every UIP provides the opportunity to learn a constraint.%, as has already been described above.
We focused on the first and last UIP in our experiments, because
ground CDNL very successfully learns only from the first UIP, %based on empirical indicators,
and because we expect constraints from the last UIP also to be useful since they contain the decision from the current decision level.
Investigating the usefulness of %nogoods learned on
other UIPs, and finding other quality criteria to discriminate among learned constraints, belong to future work.

%\subsection{Conflict generalisation with limited resources.}
%\label{sec:conflict-generalisation:limits}

If no limit is imposed on the conflict generalisation algorithm, it runs until the problem is solved (one or more answer sets are found, or unsatisfiability is proven).
To increase efficiency, resource consumption may be limited by stopping
after a certain time or number of conflicts and then collecting results as if the problem had been solved.
In our experiments, conflict generalisation was very effective even if resources were heavily limited, because the class of nogoods violated most often emerged at the very beginning of the solving process.

\subsection{Continuation of the motivating example}

The House Configuration Problem (HCP) was presented as a motivating example in \cref{sec:example}.
We have already presented the constraint \labelcref{eq:1uip_constraint} that can automatically be learned at the first UIP.

When continuing along the implication graph until the last UIP, two different non-ground nogoods can be learned: namely, $\Omega_\mathrm{min} \cup \{ \nongroundchoiceatom{r_\mathrm{r2c}} \}$ and $\Omega_\mathrm{min} \cup \body(r_\mathrm{r2c})$, where:
\begin{align*}
	\Omega_\mathrm{min} =	&~\{ \sigT~\mr{cabinetTOthing}(C, \mi{T1}), \sigT~\mr{personTOthing}(\mi{P1}, \mi{T1}),							\\
							&~~~ \sigT~\mr{cabinetTOthing}(C, \mi{T2}), \sigT~\mr{personTOthing}(\mi{P2}, \mi{T2}), \mi{P1} < \mi{P2} \}	\\
	r_\mathrm{r2c}\colon	& \mr{roomTOcabinet}(R,C) \leftarrow \mr{roomDomain}(R), \mr{cabinet}(C), \naf{\widehat{\mr{roomTOcabinet}(R,C)}}.
\end{align*}
In both nogoods, internal atoms can be replaced by ordinary atoms as described above to yield the following unique non-ground nogood\footnote{We use nogood notation and constraint notation interchangeably for learned nogoods.} learned at the last UIP:
\begin{align}
\label{eq:luip_constraint}
\leftarrow\;
& \mr{roomTOcabinet}(R, C), \underline{\smash{\mr{roomDomain}(R)}}, \underline{\smash{\mr{cabinet}(C)}}, \nonumber \\
& \mr{cabinetTOthing}(C, \mi{T1}), \mr{personTOthing}(\mi{P1}, \mi{T1}), \nonumber \\
& \mr{cabinetTOthing}(C, \mi{T2}), \mr{personTOthing}(\mi{P2}, \mi{T2}),
  \mi{P1} < \mi{P2}.
\end{align}
This nogood contains additional literals of domain predicates (underlined for emphasis),
which is the only difference to the nogood \labelcref{eq:1uip_constraint} learned at the first UIP.

Reduction of the non-minimal learned nogood as described in \cref{sec:example} has not yet been implemented in our system.
An implementation based on first-order theorem proving is conceivable, in which undecidability could be avoided by imposing a bound on the number of constants, e.g., a maximal number of entities in a configuration.

\section{Experimental Results}
\label{sec:experiments}

We conducted a set of experiments on encodings of the House Reconfiguration Problem (HRP) and on a Graph Colouring problem (\cref{lst:3cc}) to demonstrate the feasibility of our approach.
HRP extends HCP from \cref{sec:example}, which has disregarded reconfiguration.
All encodings and instances used for our experiments are available on our website.\footnote{\url{http://ainf.aau.at/dynacon}}
The HRP encoding was closely based on the encoding by \citeN{Ryabokon:Dissertation}, except that the redundant constraint \labelcref{eq:redundant_constraint} has been removed, as described in \cref{sec:example}, and due to syntactic restrictions of \alphaslv\ some aggregates had to be rewritten and optimization statements were not used at all.
Note that for reasons of a fair comparison, all solvers used the same encodings in our experiments, even though solvers supporting aggregates could have profited from a more sophisticated encoding.

Graph Colouring problems are an abstraction to which many real-world problems can be mapped.
We have designed 100 satisfiable and 100 unsatisfiable graph instances (\enquote{3CC} for \enquote{3-colourable chains}) containing repeated patterns that force some pairs of nodes to have the same colour.
On 3CC, our approach was able to learn constraints that represent this pattern.
This was only possible by adapting procedure \procedure{FindNextLiteralForResolution} in \cref{alg:AnalyzeConflict} to also use literals from the next-lower decision level ($\dl(l) \geq \dl(\assignment)-1$) for resolution.
The question how many decision levels to consider is an interesting topic for future work.
From the constraints learned from the conflicts occurring most often, one from the first UIP and one from the last UIP were chosen manually for the experiments.
Since these constraints represent only the conflict for one specific combination of the three colours, they were multiplied manually to cover all possible combinations.\footnote{Each colour is represented by a predicate in our encoding.
Depending on the actual conflicts during a specific solver run, constraints representing other colour combinations may be produced automatically.}
The constraints learned at the first UIP were then reduced manually, similarly as has been described in \cref{sec:example}, yielding the following final constraint \labelcref{eq:3cc_reduced_constraints} for red (and two more constraints for the other colours):
\begin{align}
\label{eq:3cc_reduced_constraints}
\leftarrow\;
& \mr{red}(\mi{N12}), \naf{\mr{red}(\mi{N22})}, \nonumber \\
& \mr{link}(\mi{N12}, \mi{N11}), \mr{link}(\mi{N12}, \mi{N21}), \mr{link}(\mi{N11}, \mi{N21}), \mr{link}(\mi{N11}, \mi{N22}), \mr{link}(\mi{N21}, \mi{N22}). 
\end{align}

\begin{lstlisting}[float,label={lst:3cc},caption={An encoding for our Graph Colouring Problem}]
blue(N) :- node(N), not red(N), not green(N).
red(N) :- node(N), not blue(N), not green(N).
green(N) :- node(N), not red(N), not blue(N).
:- link(N1,N2), blue(N1), blue(N2).
:- link(N1,N2), red(N1), red(N2).
:- link(N1,N2), green(N1), green(N2).
\end{lstlisting}

The ASP solvers \alphaslv\footnote{\url{https://github.com/alpha-asp/Alpha}} v0.5.0 \cite{DBLP:conf/lpnmr/Weinzierl17}, \slv{dlv} 2.0 \cite{DBLP:conf/lpnmr/AlvianoCDFLPRVZ17}, and \slv{clingo} 5.4.0 \cite{DBLP:journals/corr/GebserKKS14} were used.
\alphaslv\ was configured to ground rules strictly lazily and constraints permissively, as recommended by \citeN{DBLP:conf/lpnmr/TaupeWF19}.
The JVM was called with parameters \texttt{-Xms1G -Xmx32G}.
For each problem instance, solvers searched for 10 answer sets.\footnote{Obtaining more than one (maybe trivial) answer set is often desirable. The number 10 has been chosen arbitrarily.}

Experiments were run on %a cluster of
machines each with two
Intel\textsuperscript{\textregistered} Xeon\textsuperscript{\textregistered} CPU E5-2650 v4 @ 2.20GHz with 12 cores each, 252 GB of memory, and Ubuntu 16.04.1 LTS Linux.
Benchmarks were scheduled with ABC Benchmarking System \cite{DBLP:conf/aiia/Redl16} and HTCondor\textsuperscript{\texttrademark}.\footnote{\url{https://github.com/credl/abcbenchmarking}, \url{http://research.cs.wisc.edu/htcondor}}
\slv{pyrunlim}\footnote{\url{https://alviano.com/software/pyrunlim/}} was used to measure time and memory consumption
and to limit time consumption to 10 minutes per instance, memory to 40 GiB and swapping to 0.
Care was taken to avoid interference between CPUs, e.g., by not running different benchmarks concurrently on the same machine.
% for the use of \textregistered and \texttrademark, cf.\ https://academia.stackexchange.com/questions/21521/is-it-mandatory-to-include-the-registered-trademark-symbol-next-to-the-name-of

% two figures side by side: https://tex.stackexchange.com/a/37597/82973
\begin{figure}
	\centering
	\begin{minipage}{\textwidth}
		\begin{minipage}{.45\textwidth}
			\centering
			\resizebox{\linewidth}{!}{\input{figures/cactus_hrp_kco_permissive.pgf}}
			\caption{HRP results with \alphaslv}%, $\kco = \infty$}
			\label{fig:cactus_hrp_kco_permissive}
		\end{minipage}%
		\hfill
		\begin{minipage}{.45\textwidth}
			\centering
			\resizebox{\linewidth}{!}{\input{figures/cactus_3col_crafted_kco_permissive.pgf}}
			\caption{3CC results with \alphaslv}%, $\kco = \infty$}
			\label{fig:cactus_3cc_kco_permissive}
		\end{minipage}%
	\end{minipage}%
	\vspace{\floatsep}
	\begin{minipage}{\textwidth}
		\begin{minipage}{.45\textwidth}
			\centering
			\resizebox{\linewidth}{!}{\input{figures/cactus_hrp_dlv2.pgf}}
			\caption{HRP results with \slv{dlv2}}
			\label{fig:cactus_hrp_dlv2}
		\end{minipage}%
		\hfill
		\begin{minipage}{.45\textwidth}
			\centering
			\resizebox{\linewidth}{!}{\input{figures/cactus_3col_crafted_dlv2.pgf}}
			\caption{3CC results with \slv{dlv2}}
			\label{fig:cactus_3cc_dlv2}
		\end{minipage}%
	\end{minipage}%
	\vspace{\floatsep}
	\begin{minipage}{\textwidth}
		\begin{minipage}{.45\textwidth}
			\centering
			\resizebox{\linewidth}{!}{\input{figures/cactus_hrp_clingo.pgf}}
			\caption{HRP results with \slv{clingo}}
			\label{fig:cactus_hrp_clingo}
		\end{minipage}%
		\hfill
		\begin{minipage}{.45\textwidth}
			\centering
			\resizebox{\linewidth}{!}{\input{figures/cactus_3col_crafted_clingo.pgf}}
			\caption{3CC results with \slv{clingo}}
			\label{fig:cactus_3cc_clingo}
		\end{minipage}%
	\end{minipage}%
\end{figure}

\begin{table}
	\caption{Number of HRP instances solved by \slv{clingo} (\textbf{\#}) per encoding; grounding and solving times in seconds (1st, \textbf{2nd}, 3rd quartile)}
	\label{tab:clingo_details}
	\begin{tabular}{ l r @{\hskip 4em} r r r @{\hskip 4em} r r r }
		\hline\hline
		                  & 			& \multicolumn{3}{c}{\textbf{grounding time}}		& \multicolumn{3}{c}{\textbf{solving time}} \\
		\textbf{Encoding} & \textbf{\#} & Q1 & Q2 & Q3	& Q1 & Q2 & Q3 \\
		\hline
		original	& 34  & 6.787 & \textbf{39.738} & ---			 	& 4.315 & \textbf{66.580} & --- \\
		\hline
		1st UIP		& 22 & 32.150 & \textbf{---} & ---			& 1.030 & \textbf{---} & --- \\
		\hline
		last UIP	& 22 & 34.833 & \textbf{---} & ---			& 0.940 & \textbf{---} & --- \\
		\hline
		reduced		& 38 & 7.850 & \textbf{52.026} & 244.256				& 1.070 & \textbf{9.270} & 90.575 \\
		\hline\hline
	\end{tabular}
\end{table}

To compare solving performance using encodings with and without learned constraints, cactus plots (\cref{fig:cactus_hrp_kco_permissive,fig:cactus_hrp_clingo,fig:cactus_hrp_dlv2,fig:cactus_3cc_kco_permissive,fig:cactus_3cc_clingo,fig:cactus_3cc_dlv2}) have been created in the usual way. The x axis gives the number of instances solved within real (i.e., wall-clock) time which is given on the y axis.
Time is accumulated over all solved instances.
Since we are investigating the effects of constraints on each solver and not comparing solvers against each other,
maximum axis values vary between solvers.
One curve has been drawn for each encoding: the original encoding, one encoding each with the additional constraint(s) learned automatically at the first/last UIP (\labelcref{eq:1uip_constraint}/\labelcref{eq:luip_constraint} for HRP), and one encoding in which only the reduced constraint(s) (\labelcref{eq:redundant_constraint} for HRP, \labelcref{eq:3cc_reduced_constraints} for 3CC) are added to the original encoding.

Learned constraints improved solving performance in many cases.
For HRP, \alphaslv\ benefited especially from the reduced constraint, but also from the constraint learned automatically at the first UIP.
\slv{dlv2} profited most from the first-UIP constraint, but the reduced one was not far behind.
The last-UIP constraint affected \slv{dlv2}'s performance negatively, however.
\slv{clingo} profited only from the reduced constraint, while automatically learned constraints caused overall solving time to increase.
Detailed analysis of grounding and solving times as reported by \slv{clingo} (\cref{tab:clingo_details}) shows that this was due to additional grounding effort induced by learned constraints.
With automatically learned constraints, more than half of the instances could not be solved within 10 minutes, therefore no median times (second quartiles) can be shown.
In almost all time-out cases, \slv{clingo} did not manage to finish grounding and start solving.
% this has been determined by manual inspection of clingo output files.
% only for the following instances clingo was killed during solving, for all others it was killed during grounding:
% inputs/learned_manual/hrp/emptyconfig_p75t375.asp
% inputs/original/hrp/emptyconfig_p55t275.asp
% inputs/original/hrp/emptyconfig_p60t300.asp
% inputs/original/hrp/emptyconfig_p65t325.asp
% inputs/original/hrp/emptyconfig_p70t350.asp
% inputs/original/hrp/emptyconfig_p75t375.asp
% inputs/original/hrp/emptyconfig_p80t400.asp
For 3CC, all solvers perform best with the reduced constraints, while the 1st-UIP constraints also improve performance.

%\rttodo{show EHRP results or not? (lack of space!)}
%
%To demonstrate the scalability of our approach, we also experimented with a novel extension of HRP in which multiple layers of hierarchy are additionally introduced that are also typical of real-world configuration problems.
%We call this variant of the problem \emph{Extended House Reconfiguration Problem} (EHRP).

Learning itself is cheap:
On the first author's computer, the relevant HRP constraints are learned using the easiest instance %(\texttt{emptyconfig\_p05t025.asp})
in less than 6 seconds if search is stopped after 50 conflicts and in less than 14 seconds if search is carried out until the first answer set is found (our implementation has not yet been tuned for optimal performance).
Performance for 3CC is similar.
Computational complexity has not yet been analysed and could be addressed in future work.

The results show that our approach is able to improve ASP solving using both lazy-grounding and ground-and-solve systems.
While effects vary between types of learned constraints and solver implementations, every system under investigation has profited from at least one learned constraint in both domains.
%\alphaslv\ solved up to 64\% more HRP instances employing learned constraints, \slv{dlv2} up to 41\% more, and \slv{clingo} solved 12\% more instances with the reduced constraint compared to without it.

\section{Conclusions}
\label{sec:conclusions}

We have proposed an extension of CDNL that, while solving one problem instance, learns non-ground nogoods that can be used to speed up the solving of other problem instances.
As far as we know, this is the first attempt to generalise and re-use knowledge learned during ASP solving.
Experimental results showed compelling benefits of our approach:
Both ground-and-solve systems and lazy-grounding systems performed significantly better on instances of a practical configuration problem and a graph colouring problem when using constraints learned by our method, solving more instances and/or solving the problems faster.

So far, we have been experimenting with encodings for two problem domains: a graph colouring problem, and an important part of many configuration domains (those where systems are composed hierarchically).
However, it seems natural to assume that many domains feature redundant constraints that may not be obvious to a human modeller.
We therefore see our approach mainly as a tool to support the design of efficient answer-set programs.

It remains to be clarified whether our approach could improve encodings used by the ASP competitions \cite{DBLP:journals/tplp/GebserMR20}, which generally are already heavily optimised, or whether it would prove more useful when applied to encodings devised by inexperienced modellers.

\paragraph{Acknowledgments.}
This work has been conducted in the scope of the research project \textit{DynaCon (FFG-PNr.:\ 861263)}, which is funded by the Austrian Federal Ministry of Transport, Innovation and Technology (BMVIT) under the program \enquote{ICT of the Future} between 2017 and 2020,\footnote{See \url{https://iktderzukunft.at/en/} for more information.} and in the scope of the research project Productive4.0, which is funded by EU-ECSEL under grant agreement no737459.

%\pagebreak

\bibliographystyle{acmtrans}
% handling of "van": [https://tex.stackexchange.com/a/40750/82973]
% here we change the meaning of \VAN to use the prefix for the bibliography
\DeclareRobustCommand{\VAN}[3]{#3}
\bibliography{conflict_generalisation}

\label{lastpage}
\end{document}

%% file: macros.tex
\usepackage{soul}
\usepackage[textsize=footnotesize,disable]{todonotes}
\presetkeys{todonotes}{inline}{}

\newcommand{\mi}[1]{\ensuremath{\mathit{#1}}}
\newcommand{\mr}[1]{\ensuremath{\mathrm{#1}}}

\newcommand{\slv}[1]{\textsc{#1}} % for ASP solvers (systems)
\newcommand{\alphaslv}{\slv{Alpha}}

\newcommand{\prog}{\ensuremath{P}}
\newcommand{\universe}{\ensuremath{U}}

\newcommand{\assignment}{\ensuremath{A}}
\newcommand{\nafsymbol}{\ensuremath{\mathrm{not}}}
\newcommand{\naf}[1]{\ensuremath{\nafsymbol~#1}}
\newcommand{\head}{\ensuremath{\mathrm{H}}}
\newcommand{\body}{\ensuremath{\mathrm{B}}}

\newcommand{\sigT}{\ensuremath{\mathbf{T}}}
\newcommand{\sigF}{\ensuremath{\mathbf{F}}}
\newcommand{\sigM}{\ensuremath{\mathbf{M}}}
\newcommand{\sigU}{\ensuremath{\mathbf{U}}}

\newcommand{\vars}{\ensuremath{\mathrm{vars}}}

\newcommand{\atom}{\ensuremath{a}}
\newcommand{\atoms}{\ensuremath{\mathcal{A}}}

\newcommand{\literals}{\ensuremath{\mathcal{L}}}
\newcommand{\antecedent}{\ensuremath{\alpha}}
\newcommand{\dl}{\ensuremath{\delta}}
\newcommand{\nil}{\ensuremath{\mathrm{NIL}}}
\newcommand{\assignedtruth}{\ensuremath{\nu}}

\newcommand{\allvars}{\ensuremath{\mathcal{V}}}
\newcommand{\substitution}{\ensuremath{\sigma}}
\newcommand{\facts}{\ensuremath{\mathrm{facts}}}
\newcommand{\choiceatom}[2]{\ensuremath{\beta(#1,#2)}}
\newcommand{\nongroundchoiceatom}[1]{\ensuremath{\beta(#1,\vars(#1))}}
\newcommand{\procedure}[1]{\textsc{#1}}

\newcommand{\InA}{\mi{InA}}

\newcommand{\GC}{\mi{GC}}

\newcommand{\InPred}{\mi{InPred}}
\newcommand{\OutPred}{\mi{OutPred}}
\newcommand{\comp}{\mr{comp}}
\newcommand{\compR}{\mr{rComp}}
\newcommand{\modelsSkeptically}{\ensuremath{\models_\mathrm{S}}}
\newcommand{\modelsFOL}{\ensuremath{\models_\mathrm{F}}}
\newcommand{\encoding}{\ensuremath{\prog^\mathrm{E}}}

\makeatletter
\newcommand{\StatexIndent}[1][3]{%
	\setlength\@tempdima{\algorithmicindent}%
	\Statex\hskip\dimexpr#1\@tempdima\relax}
\usepackage{mathtools}
\newtagform{angles}[\textit]{$\langle$}{$\rangle$}
\usetagform{angles}
\creflabelformat{equation}{#2$\langle$\textit{#1}$\rangle$#3}

\newtheorem{definition}{Definition}[section]
\newtheorem{corollary}{Corollary}[section]

\DeclareRobustCommand{\VAN}[3]{#2} % set up for citation

%% file: figures/cactus_hrp_kco_permissive.pgf
%% Creator: Matplotlib, PGF backend
%%
%% To include the figure in your LaTeX document, write
%%   \input{<filename>.pgf}
%%
%% Make sure the required packages are loaded in your preamble
%%   \usepackage{pgf}
%%
%% and, on pdftex
%%   \usepackage[utf8]{inputenc}\DeclareUnicodeCharacter{2212}{-}
%%
%% or, on luatex and xetex
%%   \usepackage{unicode-math}
%%
%% Figures using additional raster images can only be included by \input if
%% they are in the same directory as the main LaTeX file. For loading figures
%% from other directories you can use the `import` package
%%   \usepackage{import}
%%
%% and then include the figures with
%%   \import{<path to file>}{<filename>.pgf}
%%
%% Matplotlib used the following preamble
%%   \usepackage{fontspec}
%%
\begingroup%
\makeatletter%
\begin{pgfpicture}%
\pgfpathrectangle{\pgfpointorigin}{\pgfqpoint{5.554914in}{4.243888in}}%
\pgfusepath{use as bounding box, clip}%
\begin{pgfscope}%
\pgfsetbuttcap%
\pgfsetmiterjoin%
\definecolor{currentfill}{rgb}{1.000000,1.000000,1.000000}%
\pgfsetfillcolor{currentfill}%
\pgfsetlinewidth{0.000000pt}%
\definecolor{currentstroke}{rgb}{1.000000,1.000000,1.000000}%
\pgfsetstrokecolor{currentstroke}%
\pgfsetdash{}{0pt}%
\pgfpathmoveto{\pgfqpoint{0.000000in}{0.000000in}}%
\pgfpathlineto{\pgfqpoint{5.554914in}{0.000000in}}%
\pgfpathlineto{\pgfqpoint{5.554914in}{4.243888in}}%
\pgfpathlineto{\pgfqpoint{0.000000in}{4.243888in}}%
\pgfpathclose%
\pgfusepath{fill}%
\end{pgfscope}%
\begin{pgfscope}%
\pgfsetbuttcap%
\pgfsetmiterjoin%
\definecolor{currentfill}{rgb}{1.000000,1.000000,1.000000}%
\pgfsetfillcolor{currentfill}%
\pgfsetlinewidth{0.000000pt}%
\definecolor{currentstroke}{rgb}{0.000000,0.000000,0.000000}%
\pgfsetstrokecolor{currentstroke}%
\pgfsetstrokeopacity{0.000000}%
\pgfsetdash{}{0pt}%
\pgfpathmoveto{\pgfqpoint{0.594914in}{0.547888in}}%
\pgfpathlineto{\pgfqpoint{5.554914in}{0.547888in}}%
\pgfpathlineto{\pgfqpoint{5.554914in}{4.243888in}}%
\pgfpathlineto{\pgfqpoint{0.594914in}{4.243888in}}%
\pgfpathclose%
\pgfusepath{fill}%
\end{pgfscope}%
\begin{pgfscope}%
\pgfsetbuttcap%
\pgfsetroundjoin%
\definecolor{currentfill}{rgb}{0.000000,0.000000,0.000000}%
\pgfsetfillcolor{currentfill}%
\pgfsetlinewidth{0.803000pt}%
\definecolor{currentstroke}{rgb}{0.000000,0.000000,0.000000}%
\pgfsetstrokecolor{currentstroke}%
\pgfsetdash{}{0pt}%
\pgfsys@defobject{currentmarker}{\pgfqpoint{0.000000in}{-0.048611in}}{\pgfqpoint{0.000000in}{0.000000in}}{%
\pgfpathmoveto{\pgfqpoint{0.000000in}{0.000000in}}%
\pgfpathlineto{\pgfqpoint{0.000000in}{-0.048611in}}%
\pgfusepath{stroke,fill}%
}%
\begin{pgfscope}%
\pgfsys@transformshift{1.085609in}{0.547888in}%
\pgfsys@useobject{currentmarker}{}%
\end{pgfscope}%
\end{pgfscope}%
\begin{pgfscope}%
\definecolor{textcolor}{rgb}{0.000000,0.000000,0.000000}%
\pgfsetstrokecolor{textcolor}%
\pgfsetfillcolor{textcolor}%
\pgftext[x=1.085609in,y=0.450666in,,top]{\color{textcolor}\fontsize{16.000000}{19.200000}\selectfont \(\displaystyle 2\)}%
\end{pgfscope}%
\begin{pgfscope}%
\pgfsetbuttcap%
\pgfsetroundjoin%
\definecolor{currentfill}{rgb}{0.000000,0.000000,0.000000}%
\pgfsetfillcolor{currentfill}%
\pgfsetlinewidth{0.803000pt}%
\definecolor{currentstroke}{rgb}{0.000000,0.000000,0.000000}%
\pgfsetstrokecolor{currentstroke}%
\pgfsetdash{}{0pt}%
\pgfsys@defobject{currentmarker}{\pgfqpoint{0.000000in}{-0.048611in}}{\pgfqpoint{0.000000in}{0.000000in}}{%
\pgfpathmoveto{\pgfqpoint{0.000000in}{0.000000in}}%
\pgfpathlineto{\pgfqpoint{0.000000in}{-0.048611in}}%
\pgfusepath{stroke,fill}%
}%
\begin{pgfscope}%
\pgfsys@transformshift{1.616090in}{0.547888in}%
\pgfsys@useobject{currentmarker}{}%
\end{pgfscope}%
\end{pgfscope}%
\begin{pgfscope}%
\definecolor{textcolor}{rgb}{0.000000,0.000000,0.000000}%
\pgfsetstrokecolor{textcolor}%
\pgfsetfillcolor{textcolor}%
\pgftext[x=1.616090in,y=0.450666in,,top]{\color{textcolor}\fontsize{16.000000}{19.200000}\selectfont \(\displaystyle 4\)}%
\end{pgfscope}%
\begin{pgfscope}%
\pgfsetbuttcap%
\pgfsetroundjoin%
\definecolor{currentfill}{rgb}{0.000000,0.000000,0.000000}%
\pgfsetfillcolor{currentfill}%
\pgfsetlinewidth{0.803000pt}%
\definecolor{currentstroke}{rgb}{0.000000,0.000000,0.000000}%
\pgfsetstrokecolor{currentstroke}%
\pgfsetdash{}{0pt}%
\pgfsys@defobject{currentmarker}{\pgfqpoint{0.000000in}{-0.048611in}}{\pgfqpoint{0.000000in}{0.000000in}}{%
\pgfpathmoveto{\pgfqpoint{0.000000in}{0.000000in}}%
\pgfpathlineto{\pgfqpoint{0.000000in}{-0.048611in}}%
\pgfusepath{stroke,fill}%
}%
\begin{pgfscope}%
\pgfsys@transformshift{2.146572in}{0.547888in}%
\pgfsys@useobject{currentmarker}{}%
\end{pgfscope}%
\end{pgfscope}%
\begin{pgfscope}%
\definecolor{textcolor}{rgb}{0.000000,0.000000,0.000000}%
\pgfsetstrokecolor{textcolor}%
\pgfsetfillcolor{textcolor}%
\pgftext[x=2.146572in,y=0.450666in,,top]{\color{textcolor}\fontsize{16.000000}{19.200000}\selectfont \(\displaystyle 6\)}%
\end{pgfscope}%
\begin{pgfscope}%
\pgfsetbuttcap%
\pgfsetroundjoin%
\definecolor{currentfill}{rgb}{0.000000,0.000000,0.000000}%
\pgfsetfillcolor{currentfill}%
\pgfsetlinewidth{0.803000pt}%
\definecolor{currentstroke}{rgb}{0.000000,0.000000,0.000000}%
\pgfsetstrokecolor{currentstroke}%
\pgfsetdash{}{0pt}%
\pgfsys@defobject{currentmarker}{\pgfqpoint{0.000000in}{-0.048611in}}{\pgfqpoint{0.000000in}{0.000000in}}{%
\pgfpathmoveto{\pgfqpoint{0.000000in}{0.000000in}}%
\pgfpathlineto{\pgfqpoint{0.000000in}{-0.048611in}}%
\pgfusepath{stroke,fill}%
}%
\begin{pgfscope}%
\pgfsys@transformshift{2.677053in}{0.547888in}%
\pgfsys@useobject{currentmarker}{}%
\end{pgfscope}%
\end{pgfscope}%
\begin{pgfscope}%
\definecolor{textcolor}{rgb}{0.000000,0.000000,0.000000}%
\pgfsetstrokecolor{textcolor}%
\pgfsetfillcolor{textcolor}%
\pgftext[x=2.677053in,y=0.450666in,,top]{\color{textcolor}\fontsize{16.000000}{19.200000}\selectfont \(\displaystyle 8\)}%
\end{pgfscope}%
\begin{pgfscope}%
\pgfsetbuttcap%
\pgfsetroundjoin%
\definecolor{currentfill}{rgb}{0.000000,0.000000,0.000000}%
\pgfsetfillcolor{currentfill}%
\pgfsetlinewidth{0.803000pt}%
\definecolor{currentstroke}{rgb}{0.000000,0.000000,0.000000}%
\pgfsetstrokecolor{currentstroke}%
\pgfsetdash{}{0pt}%
\pgfsys@defobject{currentmarker}{\pgfqpoint{0.000000in}{-0.048611in}}{\pgfqpoint{0.000000in}{0.000000in}}{%
\pgfpathmoveto{\pgfqpoint{0.000000in}{0.000000in}}%
\pgfpathlineto{\pgfqpoint{0.000000in}{-0.048611in}}%
\pgfusepath{stroke,fill}%
}%
\begin{pgfscope}%
\pgfsys@transformshift{3.207534in}{0.547888in}%
\pgfsys@useobject{currentmarker}{}%
\end{pgfscope}%
\end{pgfscope}%
\begin{pgfscope}%
\definecolor{textcolor}{rgb}{0.000000,0.000000,0.000000}%
\pgfsetstrokecolor{textcolor}%
\pgfsetfillcolor{textcolor}%
\pgftext[x=3.207534in,y=0.450666in,,top]{\color{textcolor}\fontsize{16.000000}{19.200000}\selectfont \(\displaystyle 10\)}%
\end{pgfscope}%
\begin{pgfscope}%
\pgfsetbuttcap%
\pgfsetroundjoin%
\definecolor{currentfill}{rgb}{0.000000,0.000000,0.000000}%
\pgfsetfillcolor{currentfill}%
\pgfsetlinewidth{0.803000pt}%
\definecolor{currentstroke}{rgb}{0.000000,0.000000,0.000000}%
\pgfsetstrokecolor{currentstroke}%
\pgfsetdash{}{0pt}%
\pgfsys@defobject{currentmarker}{\pgfqpoint{0.000000in}{-0.048611in}}{\pgfqpoint{0.000000in}{0.000000in}}{%
\pgfpathmoveto{\pgfqpoint{0.000000in}{0.000000in}}%
\pgfpathlineto{\pgfqpoint{0.000000in}{-0.048611in}}%
\pgfusepath{stroke,fill}%
}%
\begin{pgfscope}%
\pgfsys@transformshift{3.738015in}{0.547888in}%
\pgfsys@useobject{currentmarker}{}%
\end{pgfscope}%
\end{pgfscope}%
\begin{pgfscope}%
\definecolor{textcolor}{rgb}{0.000000,0.000000,0.000000}%
\pgfsetstrokecolor{textcolor}%
\pgfsetfillcolor{textcolor}%
\pgftext[x=3.738015in,y=0.450666in,,top]{\color{textcolor}\fontsize{16.000000}{19.200000}\selectfont \(\displaystyle 12\)}%
\end{pgfscope}%
\begin{pgfscope}%
\pgfsetbuttcap%
\pgfsetroundjoin%
\definecolor{currentfill}{rgb}{0.000000,0.000000,0.000000}%
\pgfsetfillcolor{currentfill}%
\pgfsetlinewidth{0.803000pt}%
\definecolor{currentstroke}{rgb}{0.000000,0.000000,0.000000}%
\pgfsetstrokecolor{currentstroke}%
\pgfsetdash{}{0pt}%
\pgfsys@defobject{currentmarker}{\pgfqpoint{0.000000in}{-0.048611in}}{\pgfqpoint{0.000000in}{0.000000in}}{%
\pgfpathmoveto{\pgfqpoint{0.000000in}{0.000000in}}%
\pgfpathlineto{\pgfqpoint{0.000000in}{-0.048611in}}%
\pgfusepath{stroke,fill}%
}%
\begin{pgfscope}%
\pgfsys@transformshift{4.268497in}{0.547888in}%
\pgfsys@useobject{currentmarker}{}%
\end{pgfscope}%
\end{pgfscope}%
\begin{pgfscope}%
\definecolor{textcolor}{rgb}{0.000000,0.000000,0.000000}%
\pgfsetstrokecolor{textcolor}%
\pgfsetfillcolor{textcolor}%
\pgftext[x=4.268497in,y=0.450666in,,top]{\color{textcolor}\fontsize{16.000000}{19.200000}\selectfont \(\displaystyle 14\)}%
\end{pgfscope}%
\begin{pgfscope}%
\pgfsetbuttcap%
\pgfsetroundjoin%
\definecolor{currentfill}{rgb}{0.000000,0.000000,0.000000}%
\pgfsetfillcolor{currentfill}%
\pgfsetlinewidth{0.803000pt}%
\definecolor{currentstroke}{rgb}{0.000000,0.000000,0.000000}%
\pgfsetstrokecolor{currentstroke}%
\pgfsetdash{}{0pt}%
\pgfsys@defobject{currentmarker}{\pgfqpoint{0.000000in}{-0.048611in}}{\pgfqpoint{0.000000in}{0.000000in}}{%
\pgfpathmoveto{\pgfqpoint{0.000000in}{0.000000in}}%
\pgfpathlineto{\pgfqpoint{0.000000in}{-0.048611in}}%
\pgfusepath{stroke,fill}%
}%
\begin{pgfscope}%
\pgfsys@transformshift{4.798978in}{0.547888in}%
\pgfsys@useobject{currentmarker}{}%
\end{pgfscope}%
\end{pgfscope}%
\begin{pgfscope}%
\definecolor{textcolor}{rgb}{0.000000,0.000000,0.000000}%
\pgfsetstrokecolor{textcolor}%
\pgfsetfillcolor{textcolor}%
\pgftext[x=4.798978in,y=0.450666in,,top]{\color{textcolor}\fontsize{16.000000}{19.200000}\selectfont \(\displaystyle 16\)}%
\end{pgfscope}%
\begin{pgfscope}%
\pgfsetbuttcap%
\pgfsetroundjoin%
\definecolor{currentfill}{rgb}{0.000000,0.000000,0.000000}%
\pgfsetfillcolor{currentfill}%
\pgfsetlinewidth{0.803000pt}%
\definecolor{currentstroke}{rgb}{0.000000,0.000000,0.000000}%
\pgfsetstrokecolor{currentstroke}%
\pgfsetdash{}{0pt}%
\pgfsys@defobject{currentmarker}{\pgfqpoint{0.000000in}{-0.048611in}}{\pgfqpoint{0.000000in}{0.000000in}}{%
\pgfpathmoveto{\pgfqpoint{0.000000in}{0.000000in}}%
\pgfpathlineto{\pgfqpoint{0.000000in}{-0.048611in}}%
\pgfusepath{stroke,fill}%
}%
\begin{pgfscope}%
\pgfsys@transformshift{5.329459in}{0.547888in}%
\pgfsys@useobject{currentmarker}{}%
\end{pgfscope}%
\end{pgfscope}%
\begin{pgfscope}%
\definecolor{textcolor}{rgb}{0.000000,0.000000,0.000000}%
\pgfsetstrokecolor{textcolor}%
\pgfsetfillcolor{textcolor}%
\pgftext[x=5.329459in,y=0.450666in,,top]{\color{textcolor}\fontsize{16.000000}{19.200000}\selectfont \(\displaystyle 18\)}%
\end{pgfscope}%
\begin{pgfscope}%
\definecolor{textcolor}{rgb}{0.000000,0.000000,0.000000}%
\pgfsetstrokecolor{textcolor}%
\pgfsetfillcolor{textcolor}%
\pgftext[x=3.074914in,y=0.197555in,,top]{\color{textcolor}\fontsize{16.000000}{19.200000}\selectfont Number of instances}%
\end{pgfscope}%
\begin{pgfscope}%
\pgfsetbuttcap%
\pgfsetroundjoin%
\definecolor{currentfill}{rgb}{0.000000,0.000000,0.000000}%
\pgfsetfillcolor{currentfill}%
\pgfsetlinewidth{0.803000pt}%
\definecolor{currentstroke}{rgb}{0.000000,0.000000,0.000000}%
\pgfsetstrokecolor{currentstroke}%
\pgfsetdash{}{0pt}%
\pgfsys@defobject{currentmarker}{\pgfqpoint{-0.048611in}{0.000000in}}{\pgfqpoint{0.000000in}{0.000000in}}{%
\pgfpathmoveto{\pgfqpoint{0.000000in}{0.000000in}}%
\pgfpathlineto{\pgfqpoint{-0.048611in}{0.000000in}}%
\pgfusepath{stroke,fill}%
}%
\begin{pgfscope}%
\pgfsys@transformshift{0.594914in}{0.708847in}%
\pgfsys@useobject{currentmarker}{}%
\end{pgfscope}%
\end{pgfscope}%
\begin{pgfscope}%
\definecolor{textcolor}{rgb}{0.000000,0.000000,0.000000}%
\pgfsetstrokecolor{textcolor}%
\pgfsetfillcolor{textcolor}%
\pgftext[x=0.387623in, y=0.631736in, left, base]{\color{textcolor}\fontsize{16.000000}{19.200000}\selectfont \(\displaystyle 0\)}%
\end{pgfscope}%
\begin{pgfscope}%
\pgfsetbuttcap%
\pgfsetroundjoin%
\definecolor{currentfill}{rgb}{0.000000,0.000000,0.000000}%
\pgfsetfillcolor{currentfill}%
\pgfsetlinewidth{0.803000pt}%
\definecolor{currentstroke}{rgb}{0.000000,0.000000,0.000000}%
\pgfsetstrokecolor{currentstroke}%
\pgfsetdash{}{0pt}%
\pgfsys@defobject{currentmarker}{\pgfqpoint{-0.048611in}{0.000000in}}{\pgfqpoint{0.000000in}{0.000000in}}{%
\pgfpathmoveto{\pgfqpoint{0.000000in}{0.000000in}}%
\pgfpathlineto{\pgfqpoint{-0.048611in}{0.000000in}}%
\pgfusepath{stroke,fill}%
}%
\begin{pgfscope}%
\pgfsys@transformshift{0.594914in}{1.119268in}%
\pgfsys@useobject{currentmarker}{}%
\end{pgfscope}%
\end{pgfscope}%
\begin{pgfscope}%
\definecolor{textcolor}{rgb}{0.000000,0.000000,0.000000}%
\pgfsetstrokecolor{textcolor}%
\pgfsetfillcolor{textcolor}%
\pgftext[x=0.387623in, y=1.042157in, left, base]{\color{textcolor}\fontsize{16.000000}{19.200000}\selectfont \(\displaystyle 3\)}%
\end{pgfscope}%
\begin{pgfscope}%
\pgfsetbuttcap%
\pgfsetroundjoin%
\definecolor{currentfill}{rgb}{0.000000,0.000000,0.000000}%
\pgfsetfillcolor{currentfill}%
\pgfsetlinewidth{0.803000pt}%
\definecolor{currentstroke}{rgb}{0.000000,0.000000,0.000000}%
\pgfsetstrokecolor{currentstroke}%
\pgfsetdash{}{0pt}%
\pgfsys@defobject{currentmarker}{\pgfqpoint{-0.048611in}{0.000000in}}{\pgfqpoint{0.000000in}{0.000000in}}{%
\pgfpathmoveto{\pgfqpoint{0.000000in}{0.000000in}}%
\pgfpathlineto{\pgfqpoint{-0.048611in}{0.000000in}}%
\pgfusepath{stroke,fill}%
}%
\begin{pgfscope}%
\pgfsys@transformshift{0.594914in}{1.529688in}%
\pgfsys@useobject{currentmarker}{}%
\end{pgfscope}%
\end{pgfscope}%
\begin{pgfscope}%
\definecolor{textcolor}{rgb}{0.000000,0.000000,0.000000}%
\pgfsetstrokecolor{textcolor}%
\pgfsetfillcolor{textcolor}%
\pgftext[x=0.387623in, y=1.452577in, left, base]{\color{textcolor}\fontsize{16.000000}{19.200000}\selectfont \(\displaystyle 6\)}%
\end{pgfscope}%
\begin{pgfscope}%
\pgfsetbuttcap%
\pgfsetroundjoin%
\definecolor{currentfill}{rgb}{0.000000,0.000000,0.000000}%
\pgfsetfillcolor{currentfill}%
\pgfsetlinewidth{0.803000pt}%
\definecolor{currentstroke}{rgb}{0.000000,0.000000,0.000000}%
\pgfsetstrokecolor{currentstroke}%
\pgfsetdash{}{0pt}%
\pgfsys@defobject{currentmarker}{\pgfqpoint{-0.048611in}{0.000000in}}{\pgfqpoint{0.000000in}{0.000000in}}{%
\pgfpathmoveto{\pgfqpoint{0.000000in}{0.000000in}}%
\pgfpathlineto{\pgfqpoint{-0.048611in}{0.000000in}}%
\pgfusepath{stroke,fill}%
}%
\begin{pgfscope}%
\pgfsys@transformshift{0.594914in}{1.940109in}%
\pgfsys@useobject{currentmarker}{}%
\end{pgfscope}%
\end{pgfscope}%
\begin{pgfscope}%
\definecolor{textcolor}{rgb}{0.000000,0.000000,0.000000}%
\pgfsetstrokecolor{textcolor}%
\pgfsetfillcolor{textcolor}%
\pgftext[x=0.387623in, y=1.862997in, left, base]{\color{textcolor}\fontsize{16.000000}{19.200000}\selectfont \(\displaystyle 9\)}%
\end{pgfscope}%
\begin{pgfscope}%
\pgfsetbuttcap%
\pgfsetroundjoin%
\definecolor{currentfill}{rgb}{0.000000,0.000000,0.000000}%
\pgfsetfillcolor{currentfill}%
\pgfsetlinewidth{0.803000pt}%
\definecolor{currentstroke}{rgb}{0.000000,0.000000,0.000000}%
\pgfsetstrokecolor{currentstroke}%
\pgfsetdash{}{0pt}%
\pgfsys@defobject{currentmarker}{\pgfqpoint{-0.048611in}{0.000000in}}{\pgfqpoint{0.000000in}{0.000000in}}{%
\pgfpathmoveto{\pgfqpoint{0.000000in}{0.000000in}}%
\pgfpathlineto{\pgfqpoint{-0.048611in}{0.000000in}}%
\pgfusepath{stroke,fill}%
}%
\begin{pgfscope}%
\pgfsys@transformshift{0.594914in}{2.350529in}%
\pgfsys@useobject{currentmarker}{}%
\end{pgfscope}%
\end{pgfscope}%
\begin{pgfscope}%
\definecolor{textcolor}{rgb}{0.000000,0.000000,0.000000}%
\pgfsetstrokecolor{textcolor}%
\pgfsetfillcolor{textcolor}%
\pgftext[x=0.277555in, y=2.273418in, left, base]{\color{textcolor}\fontsize{16.000000}{19.200000}\selectfont \(\displaystyle 12\)}%
\end{pgfscope}%
\begin{pgfscope}%
\pgfsetbuttcap%
\pgfsetroundjoin%
\definecolor{currentfill}{rgb}{0.000000,0.000000,0.000000}%
\pgfsetfillcolor{currentfill}%
\pgfsetlinewidth{0.803000pt}%
\definecolor{currentstroke}{rgb}{0.000000,0.000000,0.000000}%
\pgfsetstrokecolor{currentstroke}%
\pgfsetdash{}{0pt}%
\pgfsys@defobject{currentmarker}{\pgfqpoint{-0.048611in}{0.000000in}}{\pgfqpoint{0.000000in}{0.000000in}}{%
\pgfpathmoveto{\pgfqpoint{0.000000in}{0.000000in}}%
\pgfpathlineto{\pgfqpoint{-0.048611in}{0.000000in}}%
\pgfusepath{stroke,fill}%
}%
\begin{pgfscope}%
\pgfsys@transformshift{0.594914in}{2.760949in}%
\pgfsys@useobject{currentmarker}{}%
\end{pgfscope}%
\end{pgfscope}%
\begin{pgfscope}%
\definecolor{textcolor}{rgb}{0.000000,0.000000,0.000000}%
\pgfsetstrokecolor{textcolor}%
\pgfsetfillcolor{textcolor}%
\pgftext[x=0.277555in, y=2.683838in, left, base]{\color{textcolor}\fontsize{16.000000}{19.200000}\selectfont \(\displaystyle 15\)}%
\end{pgfscope}%
\begin{pgfscope}%
\pgfsetbuttcap%
\pgfsetroundjoin%
\definecolor{currentfill}{rgb}{0.000000,0.000000,0.000000}%
\pgfsetfillcolor{currentfill}%
\pgfsetlinewidth{0.803000pt}%
\definecolor{currentstroke}{rgb}{0.000000,0.000000,0.000000}%
\pgfsetstrokecolor{currentstroke}%
\pgfsetdash{}{0pt}%
\pgfsys@defobject{currentmarker}{\pgfqpoint{-0.048611in}{0.000000in}}{\pgfqpoint{0.000000in}{0.000000in}}{%
\pgfpathmoveto{\pgfqpoint{0.000000in}{0.000000in}}%
\pgfpathlineto{\pgfqpoint{-0.048611in}{0.000000in}}%
\pgfusepath{stroke,fill}%
}%
\begin{pgfscope}%
\pgfsys@transformshift{0.594914in}{3.171370in}%
\pgfsys@useobject{currentmarker}{}%
\end{pgfscope}%
\end{pgfscope}%
\begin{pgfscope}%
\definecolor{textcolor}{rgb}{0.000000,0.000000,0.000000}%
\pgfsetstrokecolor{textcolor}%
\pgfsetfillcolor{textcolor}%
\pgftext[x=0.277555in, y=3.094259in, left, base]{\color{textcolor}\fontsize{16.000000}{19.200000}\selectfont \(\displaystyle 18\)}%
\end{pgfscope}%
\begin{pgfscope}%
\pgfsetbuttcap%
\pgfsetroundjoin%
\definecolor{currentfill}{rgb}{0.000000,0.000000,0.000000}%
\pgfsetfillcolor{currentfill}%
\pgfsetlinewidth{0.803000pt}%
\definecolor{currentstroke}{rgb}{0.000000,0.000000,0.000000}%
\pgfsetstrokecolor{currentstroke}%
\pgfsetdash{}{0pt}%
\pgfsys@defobject{currentmarker}{\pgfqpoint{-0.048611in}{0.000000in}}{\pgfqpoint{0.000000in}{0.000000in}}{%
\pgfpathmoveto{\pgfqpoint{0.000000in}{0.000000in}}%
\pgfpathlineto{\pgfqpoint{-0.048611in}{0.000000in}}%
\pgfusepath{stroke,fill}%
}%
\begin{pgfscope}%
\pgfsys@transformshift{0.594914in}{3.581790in}%
\pgfsys@useobject{currentmarker}{}%
\end{pgfscope}%
\end{pgfscope}%
\begin{pgfscope}%
\definecolor{textcolor}{rgb}{0.000000,0.000000,0.000000}%
\pgfsetstrokecolor{textcolor}%
\pgfsetfillcolor{textcolor}%
\pgftext[x=0.277555in, y=3.504679in, left, base]{\color{textcolor}\fontsize{16.000000}{19.200000}\selectfont \(\displaystyle 21\)}%
\end{pgfscope}%
\begin{pgfscope}%
\pgfsetbuttcap%
\pgfsetroundjoin%
\definecolor{currentfill}{rgb}{0.000000,0.000000,0.000000}%
\pgfsetfillcolor{currentfill}%
\pgfsetlinewidth{0.803000pt}%
\definecolor{currentstroke}{rgb}{0.000000,0.000000,0.000000}%
\pgfsetstrokecolor{currentstroke}%
\pgfsetdash{}{0pt}%
\pgfsys@defobject{currentmarker}{\pgfqpoint{-0.048611in}{0.000000in}}{\pgfqpoint{0.000000in}{0.000000in}}{%
\pgfpathmoveto{\pgfqpoint{0.000000in}{0.000000in}}%
\pgfpathlineto{\pgfqpoint{-0.048611in}{0.000000in}}%
\pgfusepath{stroke,fill}%
}%
\begin{pgfscope}%
\pgfsys@transformshift{0.594914in}{3.992210in}%
\pgfsys@useobject{currentmarker}{}%
\end{pgfscope}%
\end{pgfscope}%
\begin{pgfscope}%
\definecolor{textcolor}{rgb}{0.000000,0.000000,0.000000}%
\pgfsetstrokecolor{textcolor}%
\pgfsetfillcolor{textcolor}%
\pgftext[x=0.277555in, y=3.915099in, left, base]{\color{textcolor}\fontsize{16.000000}{19.200000}\selectfont \(\displaystyle 24\)}%
\end{pgfscope}%
\begin{pgfscope}%
\definecolor{textcolor}{rgb}{0.000000,0.000000,0.000000}%
\pgfsetstrokecolor{textcolor}%
\pgfsetfillcolor{textcolor}%
\pgftext[x=0.222000in,y=2.395888in,,bottom,rotate=90.000000]{\color{textcolor}\fontsize{16.000000}{19.200000}\selectfont Real time consumption (minutes)}%
\end{pgfscope}%
\begin{pgfscope}%
\pgfpathrectangle{\pgfqpoint{0.594914in}{0.547888in}}{\pgfqpoint{4.960000in}{3.696000in}}%
\pgfusepath{clip}%
\pgfsetrectcap%
\pgfsetroundjoin%
\pgfsetlinewidth{1.505625pt}%
\definecolor{currentstroke}{rgb}{0.498039,0.498039,0.498039}%
\pgfsetstrokecolor{currentstroke}%
\pgfsetdash{}{0pt}%
\pgfpathmoveto{\pgfqpoint{0.820368in}{0.715973in}}%
\pgfpathlineto{\pgfqpoint{1.085609in}{0.724149in}}%
\pgfpathlineto{\pgfqpoint{1.350850in}{0.735700in}}%
\pgfpathlineto{\pgfqpoint{1.616090in}{0.778404in}}%
\pgfpathlineto{\pgfqpoint{1.881331in}{0.837282in}}%
\pgfpathlineto{\pgfqpoint{2.146572in}{0.951463in}}%
\pgfpathlineto{\pgfqpoint{2.411812in}{1.125046in}}%
\pgfpathlineto{\pgfqpoint{2.677053in}{1.654606in}}%
\pgfpathlineto{\pgfqpoint{2.942294in}{2.219313in}}%
\pgfpathlineto{\pgfqpoint{3.207534in}{3.086269in}}%
\pgfpathlineto{\pgfqpoint{3.472775in}{4.075888in}}%
\pgfusepath{stroke}%
\end{pgfscope}%
\begin{pgfscope}%
\pgfpathrectangle{\pgfqpoint{0.594914in}{0.547888in}}{\pgfqpoint{4.960000in}{3.696000in}}%
\pgfusepath{clip}%
\pgfsetbuttcap%
\pgfsetroundjoin%
\pgfsetlinewidth{1.505625pt}%
\definecolor{currentstroke}{rgb}{1.000000,0.498039,0.054902}%
\pgfsetstrokecolor{currentstroke}%
\pgfsetdash{{5.550000pt}{2.400000pt}}{0.000000pt}%
\pgfpathmoveto{\pgfqpoint{0.820368in}{0.716404in}}%
\pgfpathlineto{\pgfqpoint{1.085609in}{0.725157in}}%
\pgfpathlineto{\pgfqpoint{1.350850in}{0.738377in}}%
\pgfpathlineto{\pgfqpoint{1.616090in}{0.764934in}}%
\pgfpathlineto{\pgfqpoint{1.881331in}{0.799692in}}%
\pgfpathlineto{\pgfqpoint{2.146572in}{0.881452in}}%
\pgfpathlineto{\pgfqpoint{2.411812in}{1.006493in}}%
\pgfpathlineto{\pgfqpoint{2.677053in}{1.140735in}}%
\pgfpathlineto{\pgfqpoint{2.942294in}{1.334114in}}%
\pgfpathlineto{\pgfqpoint{3.207534in}{1.598424in}}%
\pgfpathlineto{\pgfqpoint{3.472775in}{1.900544in}}%
\pgfpathlineto{\pgfqpoint{3.738015in}{2.550762in}}%
\pgfpathlineto{\pgfqpoint{4.003256in}{3.646046in}}%
\pgfusepath{stroke}%
\end{pgfscope}%
\begin{pgfscope}%
\pgfpathrectangle{\pgfqpoint{0.594914in}{0.547888in}}{\pgfqpoint{4.960000in}{3.696000in}}%
\pgfusepath{clip}%
\pgfsetbuttcap%
\pgfsetroundjoin%
\pgfsetlinewidth{1.505625pt}%
\definecolor{currentstroke}{rgb}{0.839216,0.152941,0.156863}%
\pgfsetstrokecolor{currentstroke}%
\pgfsetdash{{1.500000pt}{2.475000pt}}{0.000000pt}%
\pgfpathmoveto{\pgfqpoint{0.820368in}{0.717097in}}%
\pgfpathlineto{\pgfqpoint{1.085609in}{0.729248in}}%
\pgfpathlineto{\pgfqpoint{1.350850in}{0.741503in}}%
\pgfpathlineto{\pgfqpoint{1.616090in}{0.791351in}}%
\pgfpathlineto{\pgfqpoint{1.881331in}{0.844838in}}%
\pgfpathlineto{\pgfqpoint{2.146572in}{0.980384in}}%
\pgfpathlineto{\pgfqpoint{2.411812in}{1.294515in}}%
\pgfpathlineto{\pgfqpoint{2.677053in}{1.668059in}}%
\pgfpathlineto{\pgfqpoint{2.942294in}{2.181012in}}%
\pgfpathlineto{\pgfqpoint{3.207534in}{2.715805in}}%
\pgfusepath{stroke}%
\end{pgfscope}%
\begin{pgfscope}%
\pgfpathrectangle{\pgfqpoint{0.594914in}{0.547888in}}{\pgfqpoint{4.960000in}{3.696000in}}%
\pgfusepath{clip}%
\pgfsetbuttcap%
\pgfsetroundjoin%
\pgfsetlinewidth{1.505625pt}%
\definecolor{currentstroke}{rgb}{0.172549,0.627451,0.172549}%
\pgfsetstrokecolor{currentstroke}%
\pgfsetdash{{9.600000pt}{2.400000pt}{1.500000pt}{2.400000pt}}{0.000000pt}%
\pgfpathmoveto{\pgfqpoint{0.820368in}{0.715888in}}%
\pgfpathlineto{\pgfqpoint{1.085609in}{0.722975in}}%
\pgfpathlineto{\pgfqpoint{1.350850in}{0.734720in}}%
\pgfpathlineto{\pgfqpoint{1.616090in}{0.758741in}}%
\pgfpathlineto{\pgfqpoint{1.881331in}{0.785236in}}%
\pgfpathlineto{\pgfqpoint{2.146572in}{0.850114in}}%
\pgfpathlineto{\pgfqpoint{2.411812in}{0.917774in}}%
\pgfpathlineto{\pgfqpoint{2.677053in}{0.993893in}}%
\pgfpathlineto{\pgfqpoint{2.942294in}{1.112881in}}%
\pgfpathlineto{\pgfqpoint{3.207534in}{1.232453in}}%
\pgfpathlineto{\pgfqpoint{3.472775in}{1.378820in}}%
\pgfpathlineto{\pgfqpoint{3.738015in}{1.542356in}}%
\pgfpathlineto{\pgfqpoint{4.003256in}{1.786278in}}%
\pgfpathlineto{\pgfqpoint{4.268497in}{2.042486in}}%
\pgfpathlineto{\pgfqpoint{4.533737in}{2.323856in}}%
\pgfpathlineto{\pgfqpoint{4.798978in}{2.626899in}}%
\pgfpathlineto{\pgfqpoint{5.064219in}{3.083002in}}%
\pgfpathlineto{\pgfqpoint{5.329459in}{3.910026in}}%
\pgfusepath{stroke}%
\end{pgfscope}%
\begin{pgfscope}%
\pgfsetrectcap%
\pgfsetmiterjoin%
\pgfsetlinewidth{0.803000pt}%
\definecolor{currentstroke}{rgb}{0.000000,0.000000,0.000000}%
\pgfsetstrokecolor{currentstroke}%
\pgfsetdash{}{0pt}%
\pgfpathmoveto{\pgfqpoint{0.594914in}{0.547888in}}%
\pgfpathlineto{\pgfqpoint{0.594914in}{4.243888in}}%
\pgfusepath{stroke}%
\end{pgfscope}%
\begin{pgfscope}%
\pgfsetrectcap%
\pgfsetmiterjoin%
\pgfsetlinewidth{0.803000pt}%
\definecolor{currentstroke}{rgb}{0.000000,0.000000,0.000000}%
\pgfsetstrokecolor{currentstroke}%
\pgfsetdash{}{0pt}%
\pgfpathmoveto{\pgfqpoint{5.554914in}{0.547888in}}%
\pgfpathlineto{\pgfqpoint{5.554914in}{4.243888in}}%
\pgfusepath{stroke}%
\end{pgfscope}%
\begin{pgfscope}%
\pgfsetrectcap%
\pgfsetmiterjoin%
\pgfsetlinewidth{0.803000pt}%
\definecolor{currentstroke}{rgb}{0.000000,0.000000,0.000000}%
\pgfsetstrokecolor{currentstroke}%
\pgfsetdash{}{0pt}%
\pgfpathmoveto{\pgfqpoint{0.594914in}{0.547888in}}%
\pgfpathlineto{\pgfqpoint{5.554914in}{0.547888in}}%
\pgfusepath{stroke}%
\end{pgfscope}%
\begin{pgfscope}%
\pgfsetrectcap%
\pgfsetmiterjoin%
\pgfsetlinewidth{0.803000pt}%
\definecolor{currentstroke}{rgb}{0.000000,0.000000,0.000000}%
\pgfsetstrokecolor{currentstroke}%
\pgfsetdash{}{0pt}%
\pgfpathmoveto{\pgfqpoint{0.594914in}{4.243888in}}%
\pgfpathlineto{\pgfqpoint{5.554914in}{4.243888in}}%
\pgfusepath{stroke}%
\end{pgfscope}%
\begin{pgfscope}%
\pgfsetbuttcap%
\pgfsetmiterjoin%
\definecolor{currentfill}{rgb}{1.000000,1.000000,1.000000}%
\pgfsetfillcolor{currentfill}%
\pgfsetfillopacity{0.800000}%
\pgfsetlinewidth{1.003750pt}%
\definecolor{currentstroke}{rgb}{0.800000,0.800000,0.800000}%
\pgfsetstrokecolor{currentstroke}%
\pgfsetstrokeopacity{0.800000}%
\pgfsetdash{}{0pt}%
\pgfpathmoveto{\pgfqpoint{0.750469in}{2.824112in}}%
\pgfpathlineto{\pgfqpoint{3.201803in}{2.824112in}}%
\pgfpathquadraticcurveto{\pgfqpoint{3.246247in}{2.824112in}}{\pgfqpoint{3.246247in}{2.868556in}}%
\pgfpathlineto{\pgfqpoint{3.246247in}{4.088333in}}%
\pgfpathquadraticcurveto{\pgfqpoint{3.246247in}{4.132777in}}{\pgfqpoint{3.201803in}{4.132777in}}%
\pgfpathlineto{\pgfqpoint{0.750469in}{4.132777in}}%
\pgfpathquadraticcurveto{\pgfqpoint{0.706025in}{4.132777in}}{\pgfqpoint{0.706025in}{4.088333in}}%
\pgfpathlineto{\pgfqpoint{0.706025in}{2.868556in}}%
\pgfpathquadraticcurveto{\pgfqpoint{0.706025in}{2.824112in}}{\pgfqpoint{0.750469in}{2.824112in}}%
\pgfpathclose%
\pgfusepath{stroke,fill}%
\end{pgfscope}%
\begin{pgfscope}%
\pgfsetrectcap%
\pgfsetroundjoin%
\pgfsetlinewidth{1.505625pt}%
\definecolor{currentstroke}{rgb}{0.498039,0.498039,0.498039}%
\pgfsetstrokecolor{currentstroke}%
\pgfsetdash{}{0pt}%
\pgfpathmoveto{\pgfqpoint{0.794914in}{3.966111in}}%
\pgfpathlineto{\pgfqpoint{1.239358in}{3.966111in}}%
\pgfusepath{stroke}%
\end{pgfscope}%
\begin{pgfscope}%
\definecolor{textcolor}{rgb}{0.000000,0.000000,0.000000}%
\pgfsetstrokecolor{textcolor}%
\pgfsetfillcolor{textcolor}%
\pgftext[x=1.417136in,y=3.888333in,left,base]{\color{textcolor}\fontsize{16.000000}{19.200000}\selectfont original encoding}%
\end{pgfscope}%
\begin{pgfscope}%
\pgfsetbuttcap%
\pgfsetroundjoin%
\pgfsetlinewidth{1.505625pt}%
\definecolor{currentstroke}{rgb}{1.000000,0.498039,0.054902}%
\pgfsetstrokecolor{currentstroke}%
\pgfsetdash{{5.550000pt}{2.400000pt}}{0.000000pt}%
\pgfpathmoveto{\pgfqpoint{0.794914in}{3.654111in}}%
\pgfpathlineto{\pgfqpoint{1.239358in}{3.654111in}}%
\pgfusepath{stroke}%
\end{pgfscope}%
\begin{pgfscope}%
\definecolor{textcolor}{rgb}{0.000000,0.000000,0.000000}%
\pgfsetstrokecolor{textcolor}%
\pgfsetfillcolor{textcolor}%
\pgftext[x=1.417136in,y=3.576333in,left,base]{\color{textcolor}\fontsize{16.000000}{19.200000}\selectfont learned at first UIP}%
\end{pgfscope}%
\begin{pgfscope}%
\pgfsetbuttcap%
\pgfsetroundjoin%
\pgfsetlinewidth{1.505625pt}%
\definecolor{currentstroke}{rgb}{0.839216,0.152941,0.156863}%
\pgfsetstrokecolor{currentstroke}%
\pgfsetdash{{1.500000pt}{2.475000pt}}{0.000000pt}%
\pgfpathmoveto{\pgfqpoint{0.794914in}{3.344111in}}%
\pgfpathlineto{\pgfqpoint{1.239358in}{3.344111in}}%
\pgfusepath{stroke}%
\end{pgfscope}%
\begin{pgfscope}%
\definecolor{textcolor}{rgb}{0.000000,0.000000,0.000000}%
\pgfsetstrokecolor{textcolor}%
\pgfsetfillcolor{textcolor}%
\pgftext[x=1.417136in,y=3.266333in,left,base]{\color{textcolor}\fontsize{16.000000}{19.200000}\selectfont learned at last UIP}%
\end{pgfscope}%
\begin{pgfscope}%
\pgfsetbuttcap%
\pgfsetroundjoin%
\pgfsetlinewidth{1.505625pt}%
\definecolor{currentstroke}{rgb}{0.172549,0.627451,0.172549}%
\pgfsetstrokecolor{currentstroke}%
\pgfsetdash{{9.600000pt}{2.400000pt}{1.500000pt}{2.400000pt}}{0.000000pt}%
\pgfpathmoveto{\pgfqpoint{0.794914in}{3.034111in}}%
\pgfpathlineto{\pgfqpoint{1.239358in}{3.034111in}}%
\pgfusepath{stroke}%
\end{pgfscope}%
\begin{pgfscope}%
\definecolor{textcolor}{rgb}{0.000000,0.000000,0.000000}%
\pgfsetstrokecolor{textcolor}%
\pgfsetfillcolor{textcolor}%
\pgftext[x=1.417136in,y=2.956333in,left,base]{\color{textcolor}\fontsize{16.000000}{19.200000}\selectfont reduced constraint}%
\end{pgfscope}%
\end{pgfpicture}%
\makeatother%
\endgroup%

%% file: figures/cactus_3col_crafted_kco_permissive.pgf
%% Creator: Matplotlib, PGF backend
%%
%% To include the figure in your LaTeX document, write
%%   \input{<filename>.pgf}
%%
%% Make sure the required packages are loaded in your preamble
%%   \usepackage{pgf}
%%
%% and, on pdftex
%%   \usepackage[utf8]{inputenc}\DeclareUnicodeCharacter{2212}{-}
%%
%% or, on luatex and xetex
%%   \usepackage{unicode-math}
%%
%% Figures using additional raster images can only be included by \input if
%% they are in the same directory as the main LaTeX file. For loading figures
%% from other directories you can use the `import` package
%%   \usepackage{import}
%%
%% and then include the figures with
%%   \import{<path to file>}{<filename>.pgf}
%%
%% Matplotlib used the following preamble
%%   \usepackage{fontspec}
%%
\begingroup%
\makeatletter%
\begin{pgfpicture}%
\pgfpathrectangle{\pgfpointorigin}{\pgfqpoint{5.664982in}{4.320162in}}%
\pgfusepath{use as bounding box, clip}%
\begin{pgfscope}%
\pgfsetbuttcap%
\pgfsetmiterjoin%
\definecolor{currentfill}{rgb}{1.000000,1.000000,1.000000}%
\pgfsetfillcolor{currentfill}%
\pgfsetlinewidth{0.000000pt}%
\definecolor{currentstroke}{rgb}{1.000000,1.000000,1.000000}%
\pgfsetstrokecolor{currentstroke}%
\pgfsetdash{}{0pt}%
\pgfpathmoveto{\pgfqpoint{0.000000in}{0.000000in}}%
\pgfpathlineto{\pgfqpoint{5.664982in}{0.000000in}}%
\pgfpathlineto{\pgfqpoint{5.664982in}{4.320162in}}%
\pgfpathlineto{\pgfqpoint{0.000000in}{4.320162in}}%
\pgfpathclose%
\pgfusepath{fill}%
\end{pgfscope}%
\begin{pgfscope}%
\pgfsetbuttcap%
\pgfsetmiterjoin%
\definecolor{currentfill}{rgb}{1.000000,1.000000,1.000000}%
\pgfsetfillcolor{currentfill}%
\pgfsetlinewidth{0.000000pt}%
\definecolor{currentstroke}{rgb}{0.000000,0.000000,0.000000}%
\pgfsetstrokecolor{currentstroke}%
\pgfsetstrokeopacity{0.000000}%
\pgfsetdash{}{0pt}%
\pgfpathmoveto{\pgfqpoint{0.704982in}{0.547888in}}%
\pgfpathlineto{\pgfqpoint{5.664982in}{0.547888in}}%
\pgfpathlineto{\pgfqpoint{5.664982in}{4.243888in}}%
\pgfpathlineto{\pgfqpoint{0.704982in}{4.243888in}}%
\pgfpathclose%
\pgfusepath{fill}%
\end{pgfscope}%
\begin{pgfscope}%
\pgfsetbuttcap%
\pgfsetroundjoin%
\definecolor{currentfill}{rgb}{0.000000,0.000000,0.000000}%
\pgfsetfillcolor{currentfill}%
\pgfsetlinewidth{0.803000pt}%
\definecolor{currentstroke}{rgb}{0.000000,0.000000,0.000000}%
\pgfsetstrokecolor{currentstroke}%
\pgfsetdash{}{0pt}%
\pgfsys@defobject{currentmarker}{\pgfqpoint{0.000000in}{-0.048611in}}{\pgfqpoint{0.000000in}{0.000000in}}{%
\pgfpathmoveto{\pgfqpoint{0.000000in}{0.000000in}}%
\pgfpathlineto{\pgfqpoint{0.000000in}{-0.048611in}}%
\pgfusepath{stroke,fill}%
}%
\begin{pgfscope}%
\pgfsys@transformshift{0.907778in}{0.547888in}%
\pgfsys@useobject{currentmarker}{}%
\end{pgfscope}%
\end{pgfscope}%
\begin{pgfscope}%
\definecolor{textcolor}{rgb}{0.000000,0.000000,0.000000}%
\pgfsetstrokecolor{textcolor}%
\pgfsetfillcolor{textcolor}%
\pgftext[x=0.907778in,y=0.450666in,,top]{\color{textcolor}\fontsize{16.000000}{19.200000}\selectfont \(\displaystyle 0\)}%
\end{pgfscope}%
\begin{pgfscope}%
\pgfsetbuttcap%
\pgfsetroundjoin%
\definecolor{currentfill}{rgb}{0.000000,0.000000,0.000000}%
\pgfsetfillcolor{currentfill}%
\pgfsetlinewidth{0.803000pt}%
\definecolor{currentstroke}{rgb}{0.000000,0.000000,0.000000}%
\pgfsetstrokecolor{currentstroke}%
\pgfsetdash{}{0pt}%
\pgfsys@defobject{currentmarker}{\pgfqpoint{0.000000in}{-0.048611in}}{\pgfqpoint{0.000000in}{0.000000in}}{%
\pgfpathmoveto{\pgfqpoint{0.000000in}{0.000000in}}%
\pgfpathlineto{\pgfqpoint{0.000000in}{-0.048611in}}%
\pgfusepath{stroke,fill}%
}%
\begin{pgfscope}%
\pgfsys@transformshift{1.474247in}{0.547888in}%
\pgfsys@useobject{currentmarker}{}%
\end{pgfscope}%
\end{pgfscope}%
\begin{pgfscope}%
\definecolor{textcolor}{rgb}{0.000000,0.000000,0.000000}%
\pgfsetstrokecolor{textcolor}%
\pgfsetfillcolor{textcolor}%
\pgftext[x=1.474247in,y=0.450666in,,top]{\color{textcolor}\fontsize{16.000000}{19.200000}\selectfont \(\displaystyle 25\)}%
\end{pgfscope}%
\begin{pgfscope}%
\pgfsetbuttcap%
\pgfsetroundjoin%
\definecolor{currentfill}{rgb}{0.000000,0.000000,0.000000}%
\pgfsetfillcolor{currentfill}%
\pgfsetlinewidth{0.803000pt}%
\definecolor{currentstroke}{rgb}{0.000000,0.000000,0.000000}%
\pgfsetstrokecolor{currentstroke}%
\pgfsetdash{}{0pt}%
\pgfsys@defobject{currentmarker}{\pgfqpoint{0.000000in}{-0.048611in}}{\pgfqpoint{0.000000in}{0.000000in}}{%
\pgfpathmoveto{\pgfqpoint{0.000000in}{0.000000in}}%
\pgfpathlineto{\pgfqpoint{0.000000in}{-0.048611in}}%
\pgfusepath{stroke,fill}%
}%
\begin{pgfscope}%
\pgfsys@transformshift{2.040715in}{0.547888in}%
\pgfsys@useobject{currentmarker}{}%
\end{pgfscope}%
\end{pgfscope}%
\begin{pgfscope}%
\definecolor{textcolor}{rgb}{0.000000,0.000000,0.000000}%
\pgfsetstrokecolor{textcolor}%
\pgfsetfillcolor{textcolor}%
\pgftext[x=2.040715in,y=0.450666in,,top]{\color{textcolor}\fontsize{16.000000}{19.200000}\selectfont \(\displaystyle 50\)}%
\end{pgfscope}%
\begin{pgfscope}%
\pgfsetbuttcap%
\pgfsetroundjoin%
\definecolor{currentfill}{rgb}{0.000000,0.000000,0.000000}%
\pgfsetfillcolor{currentfill}%
\pgfsetlinewidth{0.803000pt}%
\definecolor{currentstroke}{rgb}{0.000000,0.000000,0.000000}%
\pgfsetstrokecolor{currentstroke}%
\pgfsetdash{}{0pt}%
\pgfsys@defobject{currentmarker}{\pgfqpoint{0.000000in}{-0.048611in}}{\pgfqpoint{0.000000in}{0.000000in}}{%
\pgfpathmoveto{\pgfqpoint{0.000000in}{0.000000in}}%
\pgfpathlineto{\pgfqpoint{0.000000in}{-0.048611in}}%
\pgfusepath{stroke,fill}%
}%
\begin{pgfscope}%
\pgfsys@transformshift{2.607184in}{0.547888in}%
\pgfsys@useobject{currentmarker}{}%
\end{pgfscope}%
\end{pgfscope}%
\begin{pgfscope}%
\definecolor{textcolor}{rgb}{0.000000,0.000000,0.000000}%
\pgfsetstrokecolor{textcolor}%
\pgfsetfillcolor{textcolor}%
\pgftext[x=2.607184in,y=0.450666in,,top]{\color{textcolor}\fontsize{16.000000}{19.200000}\selectfont \(\displaystyle 75\)}%
\end{pgfscope}%
\begin{pgfscope}%
\pgfsetbuttcap%
\pgfsetroundjoin%
\definecolor{currentfill}{rgb}{0.000000,0.000000,0.000000}%
\pgfsetfillcolor{currentfill}%
\pgfsetlinewidth{0.803000pt}%
\definecolor{currentstroke}{rgb}{0.000000,0.000000,0.000000}%
\pgfsetstrokecolor{currentstroke}%
\pgfsetdash{}{0pt}%
\pgfsys@defobject{currentmarker}{\pgfqpoint{0.000000in}{-0.048611in}}{\pgfqpoint{0.000000in}{0.000000in}}{%
\pgfpathmoveto{\pgfqpoint{0.000000in}{0.000000in}}%
\pgfpathlineto{\pgfqpoint{0.000000in}{-0.048611in}}%
\pgfusepath{stroke,fill}%
}%
\begin{pgfscope}%
\pgfsys@transformshift{3.173653in}{0.547888in}%
\pgfsys@useobject{currentmarker}{}%
\end{pgfscope}%
\end{pgfscope}%
\begin{pgfscope}%
\definecolor{textcolor}{rgb}{0.000000,0.000000,0.000000}%
\pgfsetstrokecolor{textcolor}%
\pgfsetfillcolor{textcolor}%
\pgftext[x=3.173653in,y=0.450666in,,top]{\color{textcolor}\fontsize{16.000000}{19.200000}\selectfont \(\displaystyle 100\)}%
\end{pgfscope}%
\begin{pgfscope}%
\pgfsetbuttcap%
\pgfsetroundjoin%
\definecolor{currentfill}{rgb}{0.000000,0.000000,0.000000}%
\pgfsetfillcolor{currentfill}%
\pgfsetlinewidth{0.803000pt}%
\definecolor{currentstroke}{rgb}{0.000000,0.000000,0.000000}%
\pgfsetstrokecolor{currentstroke}%
\pgfsetdash{}{0pt}%
\pgfsys@defobject{currentmarker}{\pgfqpoint{0.000000in}{-0.048611in}}{\pgfqpoint{0.000000in}{0.000000in}}{%
\pgfpathmoveto{\pgfqpoint{0.000000in}{0.000000in}}%
\pgfpathlineto{\pgfqpoint{0.000000in}{-0.048611in}}%
\pgfusepath{stroke,fill}%
}%
\begin{pgfscope}%
\pgfsys@transformshift{3.740121in}{0.547888in}%
\pgfsys@useobject{currentmarker}{}%
\end{pgfscope}%
\end{pgfscope}%
\begin{pgfscope}%
\definecolor{textcolor}{rgb}{0.000000,0.000000,0.000000}%
\pgfsetstrokecolor{textcolor}%
\pgfsetfillcolor{textcolor}%
\pgftext[x=3.740121in,y=0.450666in,,top]{\color{textcolor}\fontsize{16.000000}{19.200000}\selectfont \(\displaystyle 125\)}%
\end{pgfscope}%
\begin{pgfscope}%
\pgfsetbuttcap%
\pgfsetroundjoin%
\definecolor{currentfill}{rgb}{0.000000,0.000000,0.000000}%
\pgfsetfillcolor{currentfill}%
\pgfsetlinewidth{0.803000pt}%
\definecolor{currentstroke}{rgb}{0.000000,0.000000,0.000000}%
\pgfsetstrokecolor{currentstroke}%
\pgfsetdash{}{0pt}%
\pgfsys@defobject{currentmarker}{\pgfqpoint{0.000000in}{-0.048611in}}{\pgfqpoint{0.000000in}{0.000000in}}{%
\pgfpathmoveto{\pgfqpoint{0.000000in}{0.000000in}}%
\pgfpathlineto{\pgfqpoint{0.000000in}{-0.048611in}}%
\pgfusepath{stroke,fill}%
}%
\begin{pgfscope}%
\pgfsys@transformshift{4.306590in}{0.547888in}%
\pgfsys@useobject{currentmarker}{}%
\end{pgfscope}%
\end{pgfscope}%
\begin{pgfscope}%
\definecolor{textcolor}{rgb}{0.000000,0.000000,0.000000}%
\pgfsetstrokecolor{textcolor}%
\pgfsetfillcolor{textcolor}%
\pgftext[x=4.306590in,y=0.450666in,,top]{\color{textcolor}\fontsize{16.000000}{19.200000}\selectfont \(\displaystyle 150\)}%
\end{pgfscope}%
\begin{pgfscope}%
\pgfsetbuttcap%
\pgfsetroundjoin%
\definecolor{currentfill}{rgb}{0.000000,0.000000,0.000000}%
\pgfsetfillcolor{currentfill}%
\pgfsetlinewidth{0.803000pt}%
\definecolor{currentstroke}{rgb}{0.000000,0.000000,0.000000}%
\pgfsetstrokecolor{currentstroke}%
\pgfsetdash{}{0pt}%
\pgfsys@defobject{currentmarker}{\pgfqpoint{0.000000in}{-0.048611in}}{\pgfqpoint{0.000000in}{0.000000in}}{%
\pgfpathmoveto{\pgfqpoint{0.000000in}{0.000000in}}%
\pgfpathlineto{\pgfqpoint{0.000000in}{-0.048611in}}%
\pgfusepath{stroke,fill}%
}%
\begin{pgfscope}%
\pgfsys@transformshift{4.873059in}{0.547888in}%
\pgfsys@useobject{currentmarker}{}%
\end{pgfscope}%
\end{pgfscope}%
\begin{pgfscope}%
\definecolor{textcolor}{rgb}{0.000000,0.000000,0.000000}%
\pgfsetstrokecolor{textcolor}%
\pgfsetfillcolor{textcolor}%
\pgftext[x=4.873059in,y=0.450666in,,top]{\color{textcolor}\fontsize{16.000000}{19.200000}\selectfont \(\displaystyle 175\)}%
\end{pgfscope}%
\begin{pgfscope}%
\pgfsetbuttcap%
\pgfsetroundjoin%
\definecolor{currentfill}{rgb}{0.000000,0.000000,0.000000}%
\pgfsetfillcolor{currentfill}%
\pgfsetlinewidth{0.803000pt}%
\definecolor{currentstroke}{rgb}{0.000000,0.000000,0.000000}%
\pgfsetstrokecolor{currentstroke}%
\pgfsetdash{}{0pt}%
\pgfsys@defobject{currentmarker}{\pgfqpoint{0.000000in}{-0.048611in}}{\pgfqpoint{0.000000in}{0.000000in}}{%
\pgfpathmoveto{\pgfqpoint{0.000000in}{0.000000in}}%
\pgfpathlineto{\pgfqpoint{0.000000in}{-0.048611in}}%
\pgfusepath{stroke,fill}%
}%
\begin{pgfscope}%
\pgfsys@transformshift{5.439528in}{0.547888in}%
\pgfsys@useobject{currentmarker}{}%
\end{pgfscope}%
\end{pgfscope}%
\begin{pgfscope}%
\definecolor{textcolor}{rgb}{0.000000,0.000000,0.000000}%
\pgfsetstrokecolor{textcolor}%
\pgfsetfillcolor{textcolor}%
\pgftext[x=5.439528in,y=0.450666in,,top]{\color{textcolor}\fontsize{16.000000}{19.200000}\selectfont \(\displaystyle 200\)}%
\end{pgfscope}%
\begin{pgfscope}%
\definecolor{textcolor}{rgb}{0.000000,0.000000,0.000000}%
\pgfsetstrokecolor{textcolor}%
\pgfsetfillcolor{textcolor}%
\pgftext[x=3.184982in,y=0.197555in,,top]{\color{textcolor}\fontsize{16.000000}{19.200000}\selectfont Number of instances}%
\end{pgfscope}%
\begin{pgfscope}%
\pgfsetbuttcap%
\pgfsetroundjoin%
\definecolor{currentfill}{rgb}{0.000000,0.000000,0.000000}%
\pgfsetfillcolor{currentfill}%
\pgfsetlinewidth{0.803000pt}%
\definecolor{currentstroke}{rgb}{0.000000,0.000000,0.000000}%
\pgfsetstrokecolor{currentstroke}%
\pgfsetdash{}{0pt}%
\pgfsys@defobject{currentmarker}{\pgfqpoint{-0.048611in}{0.000000in}}{\pgfqpoint{0.000000in}{0.000000in}}{%
\pgfpathmoveto{\pgfqpoint{0.000000in}{0.000000in}}%
\pgfpathlineto{\pgfqpoint{-0.048611in}{0.000000in}}%
\pgfusepath{stroke,fill}%
}%
\begin{pgfscope}%
\pgfsys@transformshift{0.704982in}{0.715610in}%
\pgfsys@useobject{currentmarker}{}%
\end{pgfscope}%
\end{pgfscope}%
\begin{pgfscope}%
\definecolor{textcolor}{rgb}{0.000000,0.000000,0.000000}%
\pgfsetstrokecolor{textcolor}%
\pgfsetfillcolor{textcolor}%
\pgftext[x=0.497692in, y=0.638499in, left, base]{\color{textcolor}\fontsize{16.000000}{19.200000}\selectfont \(\displaystyle 0\)}%
\end{pgfscope}%
\begin{pgfscope}%
\pgfsetbuttcap%
\pgfsetroundjoin%
\definecolor{currentfill}{rgb}{0.000000,0.000000,0.000000}%
\pgfsetfillcolor{currentfill}%
\pgfsetlinewidth{0.803000pt}%
\definecolor{currentstroke}{rgb}{0.000000,0.000000,0.000000}%
\pgfsetstrokecolor{currentstroke}%
\pgfsetdash{}{0pt}%
\pgfsys@defobject{currentmarker}{\pgfqpoint{-0.048611in}{0.000000in}}{\pgfqpoint{0.000000in}{0.000000in}}{%
\pgfpathmoveto{\pgfqpoint{0.000000in}{0.000000in}}%
\pgfpathlineto{\pgfqpoint{-0.048611in}{0.000000in}}%
\pgfusepath{stroke,fill}%
}%
\begin{pgfscope}%
\pgfsys@transformshift{0.704982in}{1.156540in}%
\pgfsys@useobject{currentmarker}{}%
\end{pgfscope}%
\end{pgfscope}%
\begin{pgfscope}%
\definecolor{textcolor}{rgb}{0.000000,0.000000,0.000000}%
\pgfsetstrokecolor{textcolor}%
\pgfsetfillcolor{textcolor}%
\pgftext[x=0.387623in, y=1.079429in, left, base]{\color{textcolor}\fontsize{16.000000}{19.200000}\selectfont \(\displaystyle 50\)}%
\end{pgfscope}%
\begin{pgfscope}%
\pgfsetbuttcap%
\pgfsetroundjoin%
\definecolor{currentfill}{rgb}{0.000000,0.000000,0.000000}%
\pgfsetfillcolor{currentfill}%
\pgfsetlinewidth{0.803000pt}%
\definecolor{currentstroke}{rgb}{0.000000,0.000000,0.000000}%
\pgfsetstrokecolor{currentstroke}%
\pgfsetdash{}{0pt}%
\pgfsys@defobject{currentmarker}{\pgfqpoint{-0.048611in}{0.000000in}}{\pgfqpoint{0.000000in}{0.000000in}}{%
\pgfpathmoveto{\pgfqpoint{0.000000in}{0.000000in}}%
\pgfpathlineto{\pgfqpoint{-0.048611in}{0.000000in}}%
\pgfusepath{stroke,fill}%
}%
\begin{pgfscope}%
\pgfsys@transformshift{0.704982in}{1.597470in}%
\pgfsys@useobject{currentmarker}{}%
\end{pgfscope}%
\end{pgfscope}%
\begin{pgfscope}%
\definecolor{textcolor}{rgb}{0.000000,0.000000,0.000000}%
\pgfsetstrokecolor{textcolor}%
\pgfsetfillcolor{textcolor}%
\pgftext[x=0.277555in, y=1.520359in, left, base]{\color{textcolor}\fontsize{16.000000}{19.200000}\selectfont \(\displaystyle 100\)}%
\end{pgfscope}%
\begin{pgfscope}%
\pgfsetbuttcap%
\pgfsetroundjoin%
\definecolor{currentfill}{rgb}{0.000000,0.000000,0.000000}%
\pgfsetfillcolor{currentfill}%
\pgfsetlinewidth{0.803000pt}%
\definecolor{currentstroke}{rgb}{0.000000,0.000000,0.000000}%
\pgfsetstrokecolor{currentstroke}%
\pgfsetdash{}{0pt}%
\pgfsys@defobject{currentmarker}{\pgfqpoint{-0.048611in}{0.000000in}}{\pgfqpoint{0.000000in}{0.000000in}}{%
\pgfpathmoveto{\pgfqpoint{0.000000in}{0.000000in}}%
\pgfpathlineto{\pgfqpoint{-0.048611in}{0.000000in}}%
\pgfusepath{stroke,fill}%
}%
\begin{pgfscope}%
\pgfsys@transformshift{0.704982in}{2.038400in}%
\pgfsys@useobject{currentmarker}{}%
\end{pgfscope}%
\end{pgfscope}%
\begin{pgfscope}%
\definecolor{textcolor}{rgb}{0.000000,0.000000,0.000000}%
\pgfsetstrokecolor{textcolor}%
\pgfsetfillcolor{textcolor}%
\pgftext[x=0.277555in, y=1.961289in, left, base]{\color{textcolor}\fontsize{16.000000}{19.200000}\selectfont \(\displaystyle 150\)}%
\end{pgfscope}%
\begin{pgfscope}%
\pgfsetbuttcap%
\pgfsetroundjoin%
\definecolor{currentfill}{rgb}{0.000000,0.000000,0.000000}%
\pgfsetfillcolor{currentfill}%
\pgfsetlinewidth{0.803000pt}%
\definecolor{currentstroke}{rgb}{0.000000,0.000000,0.000000}%
\pgfsetstrokecolor{currentstroke}%
\pgfsetdash{}{0pt}%
\pgfsys@defobject{currentmarker}{\pgfqpoint{-0.048611in}{0.000000in}}{\pgfqpoint{0.000000in}{0.000000in}}{%
\pgfpathmoveto{\pgfqpoint{0.000000in}{0.000000in}}%
\pgfpathlineto{\pgfqpoint{-0.048611in}{0.000000in}}%
\pgfusepath{stroke,fill}%
}%
\begin{pgfscope}%
\pgfsys@transformshift{0.704982in}{2.479331in}%
\pgfsys@useobject{currentmarker}{}%
\end{pgfscope}%
\end{pgfscope}%
\begin{pgfscope}%
\definecolor{textcolor}{rgb}{0.000000,0.000000,0.000000}%
\pgfsetstrokecolor{textcolor}%
\pgfsetfillcolor{textcolor}%
\pgftext[x=0.277555in, y=2.402219in, left, base]{\color{textcolor}\fontsize{16.000000}{19.200000}\selectfont \(\displaystyle 200\)}%
\end{pgfscope}%
\begin{pgfscope}%
\pgfsetbuttcap%
\pgfsetroundjoin%
\definecolor{currentfill}{rgb}{0.000000,0.000000,0.000000}%
\pgfsetfillcolor{currentfill}%
\pgfsetlinewidth{0.803000pt}%
\definecolor{currentstroke}{rgb}{0.000000,0.000000,0.000000}%
\pgfsetstrokecolor{currentstroke}%
\pgfsetdash{}{0pt}%
\pgfsys@defobject{currentmarker}{\pgfqpoint{-0.048611in}{0.000000in}}{\pgfqpoint{0.000000in}{0.000000in}}{%
\pgfpathmoveto{\pgfqpoint{0.000000in}{0.000000in}}%
\pgfpathlineto{\pgfqpoint{-0.048611in}{0.000000in}}%
\pgfusepath{stroke,fill}%
}%
\begin{pgfscope}%
\pgfsys@transformshift{0.704982in}{2.920261in}%
\pgfsys@useobject{currentmarker}{}%
\end{pgfscope}%
\end{pgfscope}%
\begin{pgfscope}%
\definecolor{textcolor}{rgb}{0.000000,0.000000,0.000000}%
\pgfsetstrokecolor{textcolor}%
\pgfsetfillcolor{textcolor}%
\pgftext[x=0.277555in, y=2.843149in, left, base]{\color{textcolor}\fontsize{16.000000}{19.200000}\selectfont \(\displaystyle 250\)}%
\end{pgfscope}%
\begin{pgfscope}%
\pgfsetbuttcap%
\pgfsetroundjoin%
\definecolor{currentfill}{rgb}{0.000000,0.000000,0.000000}%
\pgfsetfillcolor{currentfill}%
\pgfsetlinewidth{0.803000pt}%
\definecolor{currentstroke}{rgb}{0.000000,0.000000,0.000000}%
\pgfsetstrokecolor{currentstroke}%
\pgfsetdash{}{0pt}%
\pgfsys@defobject{currentmarker}{\pgfqpoint{-0.048611in}{0.000000in}}{\pgfqpoint{0.000000in}{0.000000in}}{%
\pgfpathmoveto{\pgfqpoint{0.000000in}{0.000000in}}%
\pgfpathlineto{\pgfqpoint{-0.048611in}{0.000000in}}%
\pgfusepath{stroke,fill}%
}%
\begin{pgfscope}%
\pgfsys@transformshift{0.704982in}{3.361191in}%
\pgfsys@useobject{currentmarker}{}%
\end{pgfscope}%
\end{pgfscope}%
\begin{pgfscope}%
\definecolor{textcolor}{rgb}{0.000000,0.000000,0.000000}%
\pgfsetstrokecolor{textcolor}%
\pgfsetfillcolor{textcolor}%
\pgftext[x=0.277555in, y=3.284080in, left, base]{\color{textcolor}\fontsize{16.000000}{19.200000}\selectfont \(\displaystyle 300\)}%
\end{pgfscope}%
\begin{pgfscope}%
\pgfsetbuttcap%
\pgfsetroundjoin%
\definecolor{currentfill}{rgb}{0.000000,0.000000,0.000000}%
\pgfsetfillcolor{currentfill}%
\pgfsetlinewidth{0.803000pt}%
\definecolor{currentstroke}{rgb}{0.000000,0.000000,0.000000}%
\pgfsetstrokecolor{currentstroke}%
\pgfsetdash{}{0pt}%
\pgfsys@defobject{currentmarker}{\pgfqpoint{-0.048611in}{0.000000in}}{\pgfqpoint{0.000000in}{0.000000in}}{%
\pgfpathmoveto{\pgfqpoint{0.000000in}{0.000000in}}%
\pgfpathlineto{\pgfqpoint{-0.048611in}{0.000000in}}%
\pgfusepath{stroke,fill}%
}%
\begin{pgfscope}%
\pgfsys@transformshift{0.704982in}{3.802121in}%
\pgfsys@useobject{currentmarker}{}%
\end{pgfscope}%
\end{pgfscope}%
\begin{pgfscope}%
\definecolor{textcolor}{rgb}{0.000000,0.000000,0.000000}%
\pgfsetstrokecolor{textcolor}%
\pgfsetfillcolor{textcolor}%
\pgftext[x=0.277555in, y=3.725010in, left, base]{\color{textcolor}\fontsize{16.000000}{19.200000}\selectfont \(\displaystyle 350\)}%
\end{pgfscope}%
\begin{pgfscope}%
\pgfsetbuttcap%
\pgfsetroundjoin%
\definecolor{currentfill}{rgb}{0.000000,0.000000,0.000000}%
\pgfsetfillcolor{currentfill}%
\pgfsetlinewidth{0.803000pt}%
\definecolor{currentstroke}{rgb}{0.000000,0.000000,0.000000}%
\pgfsetstrokecolor{currentstroke}%
\pgfsetdash{}{0pt}%
\pgfsys@defobject{currentmarker}{\pgfqpoint{-0.048611in}{0.000000in}}{\pgfqpoint{0.000000in}{0.000000in}}{%
\pgfpathmoveto{\pgfqpoint{0.000000in}{0.000000in}}%
\pgfpathlineto{\pgfqpoint{-0.048611in}{0.000000in}}%
\pgfusepath{stroke,fill}%
}%
\begin{pgfscope}%
\pgfsys@transformshift{0.704982in}{4.243051in}%
\pgfsys@useobject{currentmarker}{}%
\end{pgfscope}%
\end{pgfscope}%
\begin{pgfscope}%
\definecolor{textcolor}{rgb}{0.000000,0.000000,0.000000}%
\pgfsetstrokecolor{textcolor}%
\pgfsetfillcolor{textcolor}%
\pgftext[x=0.277555in, y=4.165940in, left, base]{\color{textcolor}\fontsize{16.000000}{19.200000}\selectfont \(\displaystyle 400\)}%
\end{pgfscope}%
\begin{pgfscope}%
\definecolor{textcolor}{rgb}{0.000000,0.000000,0.000000}%
\pgfsetstrokecolor{textcolor}%
\pgfsetfillcolor{textcolor}%
\pgftext[x=0.222000in,y=2.395888in,,bottom,rotate=90.000000]{\color{textcolor}\fontsize{16.000000}{19.200000}\selectfont Real time consumption (minutes)}%
\end{pgfscope}%
\begin{pgfscope}%
\pgfpathrectangle{\pgfqpoint{0.704982in}{0.547888in}}{\pgfqpoint{4.960000in}{3.696000in}}%
\pgfusepath{clip}%
\pgfsetrectcap%
\pgfsetroundjoin%
\pgfsetlinewidth{1.505625pt}%
\definecolor{currentstroke}{rgb}{0.498039,0.498039,0.498039}%
\pgfsetstrokecolor{currentstroke}%
\pgfsetdash{}{0pt}%
\pgfpathmoveto{\pgfqpoint{0.930437in}{0.715891in}}%
\pgfpathlineto{\pgfqpoint{0.953095in}{0.716314in}}%
\pgfpathlineto{\pgfqpoint{0.975754in}{0.716811in}}%
\pgfpathlineto{\pgfqpoint{0.998413in}{0.717444in}}%
\pgfpathlineto{\pgfqpoint{1.021072in}{0.718418in}}%
\pgfpathlineto{\pgfqpoint{1.043730in}{0.719585in}}%
\pgfpathlineto{\pgfqpoint{1.066389in}{0.720895in}}%
\pgfpathlineto{\pgfqpoint{1.089048in}{0.722932in}}%
\pgfpathlineto{\pgfqpoint{1.111707in}{0.725381in}}%
\pgfpathlineto{\pgfqpoint{1.134365in}{0.727842in}}%
\pgfpathlineto{\pgfqpoint{1.157024in}{0.731203in}}%
\pgfpathlineto{\pgfqpoint{1.179683in}{0.735566in}}%
\pgfpathlineto{\pgfqpoint{1.202342in}{0.739935in}}%
\pgfpathlineto{\pgfqpoint{1.225000in}{0.745868in}}%
\pgfpathlineto{\pgfqpoint{1.247659in}{0.752605in}}%
\pgfpathlineto{\pgfqpoint{1.270318in}{0.759462in}}%
\pgfpathlineto{\pgfqpoint{1.292977in}{0.768317in}}%
\pgfpathlineto{\pgfqpoint{1.315635in}{0.777311in}}%
\pgfpathlineto{\pgfqpoint{1.338294in}{0.787226in}}%
\pgfpathlineto{\pgfqpoint{1.360953in}{0.798871in}}%
\pgfpathlineto{\pgfqpoint{1.383612in}{0.813757in}}%
\pgfpathlineto{\pgfqpoint{1.406270in}{0.828833in}}%
\pgfpathlineto{\pgfqpoint{1.428929in}{0.844248in}}%
\pgfpathlineto{\pgfqpoint{1.451588in}{0.861733in}}%
\pgfpathlineto{\pgfqpoint{1.474247in}{0.881528in}}%
\pgfpathlineto{\pgfqpoint{1.496905in}{0.902844in}}%
\pgfpathlineto{\pgfqpoint{1.519564in}{0.926590in}}%
\pgfpathlineto{\pgfqpoint{1.542223in}{0.952338in}}%
\pgfpathlineto{\pgfqpoint{1.564882in}{0.978721in}}%
\pgfpathlineto{\pgfqpoint{1.587540in}{1.009997in}}%
\pgfpathlineto{\pgfqpoint{1.610199in}{1.042737in}}%
\pgfpathlineto{\pgfqpoint{1.632858in}{1.078014in}}%
\pgfpathlineto{\pgfqpoint{1.655517in}{1.114067in}}%
\pgfpathlineto{\pgfqpoint{1.678175in}{1.151262in}}%
\pgfpathlineto{\pgfqpoint{1.700834in}{1.194971in}}%
\pgfpathlineto{\pgfqpoint{1.723493in}{1.243149in}}%
\pgfpathlineto{\pgfqpoint{1.746152in}{1.297923in}}%
\pgfpathlineto{\pgfqpoint{1.768810in}{1.352712in}}%
\pgfpathlineto{\pgfqpoint{1.791469in}{1.408967in}}%
\pgfpathlineto{\pgfqpoint{1.814128in}{1.467201in}}%
\pgfpathlineto{\pgfqpoint{1.836787in}{1.525847in}}%
\pgfpathlineto{\pgfqpoint{1.859445in}{1.586773in}}%
\pgfpathlineto{\pgfqpoint{1.882104in}{1.647798in}}%
\pgfpathlineto{\pgfqpoint{1.904763in}{1.714084in}}%
\pgfpathlineto{\pgfqpoint{1.927422in}{1.791232in}}%
\pgfpathlineto{\pgfqpoint{1.950080in}{1.873812in}}%
\pgfpathlineto{\pgfqpoint{1.972739in}{1.959071in}}%
\pgfpathlineto{\pgfqpoint{1.995398in}{2.046636in}}%
\pgfusepath{stroke}%
\end{pgfscope}%
\begin{pgfscope}%
\pgfpathrectangle{\pgfqpoint{0.704982in}{0.547888in}}{\pgfqpoint{4.960000in}{3.696000in}}%
\pgfusepath{clip}%
\pgfsetbuttcap%
\pgfsetroundjoin%
\pgfsetlinewidth{1.505625pt}%
\definecolor{currentstroke}{rgb}{1.000000,0.498039,0.054902}%
\pgfsetstrokecolor{currentstroke}%
\pgfsetdash{{5.550000pt}{2.400000pt}}{0.000000pt}%
\pgfpathmoveto{\pgfqpoint{0.930437in}{0.716062in}}%
\pgfpathlineto{\pgfqpoint{0.998413in}{0.718480in}}%
\pgfpathlineto{\pgfqpoint{1.066389in}{0.722119in}}%
\pgfpathlineto{\pgfqpoint{1.157024in}{0.729032in}}%
\pgfpathlineto{\pgfqpoint{1.202342in}{0.732933in}}%
\pgfpathlineto{\pgfqpoint{1.270318in}{0.740978in}}%
\pgfpathlineto{\pgfqpoint{1.338294in}{0.750568in}}%
\pgfpathlineto{\pgfqpoint{1.428929in}{0.765181in}}%
\pgfpathlineto{\pgfqpoint{1.519564in}{0.783125in}}%
\pgfpathlineto{\pgfqpoint{1.610199in}{0.803762in}}%
\pgfpathlineto{\pgfqpoint{1.746152in}{0.837074in}}%
\pgfpathlineto{\pgfqpoint{1.791469in}{0.849244in}}%
\pgfpathlineto{\pgfqpoint{1.859445in}{0.869199in}}%
\pgfpathlineto{\pgfqpoint{1.904763in}{0.883113in}}%
\pgfpathlineto{\pgfqpoint{1.995398in}{0.913384in}}%
\pgfpathlineto{\pgfqpoint{2.108692in}{0.952828in}}%
\pgfpathlineto{\pgfqpoint{2.199327in}{0.985627in}}%
\pgfpathlineto{\pgfqpoint{2.289962in}{1.019842in}}%
\pgfpathlineto{\pgfqpoint{2.357938in}{1.046622in}}%
\pgfpathlineto{\pgfqpoint{2.425914in}{1.075553in}}%
\pgfpathlineto{\pgfqpoint{2.516549in}{1.116074in}}%
\pgfpathlineto{\pgfqpoint{2.629843in}{1.168097in}}%
\pgfpathlineto{\pgfqpoint{2.675160in}{1.189413in}}%
\pgfpathlineto{\pgfqpoint{2.743136in}{1.224675in}}%
\pgfpathlineto{\pgfqpoint{2.811113in}{1.262123in}}%
\pgfpathlineto{\pgfqpoint{2.901748in}{1.314025in}}%
\pgfpathlineto{\pgfqpoint{3.037700in}{1.393739in}}%
\pgfpathlineto{\pgfqpoint{3.150994in}{1.461616in}}%
\pgfpathlineto{\pgfqpoint{3.264288in}{1.533065in}}%
\pgfpathlineto{\pgfqpoint{3.354923in}{1.592051in}}%
\pgfpathlineto{\pgfqpoint{3.445558in}{1.653415in}}%
\pgfpathlineto{\pgfqpoint{3.558851in}{1.732659in}}%
\pgfpathlineto{\pgfqpoint{3.649486in}{1.797677in}}%
\pgfpathlineto{\pgfqpoint{3.762780in}{1.880073in}}%
\pgfpathlineto{\pgfqpoint{3.853415in}{1.948179in}}%
\pgfpathlineto{\pgfqpoint{3.921391in}{2.000272in}}%
\pgfpathlineto{\pgfqpoint{4.034685in}{2.089634in}}%
\pgfpathlineto{\pgfqpoint{4.080003in}{2.127918in}}%
\pgfpathlineto{\pgfqpoint{4.125320in}{2.168014in}}%
\pgfpathlineto{\pgfqpoint{4.193296in}{2.230990in}}%
\pgfpathlineto{\pgfqpoint{4.283931in}{2.317639in}}%
\pgfpathlineto{\pgfqpoint{4.306590in}{2.339782in}}%
\pgfpathlineto{\pgfqpoint{4.351908in}{2.386980in}}%
\pgfpathlineto{\pgfqpoint{4.397225in}{2.438893in}}%
\pgfpathlineto{\pgfqpoint{4.487860in}{2.545147in}}%
\pgfpathlineto{\pgfqpoint{4.533178in}{2.599653in}}%
\pgfpathlineto{\pgfqpoint{4.601154in}{2.683608in}}%
\pgfpathlineto{\pgfqpoint{4.669130in}{2.774275in}}%
\pgfpathlineto{\pgfqpoint{4.714448in}{2.836627in}}%
\pgfpathlineto{\pgfqpoint{4.759765in}{2.902398in}}%
\pgfpathlineto{\pgfqpoint{4.827741in}{3.003087in}}%
\pgfpathlineto{\pgfqpoint{4.895718in}{3.107929in}}%
\pgfpathlineto{\pgfqpoint{4.963694in}{3.214820in}}%
\pgfpathlineto{\pgfqpoint{5.031670in}{3.323887in}}%
\pgfpathlineto{\pgfqpoint{5.076988in}{3.401430in}}%
\pgfpathlineto{\pgfqpoint{5.122305in}{3.480801in}}%
\pgfpathlineto{\pgfqpoint{5.190281in}{3.606994in}}%
\pgfpathlineto{\pgfqpoint{5.235599in}{3.692758in}}%
\pgfpathlineto{\pgfqpoint{5.258258in}{3.738007in}}%
\pgfpathlineto{\pgfqpoint{5.303575in}{3.833876in}}%
\pgfpathlineto{\pgfqpoint{5.348893in}{3.937234in}}%
\pgfpathlineto{\pgfqpoint{5.371551in}{3.996727in}}%
\pgfpathlineto{\pgfqpoint{5.394210in}{4.075888in}}%
\pgfpathlineto{\pgfqpoint{5.394210in}{4.075888in}}%
\pgfusepath{stroke}%
\end{pgfscope}%
\begin{pgfscope}%
\pgfpathrectangle{\pgfqpoint{0.704982in}{0.547888in}}{\pgfqpoint{4.960000in}{3.696000in}}%
\pgfusepath{clip}%
\pgfsetbuttcap%
\pgfsetroundjoin%
\pgfsetlinewidth{1.505625pt}%
\definecolor{currentstroke}{rgb}{0.839216,0.152941,0.156863}%
\pgfsetstrokecolor{currentstroke}%
\pgfsetdash{{1.500000pt}{2.475000pt}}{0.000000pt}%
\pgfpathmoveto{\pgfqpoint{0.930437in}{0.717085in}}%
\pgfpathlineto{\pgfqpoint{0.953095in}{0.720404in}}%
\pgfpathlineto{\pgfqpoint{0.975754in}{0.725884in}}%
\pgfpathlineto{\pgfqpoint{0.998413in}{0.731685in}}%
\pgfpathlineto{\pgfqpoint{1.021072in}{0.740504in}}%
\pgfpathlineto{\pgfqpoint{1.043730in}{0.754211in}}%
\pgfpathlineto{\pgfqpoint{1.066389in}{0.771985in}}%
\pgfpathlineto{\pgfqpoint{1.089048in}{0.792608in}}%
\pgfpathlineto{\pgfqpoint{1.111707in}{0.828965in}}%
\pgfpathlineto{\pgfqpoint{1.134365in}{0.865575in}}%
\pgfpathlineto{\pgfqpoint{1.157024in}{0.911089in}}%
\pgfpathlineto{\pgfqpoint{1.179683in}{0.957810in}}%
\pgfpathlineto{\pgfqpoint{1.202342in}{1.007136in}}%
\pgfpathlineto{\pgfqpoint{1.225000in}{1.058856in}}%
\pgfpathlineto{\pgfqpoint{1.247659in}{1.111259in}}%
\pgfpathlineto{\pgfqpoint{1.270318in}{1.181642in}}%
\pgfpathlineto{\pgfqpoint{1.292977in}{1.253918in}}%
\pgfusepath{stroke}%
\end{pgfscope}%
\begin{pgfscope}%
\pgfpathrectangle{\pgfqpoint{0.704982in}{0.547888in}}{\pgfqpoint{4.960000in}{3.696000in}}%
\pgfusepath{clip}%
\pgfsetbuttcap%
\pgfsetroundjoin%
\pgfsetlinewidth{1.505625pt}%
\definecolor{currentstroke}{rgb}{0.172549,0.627451,0.172549}%
\pgfsetstrokecolor{currentstroke}%
\pgfsetdash{{9.600000pt}{2.400000pt}{1.500000pt}{2.400000pt}}{0.000000pt}%
\pgfpathmoveto{\pgfqpoint{0.930437in}{0.715888in}}%
\pgfpathlineto{\pgfqpoint{1.066389in}{0.718355in}}%
\pgfpathlineto{\pgfqpoint{1.202342in}{0.721850in}}%
\pgfpathlineto{\pgfqpoint{1.338294in}{0.726443in}}%
\pgfpathlineto{\pgfqpoint{1.474247in}{0.732555in}}%
\pgfpathlineto{\pgfqpoint{1.587540in}{0.738540in}}%
\pgfpathlineto{\pgfqpoint{1.768810in}{0.749580in}}%
\pgfpathlineto{\pgfqpoint{1.927422in}{0.761142in}}%
\pgfpathlineto{\pgfqpoint{2.063374in}{0.772620in}}%
\pgfpathlineto{\pgfqpoint{2.221985in}{0.787165in}}%
\pgfpathlineto{\pgfqpoint{2.357938in}{0.801021in}}%
\pgfpathlineto{\pgfqpoint{2.539208in}{0.821190in}}%
\pgfpathlineto{\pgfqpoint{2.720478in}{0.842559in}}%
\pgfpathlineto{\pgfqpoint{2.856430in}{0.860373in}}%
\pgfpathlineto{\pgfqpoint{3.015041in}{0.883086in}}%
\pgfpathlineto{\pgfqpoint{3.105676in}{0.896979in}}%
\pgfpathlineto{\pgfqpoint{3.241629in}{0.919289in}}%
\pgfpathlineto{\pgfqpoint{3.468216in}{0.958767in}}%
\pgfpathlineto{\pgfqpoint{3.626828in}{0.987589in}}%
\pgfpathlineto{\pgfqpoint{3.808098in}{1.021677in}}%
\pgfpathlineto{\pgfqpoint{3.921391in}{1.043742in}}%
\pgfpathlineto{\pgfqpoint{4.080003in}{1.075884in}}%
\pgfpathlineto{\pgfqpoint{4.261273in}{1.114198in}}%
\pgfpathlineto{\pgfqpoint{4.397225in}{1.144109in}}%
\pgfpathlineto{\pgfqpoint{4.510519in}{1.170081in}}%
\pgfpathlineto{\pgfqpoint{4.669130in}{1.207807in}}%
\pgfpathlineto{\pgfqpoint{4.737106in}{1.224869in}}%
\pgfpathlineto{\pgfqpoint{4.850400in}{1.255194in}}%
\pgfpathlineto{\pgfqpoint{4.963694in}{1.287168in}}%
\pgfpathlineto{\pgfqpoint{5.099646in}{1.328473in}}%
\pgfpathlineto{\pgfqpoint{5.235599in}{1.371747in}}%
\pgfpathlineto{\pgfqpoint{5.348893in}{1.410257in}}%
\pgfpathlineto{\pgfqpoint{5.416869in}{1.436261in}}%
\pgfpathlineto{\pgfqpoint{5.439528in}{1.445332in}}%
\pgfpathlineto{\pgfqpoint{5.439528in}{1.445332in}}%
\pgfusepath{stroke}%
\end{pgfscope}%
\begin{pgfscope}%
\pgfsetrectcap%
\pgfsetmiterjoin%
\pgfsetlinewidth{0.803000pt}%
\definecolor{currentstroke}{rgb}{0.000000,0.000000,0.000000}%
\pgfsetstrokecolor{currentstroke}%
\pgfsetdash{}{0pt}%
\pgfpathmoveto{\pgfqpoint{0.704982in}{0.547888in}}%
\pgfpathlineto{\pgfqpoint{0.704982in}{4.243888in}}%
\pgfusepath{stroke}%
\end{pgfscope}%
\begin{pgfscope}%
\pgfsetrectcap%
\pgfsetmiterjoin%
\pgfsetlinewidth{0.803000pt}%
\definecolor{currentstroke}{rgb}{0.000000,0.000000,0.000000}%
\pgfsetstrokecolor{currentstroke}%
\pgfsetdash{}{0pt}%
\pgfpathmoveto{\pgfqpoint{5.664982in}{0.547888in}}%
\pgfpathlineto{\pgfqpoint{5.664982in}{4.243888in}}%
\pgfusepath{stroke}%
\end{pgfscope}%
\begin{pgfscope}%
\pgfsetrectcap%
\pgfsetmiterjoin%
\pgfsetlinewidth{0.803000pt}%
\definecolor{currentstroke}{rgb}{0.000000,0.000000,0.000000}%
\pgfsetstrokecolor{currentstroke}%
\pgfsetdash{}{0pt}%
\pgfpathmoveto{\pgfqpoint{0.704982in}{0.547888in}}%
\pgfpathlineto{\pgfqpoint{5.664982in}{0.547888in}}%
\pgfusepath{stroke}%
\end{pgfscope}%
\begin{pgfscope}%
\pgfsetrectcap%
\pgfsetmiterjoin%
\pgfsetlinewidth{0.803000pt}%
\definecolor{currentstroke}{rgb}{0.000000,0.000000,0.000000}%
\pgfsetstrokecolor{currentstroke}%
\pgfsetdash{}{0pt}%
\pgfpathmoveto{\pgfqpoint{0.704982in}{4.243888in}}%
\pgfpathlineto{\pgfqpoint{5.664982in}{4.243888in}}%
\pgfusepath{stroke}%
\end{pgfscope}%
\begin{pgfscope}%
\pgfsetbuttcap%
\pgfsetmiterjoin%
\definecolor{currentfill}{rgb}{1.000000,1.000000,1.000000}%
\pgfsetfillcolor{currentfill}%
\pgfsetfillopacity{0.800000}%
\pgfsetlinewidth{1.003750pt}%
\definecolor{currentstroke}{rgb}{0.800000,0.800000,0.800000}%
\pgfsetstrokecolor{currentstroke}%
\pgfsetstrokeopacity{0.800000}%
\pgfsetdash{}{0pt}%
\pgfpathmoveto{\pgfqpoint{0.860538in}{2.824112in}}%
\pgfpathlineto{\pgfqpoint{3.311871in}{2.824112in}}%
\pgfpathquadraticcurveto{\pgfqpoint{3.356315in}{2.824112in}}{\pgfqpoint{3.356315in}{2.868556in}}%
\pgfpathlineto{\pgfqpoint{3.356315in}{4.088333in}}%
\pgfpathquadraticcurveto{\pgfqpoint{3.356315in}{4.132777in}}{\pgfqpoint{3.311871in}{4.132777in}}%
\pgfpathlineto{\pgfqpoint{0.860538in}{4.132777in}}%
\pgfpathquadraticcurveto{\pgfqpoint{0.816093in}{4.132777in}}{\pgfqpoint{0.816093in}{4.088333in}}%
\pgfpathlineto{\pgfqpoint{0.816093in}{2.868556in}}%
\pgfpathquadraticcurveto{\pgfqpoint{0.816093in}{2.824112in}}{\pgfqpoint{0.860538in}{2.824112in}}%
\pgfpathclose%
\pgfusepath{stroke,fill}%
\end{pgfscope}%
\begin{pgfscope}%
\pgfsetrectcap%
\pgfsetroundjoin%
\pgfsetlinewidth{1.505625pt}%
\definecolor{currentstroke}{rgb}{0.498039,0.498039,0.498039}%
\pgfsetstrokecolor{currentstroke}%
\pgfsetdash{}{0pt}%
\pgfpathmoveto{\pgfqpoint{0.904982in}{3.966111in}}%
\pgfpathlineto{\pgfqpoint{1.349427in}{3.966111in}}%
\pgfusepath{stroke}%
\end{pgfscope}%
\begin{pgfscope}%
\definecolor{textcolor}{rgb}{0.000000,0.000000,0.000000}%
\pgfsetstrokecolor{textcolor}%
\pgfsetfillcolor{textcolor}%
\pgftext[x=1.527204in,y=3.888333in,left,base]{\color{textcolor}\fontsize{16.000000}{19.200000}\selectfont original encoding}%
\end{pgfscope}%
\begin{pgfscope}%
\pgfsetbuttcap%
\pgfsetroundjoin%
\pgfsetlinewidth{1.505625pt}%
\definecolor{currentstroke}{rgb}{1.000000,0.498039,0.054902}%
\pgfsetstrokecolor{currentstroke}%
\pgfsetdash{{5.550000pt}{2.400000pt}}{0.000000pt}%
\pgfpathmoveto{\pgfqpoint{0.904982in}{3.654111in}}%
\pgfpathlineto{\pgfqpoint{1.349427in}{3.654111in}}%
\pgfusepath{stroke}%
\end{pgfscope}%
\begin{pgfscope}%
\definecolor{textcolor}{rgb}{0.000000,0.000000,0.000000}%
\pgfsetstrokecolor{textcolor}%
\pgfsetfillcolor{textcolor}%
\pgftext[x=1.527204in,y=3.576333in,left,base]{\color{textcolor}\fontsize{16.000000}{19.200000}\selectfont learned at first UIP}%
\end{pgfscope}%
\begin{pgfscope}%
\pgfsetbuttcap%
\pgfsetroundjoin%
\pgfsetlinewidth{1.505625pt}%
\definecolor{currentstroke}{rgb}{0.839216,0.152941,0.156863}%
\pgfsetstrokecolor{currentstroke}%
\pgfsetdash{{1.500000pt}{2.475000pt}}{0.000000pt}%
\pgfpathmoveto{\pgfqpoint{0.904982in}{3.344111in}}%
\pgfpathlineto{\pgfqpoint{1.349427in}{3.344111in}}%
\pgfusepath{stroke}%
\end{pgfscope}%
\begin{pgfscope}%
\definecolor{textcolor}{rgb}{0.000000,0.000000,0.000000}%
\pgfsetstrokecolor{textcolor}%
\pgfsetfillcolor{textcolor}%
\pgftext[x=1.527204in,y=3.266333in,left,base]{\color{textcolor}\fontsize{16.000000}{19.200000}\selectfont learned at last UIP}%
\end{pgfscope}%
\begin{pgfscope}%
\pgfsetbuttcap%
\pgfsetroundjoin%
\pgfsetlinewidth{1.505625pt}%
\definecolor{currentstroke}{rgb}{0.172549,0.627451,0.172549}%
\pgfsetstrokecolor{currentstroke}%
\pgfsetdash{{9.600000pt}{2.400000pt}{1.500000pt}{2.400000pt}}{0.000000pt}%
\pgfpathmoveto{\pgfqpoint{0.904982in}{3.034111in}}%
\pgfpathlineto{\pgfqpoint{1.349427in}{3.034111in}}%
\pgfusepath{stroke}%
\end{pgfscope}%
\begin{pgfscope}%
\definecolor{textcolor}{rgb}{0.000000,0.000000,0.000000}%
\pgfsetstrokecolor{textcolor}%
\pgfsetfillcolor{textcolor}%
\pgftext[x=1.527204in,y=2.956333in,left,base]{\color{textcolor}\fontsize{16.000000}{19.200000}\selectfont reduced constraint}%
\end{pgfscope}%
\end{pgfpicture}%
\makeatother%
\endgroup%

%% file: figures/cactus_hrp_dlv2.pgf
%% Creator: Matplotlib, PGF backend
%%
%% To include the figure in your LaTeX document, write
%%   \input{<filename>.pgf}
%%
%% Make sure the required packages are loaded in your preamble
%%   \usepackage{pgf}
%%
%% and, on pdftex
%%   \usepackage[utf8]{inputenc}\DeclareUnicodeCharacter{2212}{-}
%%
%% or, on luatex and xetex
%%   \usepackage{unicode-math}
%%
%% Figures using additional raster images can only be included by \input if
%% they are in the same directory as the main LaTeX file. For loading figures
%% from other directories you can use the `import` package
%%   \usepackage{import}
%%
%% and then include the figures with
%%   \import{<path to file>}{<filename>.pgf}
%%
%% Matplotlib used the following preamble
%%   \usepackage{fontspec}
%%
\begingroup%
\makeatletter%
\begin{pgfpicture}%
\pgfpathrectangle{\pgfpointorigin}{\pgfqpoint{5.589831in}{4.254121in}}%
\pgfusepath{use as bounding box, clip}%
\begin{pgfscope}%
\pgfsetbuttcap%
\pgfsetmiterjoin%
\definecolor{currentfill}{rgb}{1.000000,1.000000,1.000000}%
\pgfsetfillcolor{currentfill}%
\pgfsetlinewidth{0.000000pt}%
\definecolor{currentstroke}{rgb}{1.000000,1.000000,1.000000}%
\pgfsetstrokecolor{currentstroke}%
\pgfsetdash{}{0pt}%
\pgfpathmoveto{\pgfqpoint{0.000000in}{0.000000in}}%
\pgfpathlineto{\pgfqpoint{5.589831in}{0.000000in}}%
\pgfpathlineto{\pgfqpoint{5.589831in}{4.254121in}}%
\pgfpathlineto{\pgfqpoint{0.000000in}{4.254121in}}%
\pgfpathclose%
\pgfusepath{fill}%
\end{pgfscope}%
\begin{pgfscope}%
\pgfsetbuttcap%
\pgfsetmiterjoin%
\definecolor{currentfill}{rgb}{1.000000,1.000000,1.000000}%
\pgfsetfillcolor{currentfill}%
\pgfsetlinewidth{0.000000pt}%
\definecolor{currentstroke}{rgb}{0.000000,0.000000,0.000000}%
\pgfsetstrokecolor{currentstroke}%
\pgfsetstrokeopacity{0.000000}%
\pgfsetdash{}{0pt}%
\pgfpathmoveto{\pgfqpoint{0.594914in}{0.547888in}}%
\pgfpathlineto{\pgfqpoint{5.554914in}{0.547888in}}%
\pgfpathlineto{\pgfqpoint{5.554914in}{4.243888in}}%
\pgfpathlineto{\pgfqpoint{0.594914in}{4.243888in}}%
\pgfpathclose%
\pgfusepath{fill}%
\end{pgfscope}%
\begin{pgfscope}%
\pgfsetbuttcap%
\pgfsetroundjoin%
\definecolor{currentfill}{rgb}{0.000000,0.000000,0.000000}%
\pgfsetfillcolor{currentfill}%
\pgfsetlinewidth{0.803000pt}%
\definecolor{currentstroke}{rgb}{0.000000,0.000000,0.000000}%
\pgfsetstrokecolor{currentstroke}%
\pgfsetdash{}{0pt}%
\pgfsys@defobject{currentmarker}{\pgfqpoint{0.000000in}{-0.048611in}}{\pgfqpoint{0.000000in}{0.000000in}}{%
\pgfpathmoveto{\pgfqpoint{0.000000in}{0.000000in}}%
\pgfpathlineto{\pgfqpoint{0.000000in}{-0.048611in}}%
\pgfusepath{stroke,fill}%
}%
\begin{pgfscope}%
\pgfsys@transformshift{0.670065in}{0.547888in}%
\pgfsys@useobject{currentmarker}{}%
\end{pgfscope}%
\end{pgfscope}%
\begin{pgfscope}%
\definecolor{textcolor}{rgb}{0.000000,0.000000,0.000000}%
\pgfsetstrokecolor{textcolor}%
\pgfsetfillcolor{textcolor}%
\pgftext[x=0.670065in,y=0.450666in,,top]{\color{textcolor}\fontsize{16.000000}{19.200000}\selectfont \(\displaystyle 0\)}%
\end{pgfscope}%
\begin{pgfscope}%
\pgfsetbuttcap%
\pgfsetroundjoin%
\definecolor{currentfill}{rgb}{0.000000,0.000000,0.000000}%
\pgfsetfillcolor{currentfill}%
\pgfsetlinewidth{0.803000pt}%
\definecolor{currentstroke}{rgb}{0.000000,0.000000,0.000000}%
\pgfsetstrokecolor{currentstroke}%
\pgfsetdash{}{0pt}%
\pgfsys@defobject{currentmarker}{\pgfqpoint{0.000000in}{-0.048611in}}{\pgfqpoint{0.000000in}{0.000000in}}{%
\pgfpathmoveto{\pgfqpoint{0.000000in}{0.000000in}}%
\pgfpathlineto{\pgfqpoint{0.000000in}{-0.048611in}}%
\pgfusepath{stroke,fill}%
}%
\begin{pgfscope}%
\pgfsys@transformshift{1.271278in}{0.547888in}%
\pgfsys@useobject{currentmarker}{}%
\end{pgfscope}%
\end{pgfscope}%
\begin{pgfscope}%
\definecolor{textcolor}{rgb}{0.000000,0.000000,0.000000}%
\pgfsetstrokecolor{textcolor}%
\pgfsetfillcolor{textcolor}%
\pgftext[x=1.271278in,y=0.450666in,,top]{\color{textcolor}\fontsize{16.000000}{19.200000}\selectfont \(\displaystyle 4\)}%
\end{pgfscope}%
\begin{pgfscope}%
\pgfsetbuttcap%
\pgfsetroundjoin%
\definecolor{currentfill}{rgb}{0.000000,0.000000,0.000000}%
\pgfsetfillcolor{currentfill}%
\pgfsetlinewidth{0.803000pt}%
\definecolor{currentstroke}{rgb}{0.000000,0.000000,0.000000}%
\pgfsetstrokecolor{currentstroke}%
\pgfsetdash{}{0pt}%
\pgfsys@defobject{currentmarker}{\pgfqpoint{0.000000in}{-0.048611in}}{\pgfqpoint{0.000000in}{0.000000in}}{%
\pgfpathmoveto{\pgfqpoint{0.000000in}{0.000000in}}%
\pgfpathlineto{\pgfqpoint{0.000000in}{-0.048611in}}%
\pgfusepath{stroke,fill}%
}%
\begin{pgfscope}%
\pgfsys@transformshift{1.872490in}{0.547888in}%
\pgfsys@useobject{currentmarker}{}%
\end{pgfscope}%
\end{pgfscope}%
\begin{pgfscope}%
\definecolor{textcolor}{rgb}{0.000000,0.000000,0.000000}%
\pgfsetstrokecolor{textcolor}%
\pgfsetfillcolor{textcolor}%
\pgftext[x=1.872490in,y=0.450666in,,top]{\color{textcolor}\fontsize{16.000000}{19.200000}\selectfont \(\displaystyle 8\)}%
\end{pgfscope}%
\begin{pgfscope}%
\pgfsetbuttcap%
\pgfsetroundjoin%
\definecolor{currentfill}{rgb}{0.000000,0.000000,0.000000}%
\pgfsetfillcolor{currentfill}%
\pgfsetlinewidth{0.803000pt}%
\definecolor{currentstroke}{rgb}{0.000000,0.000000,0.000000}%
\pgfsetstrokecolor{currentstroke}%
\pgfsetdash{}{0pt}%
\pgfsys@defobject{currentmarker}{\pgfqpoint{0.000000in}{-0.048611in}}{\pgfqpoint{0.000000in}{0.000000in}}{%
\pgfpathmoveto{\pgfqpoint{0.000000in}{0.000000in}}%
\pgfpathlineto{\pgfqpoint{0.000000in}{-0.048611in}}%
\pgfusepath{stroke,fill}%
}%
\begin{pgfscope}%
\pgfsys@transformshift{2.473702in}{0.547888in}%
\pgfsys@useobject{currentmarker}{}%
\end{pgfscope}%
\end{pgfscope}%
\begin{pgfscope}%
\definecolor{textcolor}{rgb}{0.000000,0.000000,0.000000}%
\pgfsetstrokecolor{textcolor}%
\pgfsetfillcolor{textcolor}%
\pgftext[x=2.473702in,y=0.450666in,,top]{\color{textcolor}\fontsize{16.000000}{19.200000}\selectfont \(\displaystyle 12\)}%
\end{pgfscope}%
\begin{pgfscope}%
\pgfsetbuttcap%
\pgfsetroundjoin%
\definecolor{currentfill}{rgb}{0.000000,0.000000,0.000000}%
\pgfsetfillcolor{currentfill}%
\pgfsetlinewidth{0.803000pt}%
\definecolor{currentstroke}{rgb}{0.000000,0.000000,0.000000}%
\pgfsetstrokecolor{currentstroke}%
\pgfsetdash{}{0pt}%
\pgfsys@defobject{currentmarker}{\pgfqpoint{0.000000in}{-0.048611in}}{\pgfqpoint{0.000000in}{0.000000in}}{%
\pgfpathmoveto{\pgfqpoint{0.000000in}{0.000000in}}%
\pgfpathlineto{\pgfqpoint{0.000000in}{-0.048611in}}%
\pgfusepath{stroke,fill}%
}%
\begin{pgfscope}%
\pgfsys@transformshift{3.074914in}{0.547888in}%
\pgfsys@useobject{currentmarker}{}%
\end{pgfscope}%
\end{pgfscope}%
\begin{pgfscope}%
\definecolor{textcolor}{rgb}{0.000000,0.000000,0.000000}%
\pgfsetstrokecolor{textcolor}%
\pgfsetfillcolor{textcolor}%
\pgftext[x=3.074914in,y=0.450666in,,top]{\color{textcolor}\fontsize{16.000000}{19.200000}\selectfont \(\displaystyle 16\)}%
\end{pgfscope}%
\begin{pgfscope}%
\pgfsetbuttcap%
\pgfsetroundjoin%
\definecolor{currentfill}{rgb}{0.000000,0.000000,0.000000}%
\pgfsetfillcolor{currentfill}%
\pgfsetlinewidth{0.803000pt}%
\definecolor{currentstroke}{rgb}{0.000000,0.000000,0.000000}%
\pgfsetstrokecolor{currentstroke}%
\pgfsetdash{}{0pt}%
\pgfsys@defobject{currentmarker}{\pgfqpoint{0.000000in}{-0.048611in}}{\pgfqpoint{0.000000in}{0.000000in}}{%
\pgfpathmoveto{\pgfqpoint{0.000000in}{0.000000in}}%
\pgfpathlineto{\pgfqpoint{0.000000in}{-0.048611in}}%
\pgfusepath{stroke,fill}%
}%
\begin{pgfscope}%
\pgfsys@transformshift{3.676126in}{0.547888in}%
\pgfsys@useobject{currentmarker}{}%
\end{pgfscope}%
\end{pgfscope}%
\begin{pgfscope}%
\definecolor{textcolor}{rgb}{0.000000,0.000000,0.000000}%
\pgfsetstrokecolor{textcolor}%
\pgfsetfillcolor{textcolor}%
\pgftext[x=3.676126in,y=0.450666in,,top]{\color{textcolor}\fontsize{16.000000}{19.200000}\selectfont \(\displaystyle 20\)}%
\end{pgfscope}%
\begin{pgfscope}%
\pgfsetbuttcap%
\pgfsetroundjoin%
\definecolor{currentfill}{rgb}{0.000000,0.000000,0.000000}%
\pgfsetfillcolor{currentfill}%
\pgfsetlinewidth{0.803000pt}%
\definecolor{currentstroke}{rgb}{0.000000,0.000000,0.000000}%
\pgfsetstrokecolor{currentstroke}%
\pgfsetdash{}{0pt}%
\pgfsys@defobject{currentmarker}{\pgfqpoint{0.000000in}{-0.048611in}}{\pgfqpoint{0.000000in}{0.000000in}}{%
\pgfpathmoveto{\pgfqpoint{0.000000in}{0.000000in}}%
\pgfpathlineto{\pgfqpoint{0.000000in}{-0.048611in}}%
\pgfusepath{stroke,fill}%
}%
\begin{pgfscope}%
\pgfsys@transformshift{4.277338in}{0.547888in}%
\pgfsys@useobject{currentmarker}{}%
\end{pgfscope}%
\end{pgfscope}%
\begin{pgfscope}%
\definecolor{textcolor}{rgb}{0.000000,0.000000,0.000000}%
\pgfsetstrokecolor{textcolor}%
\pgfsetfillcolor{textcolor}%
\pgftext[x=4.277338in,y=0.450666in,,top]{\color{textcolor}\fontsize{16.000000}{19.200000}\selectfont \(\displaystyle 24\)}%
\end{pgfscope}%
\begin{pgfscope}%
\pgfsetbuttcap%
\pgfsetroundjoin%
\definecolor{currentfill}{rgb}{0.000000,0.000000,0.000000}%
\pgfsetfillcolor{currentfill}%
\pgfsetlinewidth{0.803000pt}%
\definecolor{currentstroke}{rgb}{0.000000,0.000000,0.000000}%
\pgfsetstrokecolor{currentstroke}%
\pgfsetdash{}{0pt}%
\pgfsys@defobject{currentmarker}{\pgfqpoint{0.000000in}{-0.048611in}}{\pgfqpoint{0.000000in}{0.000000in}}{%
\pgfpathmoveto{\pgfqpoint{0.000000in}{0.000000in}}%
\pgfpathlineto{\pgfqpoint{0.000000in}{-0.048611in}}%
\pgfusepath{stroke,fill}%
}%
\begin{pgfscope}%
\pgfsys@transformshift{4.878550in}{0.547888in}%
\pgfsys@useobject{currentmarker}{}%
\end{pgfscope}%
\end{pgfscope}%
\begin{pgfscope}%
\definecolor{textcolor}{rgb}{0.000000,0.000000,0.000000}%
\pgfsetstrokecolor{textcolor}%
\pgfsetfillcolor{textcolor}%
\pgftext[x=4.878550in,y=0.450666in,,top]{\color{textcolor}\fontsize{16.000000}{19.200000}\selectfont \(\displaystyle 28\)}%
\end{pgfscope}%
\begin{pgfscope}%
\pgfsetbuttcap%
\pgfsetroundjoin%
\definecolor{currentfill}{rgb}{0.000000,0.000000,0.000000}%
\pgfsetfillcolor{currentfill}%
\pgfsetlinewidth{0.803000pt}%
\definecolor{currentstroke}{rgb}{0.000000,0.000000,0.000000}%
\pgfsetstrokecolor{currentstroke}%
\pgfsetdash{}{0pt}%
\pgfsys@defobject{currentmarker}{\pgfqpoint{0.000000in}{-0.048611in}}{\pgfqpoint{0.000000in}{0.000000in}}{%
\pgfpathmoveto{\pgfqpoint{0.000000in}{0.000000in}}%
\pgfpathlineto{\pgfqpoint{0.000000in}{-0.048611in}}%
\pgfusepath{stroke,fill}%
}%
\begin{pgfscope}%
\pgfsys@transformshift{5.479762in}{0.547888in}%
\pgfsys@useobject{currentmarker}{}%
\end{pgfscope}%
\end{pgfscope}%
\begin{pgfscope}%
\definecolor{textcolor}{rgb}{0.000000,0.000000,0.000000}%
\pgfsetstrokecolor{textcolor}%
\pgfsetfillcolor{textcolor}%
\pgftext[x=5.479762in,y=0.450666in,,top]{\color{textcolor}\fontsize{16.000000}{19.200000}\selectfont \(\displaystyle 32\)}%
\end{pgfscope}%
\begin{pgfscope}%
\definecolor{textcolor}{rgb}{0.000000,0.000000,0.000000}%
\pgfsetstrokecolor{textcolor}%
\pgfsetfillcolor{textcolor}%
\pgftext[x=3.074914in,y=0.197555in,,top]{\color{textcolor}\fontsize{16.000000}{19.200000}\selectfont Number of instances}%
\end{pgfscope}%
\begin{pgfscope}%
\pgfsetbuttcap%
\pgfsetroundjoin%
\definecolor{currentfill}{rgb}{0.000000,0.000000,0.000000}%
\pgfsetfillcolor{currentfill}%
\pgfsetlinewidth{0.803000pt}%
\definecolor{currentstroke}{rgb}{0.000000,0.000000,0.000000}%
\pgfsetstrokecolor{currentstroke}%
\pgfsetdash{}{0pt}%
\pgfsys@defobject{currentmarker}{\pgfqpoint{-0.048611in}{0.000000in}}{\pgfqpoint{0.000000in}{0.000000in}}{%
\pgfpathmoveto{\pgfqpoint{0.000000in}{0.000000in}}%
\pgfpathlineto{\pgfqpoint{-0.048611in}{0.000000in}}%
\pgfusepath{stroke,fill}%
}%
\begin{pgfscope}%
\pgfsys@transformshift{0.594914in}{0.715372in}%
\pgfsys@useobject{currentmarker}{}%
\end{pgfscope}%
\end{pgfscope}%
\begin{pgfscope}%
\definecolor{textcolor}{rgb}{0.000000,0.000000,0.000000}%
\pgfsetstrokecolor{textcolor}%
\pgfsetfillcolor{textcolor}%
\pgftext[x=0.387623in, y=0.638261in, left, base]{\color{textcolor}\fontsize{16.000000}{19.200000}\selectfont \(\displaystyle 0\)}%
\end{pgfscope}%
\begin{pgfscope}%
\pgfsetbuttcap%
\pgfsetroundjoin%
\definecolor{currentfill}{rgb}{0.000000,0.000000,0.000000}%
\pgfsetfillcolor{currentfill}%
\pgfsetlinewidth{0.803000pt}%
\definecolor{currentstroke}{rgb}{0.000000,0.000000,0.000000}%
\pgfsetstrokecolor{currentstroke}%
\pgfsetdash{}{0pt}%
\pgfsys@defobject{currentmarker}{\pgfqpoint{-0.048611in}{0.000000in}}{\pgfqpoint{0.000000in}{0.000000in}}{%
\pgfpathmoveto{\pgfqpoint{0.000000in}{0.000000in}}%
\pgfpathlineto{\pgfqpoint{-0.048611in}{0.000000in}}%
\pgfusepath{stroke,fill}%
}%
\begin{pgfscope}%
\pgfsys@transformshift{0.594914in}{1.148077in}%
\pgfsys@useobject{currentmarker}{}%
\end{pgfscope}%
\end{pgfscope}%
\begin{pgfscope}%
\definecolor{textcolor}{rgb}{0.000000,0.000000,0.000000}%
\pgfsetstrokecolor{textcolor}%
\pgfsetfillcolor{textcolor}%
\pgftext[x=0.387623in, y=1.070966in, left, base]{\color{textcolor}\fontsize{16.000000}{19.200000}\selectfont \(\displaystyle 5\)}%
\end{pgfscope}%
\begin{pgfscope}%
\pgfsetbuttcap%
\pgfsetroundjoin%
\definecolor{currentfill}{rgb}{0.000000,0.000000,0.000000}%
\pgfsetfillcolor{currentfill}%
\pgfsetlinewidth{0.803000pt}%
\definecolor{currentstroke}{rgb}{0.000000,0.000000,0.000000}%
\pgfsetstrokecolor{currentstroke}%
\pgfsetdash{}{0pt}%
\pgfsys@defobject{currentmarker}{\pgfqpoint{-0.048611in}{0.000000in}}{\pgfqpoint{0.000000in}{0.000000in}}{%
\pgfpathmoveto{\pgfqpoint{0.000000in}{0.000000in}}%
\pgfpathlineto{\pgfqpoint{-0.048611in}{0.000000in}}%
\pgfusepath{stroke,fill}%
}%
\begin{pgfscope}%
\pgfsys@transformshift{0.594914in}{1.580781in}%
\pgfsys@useobject{currentmarker}{}%
\end{pgfscope}%
\end{pgfscope}%
\begin{pgfscope}%
\definecolor{textcolor}{rgb}{0.000000,0.000000,0.000000}%
\pgfsetstrokecolor{textcolor}%
\pgfsetfillcolor{textcolor}%
\pgftext[x=0.277555in, y=1.503670in, left, base]{\color{textcolor}\fontsize{16.000000}{19.200000}\selectfont \(\displaystyle 10\)}%
\end{pgfscope}%
\begin{pgfscope}%
\pgfsetbuttcap%
\pgfsetroundjoin%
\definecolor{currentfill}{rgb}{0.000000,0.000000,0.000000}%
\pgfsetfillcolor{currentfill}%
\pgfsetlinewidth{0.803000pt}%
\definecolor{currentstroke}{rgb}{0.000000,0.000000,0.000000}%
\pgfsetstrokecolor{currentstroke}%
\pgfsetdash{}{0pt}%
\pgfsys@defobject{currentmarker}{\pgfqpoint{-0.048611in}{0.000000in}}{\pgfqpoint{0.000000in}{0.000000in}}{%
\pgfpathmoveto{\pgfqpoint{0.000000in}{0.000000in}}%
\pgfpathlineto{\pgfqpoint{-0.048611in}{0.000000in}}%
\pgfusepath{stroke,fill}%
}%
\begin{pgfscope}%
\pgfsys@transformshift{0.594914in}{2.013486in}%
\pgfsys@useobject{currentmarker}{}%
\end{pgfscope}%
\end{pgfscope}%
\begin{pgfscope}%
\definecolor{textcolor}{rgb}{0.000000,0.000000,0.000000}%
\pgfsetstrokecolor{textcolor}%
\pgfsetfillcolor{textcolor}%
\pgftext[x=0.277555in, y=1.936375in, left, base]{\color{textcolor}\fontsize{16.000000}{19.200000}\selectfont \(\displaystyle 15\)}%
\end{pgfscope}%
\begin{pgfscope}%
\pgfsetbuttcap%
\pgfsetroundjoin%
\definecolor{currentfill}{rgb}{0.000000,0.000000,0.000000}%
\pgfsetfillcolor{currentfill}%
\pgfsetlinewidth{0.803000pt}%
\definecolor{currentstroke}{rgb}{0.000000,0.000000,0.000000}%
\pgfsetstrokecolor{currentstroke}%
\pgfsetdash{}{0pt}%
\pgfsys@defobject{currentmarker}{\pgfqpoint{-0.048611in}{0.000000in}}{\pgfqpoint{0.000000in}{0.000000in}}{%
\pgfpathmoveto{\pgfqpoint{0.000000in}{0.000000in}}%
\pgfpathlineto{\pgfqpoint{-0.048611in}{0.000000in}}%
\pgfusepath{stroke,fill}%
}%
\begin{pgfscope}%
\pgfsys@transformshift{0.594914in}{2.446191in}%
\pgfsys@useobject{currentmarker}{}%
\end{pgfscope}%
\end{pgfscope}%
\begin{pgfscope}%
\definecolor{textcolor}{rgb}{0.000000,0.000000,0.000000}%
\pgfsetstrokecolor{textcolor}%
\pgfsetfillcolor{textcolor}%
\pgftext[x=0.277555in, y=2.369080in, left, base]{\color{textcolor}\fontsize{16.000000}{19.200000}\selectfont \(\displaystyle 20\)}%
\end{pgfscope}%
\begin{pgfscope}%
\pgfsetbuttcap%
\pgfsetroundjoin%
\definecolor{currentfill}{rgb}{0.000000,0.000000,0.000000}%
\pgfsetfillcolor{currentfill}%
\pgfsetlinewidth{0.803000pt}%
\definecolor{currentstroke}{rgb}{0.000000,0.000000,0.000000}%
\pgfsetstrokecolor{currentstroke}%
\pgfsetdash{}{0pt}%
\pgfsys@defobject{currentmarker}{\pgfqpoint{-0.048611in}{0.000000in}}{\pgfqpoint{0.000000in}{0.000000in}}{%
\pgfpathmoveto{\pgfqpoint{0.000000in}{0.000000in}}%
\pgfpathlineto{\pgfqpoint{-0.048611in}{0.000000in}}%
\pgfusepath{stroke,fill}%
}%
\begin{pgfscope}%
\pgfsys@transformshift{0.594914in}{2.878896in}%
\pgfsys@useobject{currentmarker}{}%
\end{pgfscope}%
\end{pgfscope}%
\begin{pgfscope}%
\definecolor{textcolor}{rgb}{0.000000,0.000000,0.000000}%
\pgfsetstrokecolor{textcolor}%
\pgfsetfillcolor{textcolor}%
\pgftext[x=0.277555in, y=2.801785in, left, base]{\color{textcolor}\fontsize{16.000000}{19.200000}\selectfont \(\displaystyle 25\)}%
\end{pgfscope}%
\begin{pgfscope}%
\pgfsetbuttcap%
\pgfsetroundjoin%
\definecolor{currentfill}{rgb}{0.000000,0.000000,0.000000}%
\pgfsetfillcolor{currentfill}%
\pgfsetlinewidth{0.803000pt}%
\definecolor{currentstroke}{rgb}{0.000000,0.000000,0.000000}%
\pgfsetstrokecolor{currentstroke}%
\pgfsetdash{}{0pt}%
\pgfsys@defobject{currentmarker}{\pgfqpoint{-0.048611in}{0.000000in}}{\pgfqpoint{0.000000in}{0.000000in}}{%
\pgfpathmoveto{\pgfqpoint{0.000000in}{0.000000in}}%
\pgfpathlineto{\pgfqpoint{-0.048611in}{0.000000in}}%
\pgfusepath{stroke,fill}%
}%
\begin{pgfscope}%
\pgfsys@transformshift{0.594914in}{3.311600in}%
\pgfsys@useobject{currentmarker}{}%
\end{pgfscope}%
\end{pgfscope}%
\begin{pgfscope}%
\definecolor{textcolor}{rgb}{0.000000,0.000000,0.000000}%
\pgfsetstrokecolor{textcolor}%
\pgfsetfillcolor{textcolor}%
\pgftext[x=0.277555in, y=3.234489in, left, base]{\color{textcolor}\fontsize{16.000000}{19.200000}\selectfont \(\displaystyle 30\)}%
\end{pgfscope}%
\begin{pgfscope}%
\pgfsetbuttcap%
\pgfsetroundjoin%
\definecolor{currentfill}{rgb}{0.000000,0.000000,0.000000}%
\pgfsetfillcolor{currentfill}%
\pgfsetlinewidth{0.803000pt}%
\definecolor{currentstroke}{rgb}{0.000000,0.000000,0.000000}%
\pgfsetstrokecolor{currentstroke}%
\pgfsetdash{}{0pt}%
\pgfsys@defobject{currentmarker}{\pgfqpoint{-0.048611in}{0.000000in}}{\pgfqpoint{0.000000in}{0.000000in}}{%
\pgfpathmoveto{\pgfqpoint{0.000000in}{0.000000in}}%
\pgfpathlineto{\pgfqpoint{-0.048611in}{0.000000in}}%
\pgfusepath{stroke,fill}%
}%
\begin{pgfscope}%
\pgfsys@transformshift{0.594914in}{3.744305in}%
\pgfsys@useobject{currentmarker}{}%
\end{pgfscope}%
\end{pgfscope}%
\begin{pgfscope}%
\definecolor{textcolor}{rgb}{0.000000,0.000000,0.000000}%
\pgfsetstrokecolor{textcolor}%
\pgfsetfillcolor{textcolor}%
\pgftext[x=0.277555in, y=3.667194in, left, base]{\color{textcolor}\fontsize{16.000000}{19.200000}\selectfont \(\displaystyle 35\)}%
\end{pgfscope}%
\begin{pgfscope}%
\pgfsetbuttcap%
\pgfsetroundjoin%
\definecolor{currentfill}{rgb}{0.000000,0.000000,0.000000}%
\pgfsetfillcolor{currentfill}%
\pgfsetlinewidth{0.803000pt}%
\definecolor{currentstroke}{rgb}{0.000000,0.000000,0.000000}%
\pgfsetstrokecolor{currentstroke}%
\pgfsetdash{}{0pt}%
\pgfsys@defobject{currentmarker}{\pgfqpoint{-0.048611in}{0.000000in}}{\pgfqpoint{0.000000in}{0.000000in}}{%
\pgfpathmoveto{\pgfqpoint{0.000000in}{0.000000in}}%
\pgfpathlineto{\pgfqpoint{-0.048611in}{0.000000in}}%
\pgfusepath{stroke,fill}%
}%
\begin{pgfscope}%
\pgfsys@transformshift{0.594914in}{4.177010in}%
\pgfsys@useobject{currentmarker}{}%
\end{pgfscope}%
\end{pgfscope}%
\begin{pgfscope}%
\definecolor{textcolor}{rgb}{0.000000,0.000000,0.000000}%
\pgfsetstrokecolor{textcolor}%
\pgfsetfillcolor{textcolor}%
\pgftext[x=0.277555in, y=4.099899in, left, base]{\color{textcolor}\fontsize{16.000000}{19.200000}\selectfont \(\displaystyle 40\)}%
\end{pgfscope}%
\begin{pgfscope}%
\definecolor{textcolor}{rgb}{0.000000,0.000000,0.000000}%
\pgfsetstrokecolor{textcolor}%
\pgfsetfillcolor{textcolor}%
\pgftext[x=0.222000in,y=2.395888in,,bottom,rotate=90.000000]{\color{textcolor}\fontsize{16.000000}{19.200000}\selectfont Real time consumption (minutes)}%
\end{pgfscope}%
\begin{pgfscope}%
\pgfpathrectangle{\pgfqpoint{0.594914in}{0.547888in}}{\pgfqpoint{4.960000in}{3.696000in}}%
\pgfusepath{clip}%
\pgfsetrectcap%
\pgfsetroundjoin%
\pgfsetlinewidth{1.505625pt}%
\definecolor{currentstroke}{rgb}{0.498039,0.498039,0.498039}%
\pgfsetstrokecolor{currentstroke}%
\pgfsetdash{}{0pt}%
\pgfpathmoveto{\pgfqpoint{0.820368in}{0.715888in}}%
\pgfpathlineto{\pgfqpoint{0.970671in}{0.716708in}}%
\pgfpathlineto{\pgfqpoint{1.120974in}{0.717876in}}%
\pgfpathlineto{\pgfqpoint{1.271278in}{0.722864in}}%
\pgfpathlineto{\pgfqpoint{1.421581in}{0.729425in}}%
\pgfpathlineto{\pgfqpoint{1.571884in}{0.736484in}}%
\pgfpathlineto{\pgfqpoint{1.722187in}{0.745821in}}%
\pgfpathlineto{\pgfqpoint{1.872490in}{0.759084in}}%
\pgfpathlineto{\pgfqpoint{2.022793in}{0.786685in}}%
\pgfpathlineto{\pgfqpoint{2.173096in}{0.829039in}}%
\pgfpathlineto{\pgfqpoint{2.323399in}{0.876888in}}%
\pgfpathlineto{\pgfqpoint{2.473702in}{0.929395in}}%
\pgfpathlineto{\pgfqpoint{2.624005in}{0.999082in}}%
\pgfpathlineto{\pgfqpoint{2.774308in}{1.072489in}}%
\pgfpathlineto{\pgfqpoint{2.924611in}{1.157465in}}%
\pgfpathlineto{\pgfqpoint{3.074914in}{1.252886in}}%
\pgfpathlineto{\pgfqpoint{3.225217in}{1.380997in}}%
\pgfpathlineto{\pgfqpoint{3.375520in}{1.566941in}}%
\pgfpathlineto{\pgfqpoint{3.525823in}{1.870635in}}%
\pgfpathlineto{\pgfqpoint{3.676126in}{2.220948in}}%
\pgfpathlineto{\pgfqpoint{3.826429in}{2.589015in}}%
\pgfpathlineto{\pgfqpoint{3.976732in}{2.997164in}}%
\pgfusepath{stroke}%
\end{pgfscope}%
\begin{pgfscope}%
\pgfpathrectangle{\pgfqpoint{0.594914in}{0.547888in}}{\pgfqpoint{4.960000in}{3.696000in}}%
\pgfusepath{clip}%
\pgfsetbuttcap%
\pgfsetroundjoin%
\pgfsetlinewidth{1.505625pt}%
\definecolor{currentstroke}{rgb}{1.000000,0.498039,0.054902}%
\pgfsetstrokecolor{currentstroke}%
\pgfsetdash{{5.550000pt}{2.400000pt}}{0.000000pt}%
\pgfpathmoveto{\pgfqpoint{0.820368in}{0.716031in}}%
\pgfpathlineto{\pgfqpoint{0.970671in}{0.716692in}}%
\pgfpathlineto{\pgfqpoint{1.120974in}{0.717853in}}%
\pgfpathlineto{\pgfqpoint{1.271278in}{0.719186in}}%
\pgfpathlineto{\pgfqpoint{1.421581in}{0.725442in}}%
\pgfpathlineto{\pgfqpoint{1.571884in}{0.731738in}}%
\pgfpathlineto{\pgfqpoint{1.722187in}{0.740351in}}%
\pgfpathlineto{\pgfqpoint{1.872490in}{0.749683in}}%
\pgfpathlineto{\pgfqpoint{2.022793in}{0.759804in}}%
\pgfpathlineto{\pgfqpoint{2.173096in}{0.771566in}}%
\pgfpathlineto{\pgfqpoint{2.323399in}{0.783881in}}%
\pgfpathlineto{\pgfqpoint{2.473702in}{0.799412in}}%
\pgfpathlineto{\pgfqpoint{2.624005in}{0.817236in}}%
\pgfpathlineto{\pgfqpoint{2.774308in}{0.839852in}}%
\pgfpathlineto{\pgfqpoint{2.924611in}{0.868932in}}%
\pgfpathlineto{\pgfqpoint{3.074914in}{0.898087in}}%
\pgfpathlineto{\pgfqpoint{3.225217in}{0.940132in}}%
\pgfpathlineto{\pgfqpoint{3.375520in}{0.985351in}}%
\pgfpathlineto{\pgfqpoint{3.525823in}{1.040216in}}%
\pgfpathlineto{\pgfqpoint{3.676126in}{1.097238in}}%
\pgfpathlineto{\pgfqpoint{3.826429in}{1.172933in}}%
\pgfpathlineto{\pgfqpoint{3.976732in}{1.262552in}}%
\pgfpathlineto{\pgfqpoint{4.127035in}{1.360528in}}%
\pgfpathlineto{\pgfqpoint{4.277338in}{1.487022in}}%
\pgfpathlineto{\pgfqpoint{4.427641in}{1.637146in}}%
\pgfpathlineto{\pgfqpoint{4.577944in}{1.830715in}}%
\pgfpathlineto{\pgfqpoint{4.728247in}{2.047097in}}%
\pgfpathlineto{\pgfqpoint{4.878550in}{2.357590in}}%
\pgfpathlineto{\pgfqpoint{5.028853in}{2.693215in}}%
\pgfpathlineto{\pgfqpoint{5.179156in}{3.061985in}}%
\pgfpathlineto{\pgfqpoint{5.329459in}{3.507888in}}%
\pgfusepath{stroke}%
\end{pgfscope}%
\begin{pgfscope}%
\pgfpathrectangle{\pgfqpoint{0.594914in}{0.547888in}}{\pgfqpoint{4.960000in}{3.696000in}}%
\pgfusepath{clip}%
\pgfsetbuttcap%
\pgfsetroundjoin%
\pgfsetlinewidth{1.505625pt}%
\definecolor{currentstroke}{rgb}{0.839216,0.152941,0.156863}%
\pgfsetstrokecolor{currentstroke}%
\pgfsetdash{{1.500000pt}{2.475000pt}}{0.000000pt}%
\pgfpathmoveto{\pgfqpoint{0.820368in}{0.716709in}}%
\pgfpathlineto{\pgfqpoint{0.970671in}{0.719147in}}%
\pgfpathlineto{\pgfqpoint{1.120974in}{0.724749in}}%
\pgfpathlineto{\pgfqpoint{1.271278in}{0.732504in}}%
\pgfpathlineto{\pgfqpoint{1.421581in}{0.742536in}}%
\pgfpathlineto{\pgfqpoint{1.571884in}{0.758547in}}%
\pgfpathlineto{\pgfqpoint{1.722187in}{0.788847in}}%
\pgfpathlineto{\pgfqpoint{1.872490in}{0.887923in}}%
\pgfpathlineto{\pgfqpoint{2.022793in}{1.048918in}}%
\pgfpathlineto{\pgfqpoint{2.173096in}{1.503472in}}%
\pgfpathlineto{\pgfqpoint{2.323399in}{2.118198in}}%
\pgfpathlineto{\pgfqpoint{2.473702in}{2.967075in}}%
\pgfusepath{stroke}%
\end{pgfscope}%
\begin{pgfscope}%
\pgfpathrectangle{\pgfqpoint{0.594914in}{0.547888in}}{\pgfqpoint{4.960000in}{3.696000in}}%
\pgfusepath{clip}%
\pgfsetbuttcap%
\pgfsetroundjoin%
\pgfsetlinewidth{1.505625pt}%
\definecolor{currentstroke}{rgb}{0.172549,0.627451,0.172549}%
\pgfsetstrokecolor{currentstroke}%
\pgfsetdash{{9.600000pt}{2.400000pt}{1.500000pt}{2.400000pt}}{0.000000pt}%
\pgfpathmoveto{\pgfqpoint{0.820368in}{0.716028in}}%
\pgfpathlineto{\pgfqpoint{0.970671in}{0.716686in}}%
\pgfpathlineto{\pgfqpoint{1.120974in}{0.717848in}}%
\pgfpathlineto{\pgfqpoint{1.271278in}{0.719184in}}%
\pgfpathlineto{\pgfqpoint{1.421581in}{0.724463in}}%
\pgfpathlineto{\pgfqpoint{1.571884in}{0.730048in}}%
\pgfpathlineto{\pgfqpoint{1.722187in}{0.735715in}}%
\pgfpathlineto{\pgfqpoint{1.872490in}{0.745806in}}%
\pgfpathlineto{\pgfqpoint{2.022793in}{0.756635in}}%
\pgfpathlineto{\pgfqpoint{2.173096in}{0.767556in}}%
\pgfpathlineto{\pgfqpoint{2.323399in}{0.780971in}}%
\pgfpathlineto{\pgfqpoint{2.473702in}{0.798920in}}%
\pgfpathlineto{\pgfqpoint{2.624005in}{0.829753in}}%
\pgfpathlineto{\pgfqpoint{2.774308in}{0.861567in}}%
\pgfpathlineto{\pgfqpoint{2.924611in}{0.900545in}}%
\pgfpathlineto{\pgfqpoint{3.074914in}{0.949298in}}%
\pgfpathlineto{\pgfqpoint{3.225217in}{1.002736in}}%
\pgfpathlineto{\pgfqpoint{3.375520in}{1.095042in}}%
\pgfpathlineto{\pgfqpoint{3.525823in}{1.196110in}}%
\pgfpathlineto{\pgfqpoint{3.676126in}{1.302988in}}%
\pgfpathlineto{\pgfqpoint{3.826429in}{1.431882in}}%
\pgfpathlineto{\pgfqpoint{3.976732in}{1.577360in}}%
\pgfpathlineto{\pgfqpoint{4.127035in}{1.731660in}}%
\pgfpathlineto{\pgfqpoint{4.277338in}{2.017686in}}%
\pgfpathlineto{\pgfqpoint{4.427641in}{2.328839in}}%
\pgfpathlineto{\pgfqpoint{4.577944in}{2.681033in}}%
\pgfpathlineto{\pgfqpoint{4.728247in}{3.272885in}}%
\pgfpathlineto{\pgfqpoint{4.878550in}{4.075888in}}%
\pgfusepath{stroke}%
\end{pgfscope}%
\begin{pgfscope}%
\pgfsetrectcap%
\pgfsetmiterjoin%
\pgfsetlinewidth{0.803000pt}%
\definecolor{currentstroke}{rgb}{0.000000,0.000000,0.000000}%
\pgfsetstrokecolor{currentstroke}%
\pgfsetdash{}{0pt}%
\pgfpathmoveto{\pgfqpoint{0.594914in}{0.547888in}}%
\pgfpathlineto{\pgfqpoint{0.594914in}{4.243888in}}%
\pgfusepath{stroke}%
\end{pgfscope}%
\begin{pgfscope}%
\pgfsetrectcap%
\pgfsetmiterjoin%
\pgfsetlinewidth{0.803000pt}%
\definecolor{currentstroke}{rgb}{0.000000,0.000000,0.000000}%
\pgfsetstrokecolor{currentstroke}%
\pgfsetdash{}{0pt}%
\pgfpathmoveto{\pgfqpoint{5.554914in}{0.547888in}}%
\pgfpathlineto{\pgfqpoint{5.554914in}{4.243888in}}%
\pgfusepath{stroke}%
\end{pgfscope}%
\begin{pgfscope}%
\pgfsetrectcap%
\pgfsetmiterjoin%
\pgfsetlinewidth{0.803000pt}%
\definecolor{currentstroke}{rgb}{0.000000,0.000000,0.000000}%
\pgfsetstrokecolor{currentstroke}%
\pgfsetdash{}{0pt}%
\pgfpathmoveto{\pgfqpoint{0.594914in}{0.547888in}}%
\pgfpathlineto{\pgfqpoint{5.554914in}{0.547888in}}%
\pgfusepath{stroke}%
\end{pgfscope}%
\begin{pgfscope}%
\pgfsetrectcap%
\pgfsetmiterjoin%
\pgfsetlinewidth{0.803000pt}%
\definecolor{currentstroke}{rgb}{0.000000,0.000000,0.000000}%
\pgfsetstrokecolor{currentstroke}%
\pgfsetdash{}{0pt}%
\pgfpathmoveto{\pgfqpoint{0.594914in}{4.243888in}}%
\pgfpathlineto{\pgfqpoint{5.554914in}{4.243888in}}%
\pgfusepath{stroke}%
\end{pgfscope}%
\begin{pgfscope}%
\pgfsetbuttcap%
\pgfsetmiterjoin%
\definecolor{currentfill}{rgb}{1.000000,1.000000,1.000000}%
\pgfsetfillcolor{currentfill}%
\pgfsetfillopacity{0.800000}%
\pgfsetlinewidth{1.003750pt}%
\definecolor{currentstroke}{rgb}{0.800000,0.800000,0.800000}%
\pgfsetstrokecolor{currentstroke}%
\pgfsetstrokeopacity{0.800000}%
\pgfsetdash{}{0pt}%
\pgfpathmoveto{\pgfqpoint{0.750469in}{2.824112in}}%
\pgfpathlineto{\pgfqpoint{3.201803in}{2.824112in}}%
\pgfpathquadraticcurveto{\pgfqpoint{3.246247in}{2.824112in}}{\pgfqpoint{3.246247in}{2.868556in}}%
\pgfpathlineto{\pgfqpoint{3.246247in}{4.088333in}}%
\pgfpathquadraticcurveto{\pgfqpoint{3.246247in}{4.132777in}}{\pgfqpoint{3.201803in}{4.132777in}}%
\pgfpathlineto{\pgfqpoint{0.750469in}{4.132777in}}%
\pgfpathquadraticcurveto{\pgfqpoint{0.706025in}{4.132777in}}{\pgfqpoint{0.706025in}{4.088333in}}%
\pgfpathlineto{\pgfqpoint{0.706025in}{2.868556in}}%
\pgfpathquadraticcurveto{\pgfqpoint{0.706025in}{2.824112in}}{\pgfqpoint{0.750469in}{2.824112in}}%
\pgfpathclose%
\pgfusepath{stroke,fill}%
\end{pgfscope}%
\begin{pgfscope}%
\pgfsetrectcap%
\pgfsetroundjoin%
\pgfsetlinewidth{1.505625pt}%
\definecolor{currentstroke}{rgb}{0.498039,0.498039,0.498039}%
\pgfsetstrokecolor{currentstroke}%
\pgfsetdash{}{0pt}%
\pgfpathmoveto{\pgfqpoint{0.794914in}{3.966111in}}%
\pgfpathlineto{\pgfqpoint{1.239358in}{3.966111in}}%
\pgfusepath{stroke}%
\end{pgfscope}%
\begin{pgfscope}%
\definecolor{textcolor}{rgb}{0.000000,0.000000,0.000000}%
\pgfsetstrokecolor{textcolor}%
\pgfsetfillcolor{textcolor}%
\pgftext[x=1.417136in,y=3.888333in,left,base]{\color{textcolor}\fontsize{16.000000}{19.200000}\selectfont original encoding}%
\end{pgfscope}%
\begin{pgfscope}%
\pgfsetbuttcap%
\pgfsetroundjoin%
\pgfsetlinewidth{1.505625pt}%
\definecolor{currentstroke}{rgb}{1.000000,0.498039,0.054902}%
\pgfsetstrokecolor{currentstroke}%
\pgfsetdash{{5.550000pt}{2.400000pt}}{0.000000pt}%
\pgfpathmoveto{\pgfqpoint{0.794914in}{3.654111in}}%
\pgfpathlineto{\pgfqpoint{1.239358in}{3.654111in}}%
\pgfusepath{stroke}%
\end{pgfscope}%
\begin{pgfscope}%
\definecolor{textcolor}{rgb}{0.000000,0.000000,0.000000}%
\pgfsetstrokecolor{textcolor}%
\pgfsetfillcolor{textcolor}%
\pgftext[x=1.417136in,y=3.576333in,left,base]{\color{textcolor}\fontsize{16.000000}{19.200000}\selectfont learned at first UIP}%
\end{pgfscope}%
\begin{pgfscope}%
\pgfsetbuttcap%
\pgfsetroundjoin%
\pgfsetlinewidth{1.505625pt}%
\definecolor{currentstroke}{rgb}{0.839216,0.152941,0.156863}%
\pgfsetstrokecolor{currentstroke}%
\pgfsetdash{{1.500000pt}{2.475000pt}}{0.000000pt}%
\pgfpathmoveto{\pgfqpoint{0.794914in}{3.344111in}}%
\pgfpathlineto{\pgfqpoint{1.239358in}{3.344111in}}%
\pgfusepath{stroke}%
\end{pgfscope}%
\begin{pgfscope}%
\definecolor{textcolor}{rgb}{0.000000,0.000000,0.000000}%
\pgfsetstrokecolor{textcolor}%
\pgfsetfillcolor{textcolor}%
\pgftext[x=1.417136in,y=3.266333in,left,base]{\color{textcolor}\fontsize{16.000000}{19.200000}\selectfont learned at last UIP}%
\end{pgfscope}%
\begin{pgfscope}%
\pgfsetbuttcap%
\pgfsetroundjoin%
\pgfsetlinewidth{1.505625pt}%
\definecolor{currentstroke}{rgb}{0.172549,0.627451,0.172549}%
\pgfsetstrokecolor{currentstroke}%
\pgfsetdash{{9.600000pt}{2.400000pt}{1.500000pt}{2.400000pt}}{0.000000pt}%
\pgfpathmoveto{\pgfqpoint{0.794914in}{3.034111in}}%
\pgfpathlineto{\pgfqpoint{1.239358in}{3.034111in}}%
\pgfusepath{stroke}%
\end{pgfscope}%
\begin{pgfscope}%
\definecolor{textcolor}{rgb}{0.000000,0.000000,0.000000}%
\pgfsetstrokecolor{textcolor}%
\pgfsetfillcolor{textcolor}%
\pgftext[x=1.417136in,y=2.956333in,left,base]{\color{textcolor}\fontsize{16.000000}{19.200000}\selectfont reduced constraint}%
\end{pgfscope}%
\end{pgfpicture}%
\makeatother%
\endgroup%

%% file: figures/cactus_3col_crafted_dlv2.pgf
%% Creator: Matplotlib, PGF backend
%%
%% To include the figure in your LaTeX document, write
%%   \input{<filename>.pgf}
%%
%% Make sure the required packages are loaded in your preamble
%%   \usepackage{pgf}
%%
%% and, on pdftex
%%   \usepackage[utf8]{inputenc}\DeclareUnicodeCharacter{2212}{-}
%%
%% or, on luatex and xetex
%%   \usepackage{unicode-math}
%%
%% Figures using additional raster images can only be included by \input if
%% they are in the same directory as the main LaTeX file. For loading figures
%% from other directories you can use the `import` package
%%   \usepackage{import}
%%
%% and then include the figures with
%%   \import{<path to file>}{<filename>.pgf}
%%
%% Matplotlib used the following preamble
%%   \usepackage{fontspec}
%%
\begingroup%
\makeatletter%
\begin{pgfpicture}%
\pgfpathrectangle{\pgfpointorigin}{\pgfqpoint{5.664982in}{4.243888in}}%
\pgfusepath{use as bounding box, clip}%
\begin{pgfscope}%
\pgfsetbuttcap%
\pgfsetmiterjoin%
\definecolor{currentfill}{rgb}{1.000000,1.000000,1.000000}%
\pgfsetfillcolor{currentfill}%
\pgfsetlinewidth{0.000000pt}%
\definecolor{currentstroke}{rgb}{1.000000,1.000000,1.000000}%
\pgfsetstrokecolor{currentstroke}%
\pgfsetdash{}{0pt}%
\pgfpathmoveto{\pgfqpoint{0.000000in}{0.000000in}}%
\pgfpathlineto{\pgfqpoint{5.664982in}{0.000000in}}%
\pgfpathlineto{\pgfqpoint{5.664982in}{4.243888in}}%
\pgfpathlineto{\pgfqpoint{0.000000in}{4.243888in}}%
\pgfpathclose%
\pgfusepath{fill}%
\end{pgfscope}%
\begin{pgfscope}%
\pgfsetbuttcap%
\pgfsetmiterjoin%
\definecolor{currentfill}{rgb}{1.000000,1.000000,1.000000}%
\pgfsetfillcolor{currentfill}%
\pgfsetlinewidth{0.000000pt}%
\definecolor{currentstroke}{rgb}{0.000000,0.000000,0.000000}%
\pgfsetstrokecolor{currentstroke}%
\pgfsetstrokeopacity{0.000000}%
\pgfsetdash{}{0pt}%
\pgfpathmoveto{\pgfqpoint{0.704982in}{0.547888in}}%
\pgfpathlineto{\pgfqpoint{5.664982in}{0.547888in}}%
\pgfpathlineto{\pgfqpoint{5.664982in}{4.243888in}}%
\pgfpathlineto{\pgfqpoint{0.704982in}{4.243888in}}%
\pgfpathclose%
\pgfusepath{fill}%
\end{pgfscope}%
\begin{pgfscope}%
\pgfsetbuttcap%
\pgfsetroundjoin%
\definecolor{currentfill}{rgb}{0.000000,0.000000,0.000000}%
\pgfsetfillcolor{currentfill}%
\pgfsetlinewidth{0.803000pt}%
\definecolor{currentstroke}{rgb}{0.000000,0.000000,0.000000}%
\pgfsetstrokecolor{currentstroke}%
\pgfsetdash{}{0pt}%
\pgfsys@defobject{currentmarker}{\pgfqpoint{0.000000in}{-0.048611in}}{\pgfqpoint{0.000000in}{0.000000in}}{%
\pgfpathmoveto{\pgfqpoint{0.000000in}{0.000000in}}%
\pgfpathlineto{\pgfqpoint{0.000000in}{-0.048611in}}%
\pgfusepath{stroke,fill}%
}%
\begin{pgfscope}%
\pgfsys@transformshift{0.907778in}{0.547888in}%
\pgfsys@useobject{currentmarker}{}%
\end{pgfscope}%
\end{pgfscope}%
\begin{pgfscope}%
\definecolor{textcolor}{rgb}{0.000000,0.000000,0.000000}%
\pgfsetstrokecolor{textcolor}%
\pgfsetfillcolor{textcolor}%
\pgftext[x=0.907778in,y=0.450666in,,top]{\color{textcolor}\fontsize{16.000000}{19.200000}\selectfont \(\displaystyle 0\)}%
\end{pgfscope}%
\begin{pgfscope}%
\pgfsetbuttcap%
\pgfsetroundjoin%
\definecolor{currentfill}{rgb}{0.000000,0.000000,0.000000}%
\pgfsetfillcolor{currentfill}%
\pgfsetlinewidth{0.803000pt}%
\definecolor{currentstroke}{rgb}{0.000000,0.000000,0.000000}%
\pgfsetstrokecolor{currentstroke}%
\pgfsetdash{}{0pt}%
\pgfsys@defobject{currentmarker}{\pgfqpoint{0.000000in}{-0.048611in}}{\pgfqpoint{0.000000in}{0.000000in}}{%
\pgfpathmoveto{\pgfqpoint{0.000000in}{0.000000in}}%
\pgfpathlineto{\pgfqpoint{0.000000in}{-0.048611in}}%
\pgfusepath{stroke,fill}%
}%
\begin{pgfscope}%
\pgfsys@transformshift{1.474247in}{0.547888in}%
\pgfsys@useobject{currentmarker}{}%
\end{pgfscope}%
\end{pgfscope}%
\begin{pgfscope}%
\definecolor{textcolor}{rgb}{0.000000,0.000000,0.000000}%
\pgfsetstrokecolor{textcolor}%
\pgfsetfillcolor{textcolor}%
\pgftext[x=1.474247in,y=0.450666in,,top]{\color{textcolor}\fontsize{16.000000}{19.200000}\selectfont \(\displaystyle 25\)}%
\end{pgfscope}%
\begin{pgfscope}%
\pgfsetbuttcap%
\pgfsetroundjoin%
\definecolor{currentfill}{rgb}{0.000000,0.000000,0.000000}%
\pgfsetfillcolor{currentfill}%
\pgfsetlinewidth{0.803000pt}%
\definecolor{currentstroke}{rgb}{0.000000,0.000000,0.000000}%
\pgfsetstrokecolor{currentstroke}%
\pgfsetdash{}{0pt}%
\pgfsys@defobject{currentmarker}{\pgfqpoint{0.000000in}{-0.048611in}}{\pgfqpoint{0.000000in}{0.000000in}}{%
\pgfpathmoveto{\pgfqpoint{0.000000in}{0.000000in}}%
\pgfpathlineto{\pgfqpoint{0.000000in}{-0.048611in}}%
\pgfusepath{stroke,fill}%
}%
\begin{pgfscope}%
\pgfsys@transformshift{2.040715in}{0.547888in}%
\pgfsys@useobject{currentmarker}{}%
\end{pgfscope}%
\end{pgfscope}%
\begin{pgfscope}%
\definecolor{textcolor}{rgb}{0.000000,0.000000,0.000000}%
\pgfsetstrokecolor{textcolor}%
\pgfsetfillcolor{textcolor}%
\pgftext[x=2.040715in,y=0.450666in,,top]{\color{textcolor}\fontsize{16.000000}{19.200000}\selectfont \(\displaystyle 50\)}%
\end{pgfscope}%
\begin{pgfscope}%
\pgfsetbuttcap%
\pgfsetroundjoin%
\definecolor{currentfill}{rgb}{0.000000,0.000000,0.000000}%
\pgfsetfillcolor{currentfill}%
\pgfsetlinewidth{0.803000pt}%
\definecolor{currentstroke}{rgb}{0.000000,0.000000,0.000000}%
\pgfsetstrokecolor{currentstroke}%
\pgfsetdash{}{0pt}%
\pgfsys@defobject{currentmarker}{\pgfqpoint{0.000000in}{-0.048611in}}{\pgfqpoint{0.000000in}{0.000000in}}{%
\pgfpathmoveto{\pgfqpoint{0.000000in}{0.000000in}}%
\pgfpathlineto{\pgfqpoint{0.000000in}{-0.048611in}}%
\pgfusepath{stroke,fill}%
}%
\begin{pgfscope}%
\pgfsys@transformshift{2.607184in}{0.547888in}%
\pgfsys@useobject{currentmarker}{}%
\end{pgfscope}%
\end{pgfscope}%
\begin{pgfscope}%
\definecolor{textcolor}{rgb}{0.000000,0.000000,0.000000}%
\pgfsetstrokecolor{textcolor}%
\pgfsetfillcolor{textcolor}%
\pgftext[x=2.607184in,y=0.450666in,,top]{\color{textcolor}\fontsize{16.000000}{19.200000}\selectfont \(\displaystyle 75\)}%
\end{pgfscope}%
\begin{pgfscope}%
\pgfsetbuttcap%
\pgfsetroundjoin%
\definecolor{currentfill}{rgb}{0.000000,0.000000,0.000000}%
\pgfsetfillcolor{currentfill}%
\pgfsetlinewidth{0.803000pt}%
\definecolor{currentstroke}{rgb}{0.000000,0.000000,0.000000}%
\pgfsetstrokecolor{currentstroke}%
\pgfsetdash{}{0pt}%
\pgfsys@defobject{currentmarker}{\pgfqpoint{0.000000in}{-0.048611in}}{\pgfqpoint{0.000000in}{0.000000in}}{%
\pgfpathmoveto{\pgfqpoint{0.000000in}{0.000000in}}%
\pgfpathlineto{\pgfqpoint{0.000000in}{-0.048611in}}%
\pgfusepath{stroke,fill}%
}%
\begin{pgfscope}%
\pgfsys@transformshift{3.173653in}{0.547888in}%
\pgfsys@useobject{currentmarker}{}%
\end{pgfscope}%
\end{pgfscope}%
\begin{pgfscope}%
\definecolor{textcolor}{rgb}{0.000000,0.000000,0.000000}%
\pgfsetstrokecolor{textcolor}%
\pgfsetfillcolor{textcolor}%
\pgftext[x=3.173653in,y=0.450666in,,top]{\color{textcolor}\fontsize{16.000000}{19.200000}\selectfont \(\displaystyle 100\)}%
\end{pgfscope}%
\begin{pgfscope}%
\pgfsetbuttcap%
\pgfsetroundjoin%
\definecolor{currentfill}{rgb}{0.000000,0.000000,0.000000}%
\pgfsetfillcolor{currentfill}%
\pgfsetlinewidth{0.803000pt}%
\definecolor{currentstroke}{rgb}{0.000000,0.000000,0.000000}%
\pgfsetstrokecolor{currentstroke}%
\pgfsetdash{}{0pt}%
\pgfsys@defobject{currentmarker}{\pgfqpoint{0.000000in}{-0.048611in}}{\pgfqpoint{0.000000in}{0.000000in}}{%
\pgfpathmoveto{\pgfqpoint{0.000000in}{0.000000in}}%
\pgfpathlineto{\pgfqpoint{0.000000in}{-0.048611in}}%
\pgfusepath{stroke,fill}%
}%
\begin{pgfscope}%
\pgfsys@transformshift{3.740121in}{0.547888in}%
\pgfsys@useobject{currentmarker}{}%
\end{pgfscope}%
\end{pgfscope}%
\begin{pgfscope}%
\definecolor{textcolor}{rgb}{0.000000,0.000000,0.000000}%
\pgfsetstrokecolor{textcolor}%
\pgfsetfillcolor{textcolor}%
\pgftext[x=3.740121in,y=0.450666in,,top]{\color{textcolor}\fontsize{16.000000}{19.200000}\selectfont \(\displaystyle 125\)}%
\end{pgfscope}%
\begin{pgfscope}%
\pgfsetbuttcap%
\pgfsetroundjoin%
\definecolor{currentfill}{rgb}{0.000000,0.000000,0.000000}%
\pgfsetfillcolor{currentfill}%
\pgfsetlinewidth{0.803000pt}%
\definecolor{currentstroke}{rgb}{0.000000,0.000000,0.000000}%
\pgfsetstrokecolor{currentstroke}%
\pgfsetdash{}{0pt}%
\pgfsys@defobject{currentmarker}{\pgfqpoint{0.000000in}{-0.048611in}}{\pgfqpoint{0.000000in}{0.000000in}}{%
\pgfpathmoveto{\pgfqpoint{0.000000in}{0.000000in}}%
\pgfpathlineto{\pgfqpoint{0.000000in}{-0.048611in}}%
\pgfusepath{stroke,fill}%
}%
\begin{pgfscope}%
\pgfsys@transformshift{4.306590in}{0.547888in}%
\pgfsys@useobject{currentmarker}{}%
\end{pgfscope}%
\end{pgfscope}%
\begin{pgfscope}%
\definecolor{textcolor}{rgb}{0.000000,0.000000,0.000000}%
\pgfsetstrokecolor{textcolor}%
\pgfsetfillcolor{textcolor}%
\pgftext[x=4.306590in,y=0.450666in,,top]{\color{textcolor}\fontsize{16.000000}{19.200000}\selectfont \(\displaystyle 150\)}%
\end{pgfscope}%
\begin{pgfscope}%
\pgfsetbuttcap%
\pgfsetroundjoin%
\definecolor{currentfill}{rgb}{0.000000,0.000000,0.000000}%
\pgfsetfillcolor{currentfill}%
\pgfsetlinewidth{0.803000pt}%
\definecolor{currentstroke}{rgb}{0.000000,0.000000,0.000000}%
\pgfsetstrokecolor{currentstroke}%
\pgfsetdash{}{0pt}%
\pgfsys@defobject{currentmarker}{\pgfqpoint{0.000000in}{-0.048611in}}{\pgfqpoint{0.000000in}{0.000000in}}{%
\pgfpathmoveto{\pgfqpoint{0.000000in}{0.000000in}}%
\pgfpathlineto{\pgfqpoint{0.000000in}{-0.048611in}}%
\pgfusepath{stroke,fill}%
}%
\begin{pgfscope}%
\pgfsys@transformshift{4.873059in}{0.547888in}%
\pgfsys@useobject{currentmarker}{}%
\end{pgfscope}%
\end{pgfscope}%
\begin{pgfscope}%
\definecolor{textcolor}{rgb}{0.000000,0.000000,0.000000}%
\pgfsetstrokecolor{textcolor}%
\pgfsetfillcolor{textcolor}%
\pgftext[x=4.873059in,y=0.450666in,,top]{\color{textcolor}\fontsize{16.000000}{19.200000}\selectfont \(\displaystyle 175\)}%
\end{pgfscope}%
\begin{pgfscope}%
\pgfsetbuttcap%
\pgfsetroundjoin%
\definecolor{currentfill}{rgb}{0.000000,0.000000,0.000000}%
\pgfsetfillcolor{currentfill}%
\pgfsetlinewidth{0.803000pt}%
\definecolor{currentstroke}{rgb}{0.000000,0.000000,0.000000}%
\pgfsetstrokecolor{currentstroke}%
\pgfsetdash{}{0pt}%
\pgfsys@defobject{currentmarker}{\pgfqpoint{0.000000in}{-0.048611in}}{\pgfqpoint{0.000000in}{0.000000in}}{%
\pgfpathmoveto{\pgfqpoint{0.000000in}{0.000000in}}%
\pgfpathlineto{\pgfqpoint{0.000000in}{-0.048611in}}%
\pgfusepath{stroke,fill}%
}%
\begin{pgfscope}%
\pgfsys@transformshift{5.439528in}{0.547888in}%
\pgfsys@useobject{currentmarker}{}%
\end{pgfscope}%
\end{pgfscope}%
\begin{pgfscope}%
\definecolor{textcolor}{rgb}{0.000000,0.000000,0.000000}%
\pgfsetstrokecolor{textcolor}%
\pgfsetfillcolor{textcolor}%
\pgftext[x=5.439528in,y=0.450666in,,top]{\color{textcolor}\fontsize{16.000000}{19.200000}\selectfont \(\displaystyle 200\)}%
\end{pgfscope}%
\begin{pgfscope}%
\definecolor{textcolor}{rgb}{0.000000,0.000000,0.000000}%
\pgfsetstrokecolor{textcolor}%
\pgfsetfillcolor{textcolor}%
\pgftext[x=3.184982in,y=0.197555in,,top]{\color{textcolor}\fontsize{16.000000}{19.200000}\selectfont Number of instances}%
\end{pgfscope}%
\begin{pgfscope}%
\pgfsetbuttcap%
\pgfsetroundjoin%
\definecolor{currentfill}{rgb}{0.000000,0.000000,0.000000}%
\pgfsetfillcolor{currentfill}%
\pgfsetlinewidth{0.803000pt}%
\definecolor{currentstroke}{rgb}{0.000000,0.000000,0.000000}%
\pgfsetstrokecolor{currentstroke}%
\pgfsetdash{}{0pt}%
\pgfsys@defobject{currentmarker}{\pgfqpoint{-0.048611in}{0.000000in}}{\pgfqpoint{0.000000in}{0.000000in}}{%
\pgfpathmoveto{\pgfqpoint{0.000000in}{0.000000in}}%
\pgfpathlineto{\pgfqpoint{-0.048611in}{0.000000in}}%
\pgfusepath{stroke,fill}%
}%
\begin{pgfscope}%
\pgfsys@transformshift{0.704982in}{0.715872in}%
\pgfsys@useobject{currentmarker}{}%
\end{pgfscope}%
\end{pgfscope}%
\begin{pgfscope}%
\definecolor{textcolor}{rgb}{0.000000,0.000000,0.000000}%
\pgfsetstrokecolor{textcolor}%
\pgfsetfillcolor{textcolor}%
\pgftext[x=0.497692in, y=0.638761in, left, base]{\color{textcolor}\fontsize{16.000000}{19.200000}\selectfont \(\displaystyle 0\)}%
\end{pgfscope}%
\begin{pgfscope}%
\pgfsetbuttcap%
\pgfsetroundjoin%
\definecolor{currentfill}{rgb}{0.000000,0.000000,0.000000}%
\pgfsetfillcolor{currentfill}%
\pgfsetlinewidth{0.803000pt}%
\definecolor{currentstroke}{rgb}{0.000000,0.000000,0.000000}%
\pgfsetstrokecolor{currentstroke}%
\pgfsetdash{}{0pt}%
\pgfsys@defobject{currentmarker}{\pgfqpoint{-0.048611in}{0.000000in}}{\pgfqpoint{0.000000in}{0.000000in}}{%
\pgfpathmoveto{\pgfqpoint{0.000000in}{0.000000in}}%
\pgfpathlineto{\pgfqpoint{-0.048611in}{0.000000in}}%
\pgfusepath{stroke,fill}%
}%
\begin{pgfscope}%
\pgfsys@transformshift{0.704982in}{1.095281in}%
\pgfsys@useobject{currentmarker}{}%
\end{pgfscope}%
\end{pgfscope}%
\begin{pgfscope}%
\definecolor{textcolor}{rgb}{0.000000,0.000000,0.000000}%
\pgfsetstrokecolor{textcolor}%
\pgfsetfillcolor{textcolor}%
\pgftext[x=0.387623in, y=1.018170in, left, base]{\color{textcolor}\fontsize{16.000000}{19.200000}\selectfont \(\displaystyle 40\)}%
\end{pgfscope}%
\begin{pgfscope}%
\pgfsetbuttcap%
\pgfsetroundjoin%
\definecolor{currentfill}{rgb}{0.000000,0.000000,0.000000}%
\pgfsetfillcolor{currentfill}%
\pgfsetlinewidth{0.803000pt}%
\definecolor{currentstroke}{rgb}{0.000000,0.000000,0.000000}%
\pgfsetstrokecolor{currentstroke}%
\pgfsetdash{}{0pt}%
\pgfsys@defobject{currentmarker}{\pgfqpoint{-0.048611in}{0.000000in}}{\pgfqpoint{0.000000in}{0.000000in}}{%
\pgfpathmoveto{\pgfqpoint{0.000000in}{0.000000in}}%
\pgfpathlineto{\pgfqpoint{-0.048611in}{0.000000in}}%
\pgfusepath{stroke,fill}%
}%
\begin{pgfscope}%
\pgfsys@transformshift{0.704982in}{1.474690in}%
\pgfsys@useobject{currentmarker}{}%
\end{pgfscope}%
\end{pgfscope}%
\begin{pgfscope}%
\definecolor{textcolor}{rgb}{0.000000,0.000000,0.000000}%
\pgfsetstrokecolor{textcolor}%
\pgfsetfillcolor{textcolor}%
\pgftext[x=0.387623in, y=1.397579in, left, base]{\color{textcolor}\fontsize{16.000000}{19.200000}\selectfont \(\displaystyle 80\)}%
\end{pgfscope}%
\begin{pgfscope}%
\pgfsetbuttcap%
\pgfsetroundjoin%
\definecolor{currentfill}{rgb}{0.000000,0.000000,0.000000}%
\pgfsetfillcolor{currentfill}%
\pgfsetlinewidth{0.803000pt}%
\definecolor{currentstroke}{rgb}{0.000000,0.000000,0.000000}%
\pgfsetstrokecolor{currentstroke}%
\pgfsetdash{}{0pt}%
\pgfsys@defobject{currentmarker}{\pgfqpoint{-0.048611in}{0.000000in}}{\pgfqpoint{0.000000in}{0.000000in}}{%
\pgfpathmoveto{\pgfqpoint{0.000000in}{0.000000in}}%
\pgfpathlineto{\pgfqpoint{-0.048611in}{0.000000in}}%
\pgfusepath{stroke,fill}%
}%
\begin{pgfscope}%
\pgfsys@transformshift{0.704982in}{1.854099in}%
\pgfsys@useobject{currentmarker}{}%
\end{pgfscope}%
\end{pgfscope}%
\begin{pgfscope}%
\definecolor{textcolor}{rgb}{0.000000,0.000000,0.000000}%
\pgfsetstrokecolor{textcolor}%
\pgfsetfillcolor{textcolor}%
\pgftext[x=0.277555in, y=1.776988in, left, base]{\color{textcolor}\fontsize{16.000000}{19.200000}\selectfont \(\displaystyle 120\)}%
\end{pgfscope}%
\begin{pgfscope}%
\pgfsetbuttcap%
\pgfsetroundjoin%
\definecolor{currentfill}{rgb}{0.000000,0.000000,0.000000}%
\pgfsetfillcolor{currentfill}%
\pgfsetlinewidth{0.803000pt}%
\definecolor{currentstroke}{rgb}{0.000000,0.000000,0.000000}%
\pgfsetstrokecolor{currentstroke}%
\pgfsetdash{}{0pt}%
\pgfsys@defobject{currentmarker}{\pgfqpoint{-0.048611in}{0.000000in}}{\pgfqpoint{0.000000in}{0.000000in}}{%
\pgfpathmoveto{\pgfqpoint{0.000000in}{0.000000in}}%
\pgfpathlineto{\pgfqpoint{-0.048611in}{0.000000in}}%
\pgfusepath{stroke,fill}%
}%
\begin{pgfscope}%
\pgfsys@transformshift{0.704982in}{2.233508in}%
\pgfsys@useobject{currentmarker}{}%
\end{pgfscope}%
\end{pgfscope}%
\begin{pgfscope}%
\definecolor{textcolor}{rgb}{0.000000,0.000000,0.000000}%
\pgfsetstrokecolor{textcolor}%
\pgfsetfillcolor{textcolor}%
\pgftext[x=0.277555in, y=2.156397in, left, base]{\color{textcolor}\fontsize{16.000000}{19.200000}\selectfont \(\displaystyle 160\)}%
\end{pgfscope}%
\begin{pgfscope}%
\pgfsetbuttcap%
\pgfsetroundjoin%
\definecolor{currentfill}{rgb}{0.000000,0.000000,0.000000}%
\pgfsetfillcolor{currentfill}%
\pgfsetlinewidth{0.803000pt}%
\definecolor{currentstroke}{rgb}{0.000000,0.000000,0.000000}%
\pgfsetstrokecolor{currentstroke}%
\pgfsetdash{}{0pt}%
\pgfsys@defobject{currentmarker}{\pgfqpoint{-0.048611in}{0.000000in}}{\pgfqpoint{0.000000in}{0.000000in}}{%
\pgfpathmoveto{\pgfqpoint{0.000000in}{0.000000in}}%
\pgfpathlineto{\pgfqpoint{-0.048611in}{0.000000in}}%
\pgfusepath{stroke,fill}%
}%
\begin{pgfscope}%
\pgfsys@transformshift{0.704982in}{2.612917in}%
\pgfsys@useobject{currentmarker}{}%
\end{pgfscope}%
\end{pgfscope}%
\begin{pgfscope}%
\definecolor{textcolor}{rgb}{0.000000,0.000000,0.000000}%
\pgfsetstrokecolor{textcolor}%
\pgfsetfillcolor{textcolor}%
\pgftext[x=0.277555in, y=2.535806in, left, base]{\color{textcolor}\fontsize{16.000000}{19.200000}\selectfont \(\displaystyle 200\)}%
\end{pgfscope}%
\begin{pgfscope}%
\pgfsetbuttcap%
\pgfsetroundjoin%
\definecolor{currentfill}{rgb}{0.000000,0.000000,0.000000}%
\pgfsetfillcolor{currentfill}%
\pgfsetlinewidth{0.803000pt}%
\definecolor{currentstroke}{rgb}{0.000000,0.000000,0.000000}%
\pgfsetstrokecolor{currentstroke}%
\pgfsetdash{}{0pt}%
\pgfsys@defobject{currentmarker}{\pgfqpoint{-0.048611in}{0.000000in}}{\pgfqpoint{0.000000in}{0.000000in}}{%
\pgfpathmoveto{\pgfqpoint{0.000000in}{0.000000in}}%
\pgfpathlineto{\pgfqpoint{-0.048611in}{0.000000in}}%
\pgfusepath{stroke,fill}%
}%
\begin{pgfscope}%
\pgfsys@transformshift{0.704982in}{2.992326in}%
\pgfsys@useobject{currentmarker}{}%
\end{pgfscope}%
\end{pgfscope}%
\begin{pgfscope}%
\definecolor{textcolor}{rgb}{0.000000,0.000000,0.000000}%
\pgfsetstrokecolor{textcolor}%
\pgfsetfillcolor{textcolor}%
\pgftext[x=0.277555in, y=2.915215in, left, base]{\color{textcolor}\fontsize{16.000000}{19.200000}\selectfont \(\displaystyle 240\)}%
\end{pgfscope}%
\begin{pgfscope}%
\pgfsetbuttcap%
\pgfsetroundjoin%
\definecolor{currentfill}{rgb}{0.000000,0.000000,0.000000}%
\pgfsetfillcolor{currentfill}%
\pgfsetlinewidth{0.803000pt}%
\definecolor{currentstroke}{rgb}{0.000000,0.000000,0.000000}%
\pgfsetstrokecolor{currentstroke}%
\pgfsetdash{}{0pt}%
\pgfsys@defobject{currentmarker}{\pgfqpoint{-0.048611in}{0.000000in}}{\pgfqpoint{0.000000in}{0.000000in}}{%
\pgfpathmoveto{\pgfqpoint{0.000000in}{0.000000in}}%
\pgfpathlineto{\pgfqpoint{-0.048611in}{0.000000in}}%
\pgfusepath{stroke,fill}%
}%
\begin{pgfscope}%
\pgfsys@transformshift{0.704982in}{3.371735in}%
\pgfsys@useobject{currentmarker}{}%
\end{pgfscope}%
\end{pgfscope}%
\begin{pgfscope}%
\definecolor{textcolor}{rgb}{0.000000,0.000000,0.000000}%
\pgfsetstrokecolor{textcolor}%
\pgfsetfillcolor{textcolor}%
\pgftext[x=0.277555in, y=3.294623in, left, base]{\color{textcolor}\fontsize{16.000000}{19.200000}\selectfont \(\displaystyle 280\)}%
\end{pgfscope}%
\begin{pgfscope}%
\pgfsetbuttcap%
\pgfsetroundjoin%
\definecolor{currentfill}{rgb}{0.000000,0.000000,0.000000}%
\pgfsetfillcolor{currentfill}%
\pgfsetlinewidth{0.803000pt}%
\definecolor{currentstroke}{rgb}{0.000000,0.000000,0.000000}%
\pgfsetstrokecolor{currentstroke}%
\pgfsetdash{}{0pt}%
\pgfsys@defobject{currentmarker}{\pgfqpoint{-0.048611in}{0.000000in}}{\pgfqpoint{0.000000in}{0.000000in}}{%
\pgfpathmoveto{\pgfqpoint{0.000000in}{0.000000in}}%
\pgfpathlineto{\pgfqpoint{-0.048611in}{0.000000in}}%
\pgfusepath{stroke,fill}%
}%
\begin{pgfscope}%
\pgfsys@transformshift{0.704982in}{3.751144in}%
\pgfsys@useobject{currentmarker}{}%
\end{pgfscope}%
\end{pgfscope}%
\begin{pgfscope}%
\definecolor{textcolor}{rgb}{0.000000,0.000000,0.000000}%
\pgfsetstrokecolor{textcolor}%
\pgfsetfillcolor{textcolor}%
\pgftext[x=0.277555in, y=3.674032in, left, base]{\color{textcolor}\fontsize{16.000000}{19.200000}\selectfont \(\displaystyle 320\)}%
\end{pgfscope}%
\begin{pgfscope}%
\pgfsetbuttcap%
\pgfsetroundjoin%
\definecolor{currentfill}{rgb}{0.000000,0.000000,0.000000}%
\pgfsetfillcolor{currentfill}%
\pgfsetlinewidth{0.803000pt}%
\definecolor{currentstroke}{rgb}{0.000000,0.000000,0.000000}%
\pgfsetstrokecolor{currentstroke}%
\pgfsetdash{}{0pt}%
\pgfsys@defobject{currentmarker}{\pgfqpoint{-0.048611in}{0.000000in}}{\pgfqpoint{0.000000in}{0.000000in}}{%
\pgfpathmoveto{\pgfqpoint{0.000000in}{0.000000in}}%
\pgfpathlineto{\pgfqpoint{-0.048611in}{0.000000in}}%
\pgfusepath{stroke,fill}%
}%
\begin{pgfscope}%
\pgfsys@transformshift{0.704982in}{4.130552in}%
\pgfsys@useobject{currentmarker}{}%
\end{pgfscope}%
\end{pgfscope}%
\begin{pgfscope}%
\definecolor{textcolor}{rgb}{0.000000,0.000000,0.000000}%
\pgfsetstrokecolor{textcolor}%
\pgfsetfillcolor{textcolor}%
\pgftext[x=0.277555in, y=4.053441in, left, base]{\color{textcolor}\fontsize{16.000000}{19.200000}\selectfont \(\displaystyle 360\)}%
\end{pgfscope}%
\begin{pgfscope}%
\definecolor{textcolor}{rgb}{0.000000,0.000000,0.000000}%
\pgfsetstrokecolor{textcolor}%
\pgfsetfillcolor{textcolor}%
\pgftext[x=0.222000in,y=2.395888in,,bottom,rotate=90.000000]{\color{textcolor}\fontsize{16.000000}{19.200000}\selectfont Real time consumption (minutes)}%
\end{pgfscope}%
\begin{pgfscope}%
\pgfpathrectangle{\pgfqpoint{0.704982in}{0.547888in}}{\pgfqpoint{4.960000in}{3.696000in}}%
\pgfusepath{clip}%
\pgfsetrectcap%
\pgfsetroundjoin%
\pgfsetlinewidth{1.505625pt}%
\definecolor{currentstroke}{rgb}{0.498039,0.498039,0.498039}%
\pgfsetstrokecolor{currentstroke}%
\pgfsetdash{}{0pt}%
\pgfpathmoveto{\pgfqpoint{0.930437in}{0.715889in}}%
\pgfpathlineto{\pgfqpoint{1.179683in}{0.716643in}}%
\pgfpathlineto{\pgfqpoint{1.383612in}{0.718647in}}%
\pgfpathlineto{\pgfqpoint{1.564882in}{0.722037in}}%
\pgfpathlineto{\pgfqpoint{1.700834in}{0.726130in}}%
\pgfpathlineto{\pgfqpoint{1.791469in}{0.729918in}}%
\pgfpathlineto{\pgfqpoint{1.882104in}{0.734732in}}%
\pgfpathlineto{\pgfqpoint{1.950080in}{0.739867in}}%
\pgfpathlineto{\pgfqpoint{2.040715in}{0.748840in}}%
\pgfpathlineto{\pgfqpoint{2.131350in}{0.758658in}}%
\pgfpathlineto{\pgfqpoint{2.312620in}{0.779869in}}%
\pgfpathlineto{\pgfqpoint{2.403255in}{0.791439in}}%
\pgfpathlineto{\pgfqpoint{2.471232in}{0.801301in}}%
\pgfpathlineto{\pgfqpoint{2.539208in}{0.812286in}}%
\pgfpathlineto{\pgfqpoint{2.607184in}{0.824814in}}%
\pgfpathlineto{\pgfqpoint{2.697819in}{0.844838in}}%
\pgfpathlineto{\pgfqpoint{2.788454in}{0.867192in}}%
\pgfpathlineto{\pgfqpoint{2.833771in}{0.878657in}}%
\pgfpathlineto{\pgfqpoint{2.901748in}{0.899438in}}%
\pgfpathlineto{\pgfqpoint{2.992383in}{0.928768in}}%
\pgfpathlineto{\pgfqpoint{3.037700in}{0.944423in}}%
\pgfpathlineto{\pgfqpoint{3.083018in}{0.961371in}}%
\pgfpathlineto{\pgfqpoint{3.150994in}{0.988414in}}%
\pgfpathlineto{\pgfqpoint{3.264288in}{1.035494in}}%
\pgfpathlineto{\pgfqpoint{3.354923in}{1.075090in}}%
\pgfpathlineto{\pgfqpoint{3.400240in}{1.095744in}}%
\pgfpathlineto{\pgfqpoint{3.445558in}{1.117584in}}%
\pgfpathlineto{\pgfqpoint{3.513534in}{1.151554in}}%
\pgfpathlineto{\pgfqpoint{3.604169in}{1.199220in}}%
\pgfpathlineto{\pgfqpoint{3.694804in}{1.249435in}}%
\pgfpathlineto{\pgfqpoint{3.740121in}{1.275437in}}%
\pgfpathlineto{\pgfqpoint{3.808098in}{1.316103in}}%
\pgfpathlineto{\pgfqpoint{3.898733in}{1.372601in}}%
\pgfpathlineto{\pgfqpoint{3.966709in}{1.416502in}}%
\pgfpathlineto{\pgfqpoint{4.034685in}{1.465495in}}%
\pgfpathlineto{\pgfqpoint{4.102661in}{1.516095in}}%
\pgfpathlineto{\pgfqpoint{4.147979in}{1.551285in}}%
\pgfpathlineto{\pgfqpoint{4.193296in}{1.588547in}}%
\pgfpathlineto{\pgfqpoint{4.261273in}{1.646640in}}%
\pgfpathlineto{\pgfqpoint{4.283931in}{1.666367in}}%
\pgfpathlineto{\pgfqpoint{4.329249in}{1.708949in}}%
\pgfpathlineto{\pgfqpoint{4.397225in}{1.775479in}}%
\pgfpathlineto{\pgfqpoint{4.442543in}{1.821237in}}%
\pgfpathlineto{\pgfqpoint{4.465201in}{1.847768in}}%
\pgfpathlineto{\pgfqpoint{4.487860in}{1.875394in}}%
\pgfpathlineto{\pgfqpoint{4.533178in}{1.935878in}}%
\pgfpathlineto{\pgfqpoint{4.555836in}{1.967285in}}%
\pgfpathlineto{\pgfqpoint{4.601154in}{2.034414in}}%
\pgfpathlineto{\pgfqpoint{4.669130in}{2.152556in}}%
\pgfpathlineto{\pgfqpoint{4.691789in}{2.193433in}}%
\pgfpathlineto{\pgfqpoint{4.759765in}{2.342748in}}%
\pgfpathlineto{\pgfqpoint{4.805083in}{2.444991in}}%
\pgfpathlineto{\pgfqpoint{4.827741in}{2.497597in}}%
\pgfpathlineto{\pgfqpoint{4.873059in}{2.623723in}}%
\pgfpathlineto{\pgfqpoint{4.918376in}{2.757032in}}%
\pgfpathlineto{\pgfqpoint{4.941035in}{2.824860in}}%
\pgfpathlineto{\pgfqpoint{4.986353in}{2.967715in}}%
\pgfpathlineto{\pgfqpoint{5.009011in}{3.043606in}}%
\pgfpathlineto{\pgfqpoint{5.076988in}{3.279874in}}%
\pgfpathlineto{\pgfqpoint{5.099646in}{3.361572in}}%
\pgfpathlineto{\pgfqpoint{5.167623in}{3.617304in}}%
\pgfpathlineto{\pgfqpoint{5.190281in}{3.704560in}}%
\pgfpathlineto{\pgfqpoint{5.258258in}{3.981715in}}%
\pgfpathlineto{\pgfqpoint{5.280916in}{4.075888in}}%
\pgfpathlineto{\pgfqpoint{5.280916in}{4.075888in}}%
\pgfusepath{stroke}%
\end{pgfscope}%
\begin{pgfscope}%
\pgfpathrectangle{\pgfqpoint{0.704982in}{0.547888in}}{\pgfqpoint{4.960000in}{3.696000in}}%
\pgfusepath{clip}%
\pgfsetbuttcap%
\pgfsetroundjoin%
\pgfsetlinewidth{1.505625pt}%
\definecolor{currentstroke}{rgb}{1.000000,0.498039,0.054902}%
\pgfsetstrokecolor{currentstroke}%
\pgfsetdash{{5.550000pt}{2.400000pt}}{0.000000pt}%
\pgfpathmoveto{\pgfqpoint{0.930437in}{0.715888in}}%
\pgfpathlineto{\pgfqpoint{1.406270in}{0.716899in}}%
\pgfpathlineto{\pgfqpoint{1.904763in}{0.719188in}}%
\pgfpathlineto{\pgfqpoint{2.335279in}{0.722519in}}%
\pgfpathlineto{\pgfqpoint{2.743136in}{0.726871in}}%
\pgfpathlineto{\pgfqpoint{3.083018in}{0.731710in}}%
\pgfpathlineto{\pgfqpoint{3.604169in}{0.740962in}}%
\pgfpathlineto{\pgfqpoint{4.034685in}{0.750489in}}%
\pgfpathlineto{\pgfqpoint{4.419884in}{0.760446in}}%
\pgfpathlineto{\pgfqpoint{4.691789in}{0.769032in}}%
\pgfpathlineto{\pgfqpoint{4.941035in}{0.778406in}}%
\pgfpathlineto{\pgfqpoint{5.144964in}{0.787604in}}%
\pgfpathlineto{\pgfqpoint{5.235599in}{0.792827in}}%
\pgfpathlineto{\pgfqpoint{5.303575in}{0.797480in}}%
\pgfpathlineto{\pgfqpoint{5.394210in}{0.805306in}}%
\pgfpathlineto{\pgfqpoint{5.416869in}{0.807491in}}%
\pgfpathlineto{\pgfqpoint{5.439528in}{0.811467in}}%
\pgfpathlineto{\pgfqpoint{5.439528in}{0.811467in}}%
\pgfusepath{stroke}%
\end{pgfscope}%
\begin{pgfscope}%
\pgfpathrectangle{\pgfqpoint{0.704982in}{0.547888in}}{\pgfqpoint{4.960000in}{3.696000in}}%
\pgfusepath{clip}%
\pgfsetbuttcap%
\pgfsetroundjoin%
\pgfsetlinewidth{1.505625pt}%
\definecolor{currentstroke}{rgb}{0.839216,0.152941,0.156863}%
\pgfsetstrokecolor{currentstroke}%
\pgfsetdash{{1.500000pt}{2.475000pt}}{0.000000pt}%
\pgfpathmoveto{\pgfqpoint{0.930437in}{0.715889in}}%
\pgfpathlineto{\pgfqpoint{1.225000in}{0.716749in}}%
\pgfpathlineto{\pgfqpoint{1.451588in}{0.718765in}}%
\pgfpathlineto{\pgfqpoint{1.655517in}{0.722285in}}%
\pgfpathlineto{\pgfqpoint{1.768810in}{0.725594in}}%
\pgfpathlineto{\pgfqpoint{1.882104in}{0.729706in}}%
\pgfpathlineto{\pgfqpoint{1.995398in}{0.735383in}}%
\pgfpathlineto{\pgfqpoint{2.086033in}{0.740931in}}%
\pgfpathlineto{\pgfqpoint{2.199327in}{0.749142in}}%
\pgfpathlineto{\pgfqpoint{2.312620in}{0.759041in}}%
\pgfpathlineto{\pgfqpoint{2.448573in}{0.772144in}}%
\pgfpathlineto{\pgfqpoint{2.561867in}{0.784902in}}%
\pgfpathlineto{\pgfqpoint{2.697819in}{0.801979in}}%
\pgfpathlineto{\pgfqpoint{2.765795in}{0.811519in}}%
\pgfpathlineto{\pgfqpoint{2.856430in}{0.826564in}}%
\pgfpathlineto{\pgfqpoint{2.924406in}{0.839033in}}%
\pgfpathlineto{\pgfqpoint{3.037700in}{0.861717in}}%
\pgfpathlineto{\pgfqpoint{3.150994in}{0.886805in}}%
\pgfpathlineto{\pgfqpoint{3.241629in}{0.908467in}}%
\pgfpathlineto{\pgfqpoint{3.354923in}{0.936606in}}%
\pgfpathlineto{\pgfqpoint{3.400240in}{0.948480in}}%
\pgfpathlineto{\pgfqpoint{3.513534in}{0.981063in}}%
\pgfpathlineto{\pgfqpoint{3.604169in}{1.010001in}}%
\pgfpathlineto{\pgfqpoint{3.672145in}{1.033086in}}%
\pgfpathlineto{\pgfqpoint{3.717463in}{1.049265in}}%
\pgfpathlineto{\pgfqpoint{3.830756in}{1.091288in}}%
\pgfpathlineto{\pgfqpoint{3.898733in}{1.118361in}}%
\pgfpathlineto{\pgfqpoint{3.989368in}{1.157311in}}%
\pgfpathlineto{\pgfqpoint{4.080003in}{1.198067in}}%
\pgfpathlineto{\pgfqpoint{4.147979in}{1.231971in}}%
\pgfpathlineto{\pgfqpoint{4.238614in}{1.278888in}}%
\pgfpathlineto{\pgfqpoint{4.283931in}{1.303491in}}%
\pgfpathlineto{\pgfqpoint{4.329249in}{1.329664in}}%
\pgfpathlineto{\pgfqpoint{4.397225in}{1.374684in}}%
\pgfpathlineto{\pgfqpoint{4.419884in}{1.391450in}}%
\pgfpathlineto{\pgfqpoint{4.442543in}{1.409621in}}%
\pgfpathlineto{\pgfqpoint{4.487860in}{1.447862in}}%
\pgfpathlineto{\pgfqpoint{4.533178in}{1.488882in}}%
\pgfpathlineto{\pgfqpoint{4.578495in}{1.538080in}}%
\pgfpathlineto{\pgfqpoint{4.623813in}{1.590339in}}%
\pgfpathlineto{\pgfqpoint{4.646471in}{1.617547in}}%
\pgfpathlineto{\pgfqpoint{4.691789in}{1.680818in}}%
\pgfpathlineto{\pgfqpoint{4.714448in}{1.714147in}}%
\pgfpathlineto{\pgfqpoint{4.737106in}{1.749353in}}%
\pgfpathlineto{\pgfqpoint{4.827741in}{1.903479in}}%
\pgfpathlineto{\pgfqpoint{4.873059in}{1.983658in}}%
\pgfpathlineto{\pgfqpoint{4.963694in}{2.150369in}}%
\pgfpathlineto{\pgfqpoint{5.009011in}{2.236793in}}%
\pgfpathlineto{\pgfqpoint{5.031670in}{2.281343in}}%
\pgfpathlineto{\pgfqpoint{5.076988in}{2.374358in}}%
\pgfpathlineto{\pgfqpoint{5.122305in}{2.468894in}}%
\pgfpathlineto{\pgfqpoint{5.144964in}{2.518059in}}%
\pgfpathlineto{\pgfqpoint{5.190281in}{2.621971in}}%
\pgfpathlineto{\pgfqpoint{5.212940in}{2.674628in}}%
\pgfpathlineto{\pgfqpoint{5.258258in}{2.787716in}}%
\pgfpathlineto{\pgfqpoint{5.280916in}{2.848793in}}%
\pgfpathlineto{\pgfqpoint{5.326234in}{2.978974in}}%
\pgfpathlineto{\pgfqpoint{5.371551in}{3.129065in}}%
\pgfpathlineto{\pgfqpoint{5.416869in}{3.308548in}}%
\pgfpathlineto{\pgfqpoint{5.416869in}{3.308548in}}%
\pgfusepath{stroke}%
\end{pgfscope}%
\begin{pgfscope}%
\pgfpathrectangle{\pgfqpoint{0.704982in}{0.547888in}}{\pgfqpoint{4.960000in}{3.696000in}}%
\pgfusepath{clip}%
\pgfsetbuttcap%
\pgfsetroundjoin%
\pgfsetlinewidth{1.505625pt}%
\definecolor{currentstroke}{rgb}{0.172549,0.627451,0.172549}%
\pgfsetstrokecolor{currentstroke}%
\pgfsetdash{{9.600000pt}{2.400000pt}{1.500000pt}{2.400000pt}}{0.000000pt}%
\pgfpathmoveto{\pgfqpoint{0.930437in}{0.715888in}}%
\pgfpathlineto{\pgfqpoint{1.496905in}{0.716961in}}%
\pgfpathlineto{\pgfqpoint{2.018057in}{0.719278in}}%
\pgfpathlineto{\pgfqpoint{2.493890in}{0.722644in}}%
\pgfpathlineto{\pgfqpoint{2.992383in}{0.727688in}}%
\pgfpathlineto{\pgfqpoint{3.490875in}{0.734315in}}%
\pgfpathlineto{\pgfqpoint{3.921391in}{0.741498in}}%
\pgfpathlineto{\pgfqpoint{4.419884in}{0.751128in}}%
\pgfpathlineto{\pgfqpoint{4.850400in}{0.761058in}}%
\pgfpathlineto{\pgfqpoint{5.280916in}{0.772246in}}%
\pgfpathlineto{\pgfqpoint{5.439528in}{0.776956in}}%
\pgfpathlineto{\pgfqpoint{5.439528in}{0.776956in}}%
\pgfusepath{stroke}%
\end{pgfscope}%
\begin{pgfscope}%
\pgfsetrectcap%
\pgfsetmiterjoin%
\pgfsetlinewidth{0.803000pt}%
\definecolor{currentstroke}{rgb}{0.000000,0.000000,0.000000}%
\pgfsetstrokecolor{currentstroke}%
\pgfsetdash{}{0pt}%
\pgfpathmoveto{\pgfqpoint{0.704982in}{0.547888in}}%
\pgfpathlineto{\pgfqpoint{0.704982in}{4.243888in}}%
\pgfusepath{stroke}%
\end{pgfscope}%
\begin{pgfscope}%
\pgfsetrectcap%
\pgfsetmiterjoin%
\pgfsetlinewidth{0.803000pt}%
\definecolor{currentstroke}{rgb}{0.000000,0.000000,0.000000}%
\pgfsetstrokecolor{currentstroke}%
\pgfsetdash{}{0pt}%
\pgfpathmoveto{\pgfqpoint{5.664982in}{0.547888in}}%
\pgfpathlineto{\pgfqpoint{5.664982in}{4.243888in}}%
\pgfusepath{stroke}%
\end{pgfscope}%
\begin{pgfscope}%
\pgfsetrectcap%
\pgfsetmiterjoin%
\pgfsetlinewidth{0.803000pt}%
\definecolor{currentstroke}{rgb}{0.000000,0.000000,0.000000}%
\pgfsetstrokecolor{currentstroke}%
\pgfsetdash{}{0pt}%
\pgfpathmoveto{\pgfqpoint{0.704982in}{0.547888in}}%
\pgfpathlineto{\pgfqpoint{5.664982in}{0.547888in}}%
\pgfusepath{stroke}%
\end{pgfscope}%
\begin{pgfscope}%
\pgfsetrectcap%
\pgfsetmiterjoin%
\pgfsetlinewidth{0.803000pt}%
\definecolor{currentstroke}{rgb}{0.000000,0.000000,0.000000}%
\pgfsetstrokecolor{currentstroke}%
\pgfsetdash{}{0pt}%
\pgfpathmoveto{\pgfqpoint{0.704982in}{4.243888in}}%
\pgfpathlineto{\pgfqpoint{5.664982in}{4.243888in}}%
\pgfusepath{stroke}%
\end{pgfscope}%
\begin{pgfscope}%
\pgfsetbuttcap%
\pgfsetmiterjoin%
\definecolor{currentfill}{rgb}{1.000000,1.000000,1.000000}%
\pgfsetfillcolor{currentfill}%
\pgfsetfillopacity{0.800000}%
\pgfsetlinewidth{1.003750pt}%
\definecolor{currentstroke}{rgb}{0.800000,0.800000,0.800000}%
\pgfsetstrokecolor{currentstroke}%
\pgfsetstrokeopacity{0.800000}%
\pgfsetdash{}{0pt}%
\pgfpathmoveto{\pgfqpoint{0.860538in}{2.824112in}}%
\pgfpathlineto{\pgfqpoint{3.311871in}{2.824112in}}%
\pgfpathquadraticcurveto{\pgfqpoint{3.356315in}{2.824112in}}{\pgfqpoint{3.356315in}{2.868556in}}%
\pgfpathlineto{\pgfqpoint{3.356315in}{4.088333in}}%
\pgfpathquadraticcurveto{\pgfqpoint{3.356315in}{4.132777in}}{\pgfqpoint{3.311871in}{4.132777in}}%
\pgfpathlineto{\pgfqpoint{0.860538in}{4.132777in}}%
\pgfpathquadraticcurveto{\pgfqpoint{0.816093in}{4.132777in}}{\pgfqpoint{0.816093in}{4.088333in}}%
\pgfpathlineto{\pgfqpoint{0.816093in}{2.868556in}}%
\pgfpathquadraticcurveto{\pgfqpoint{0.816093in}{2.824112in}}{\pgfqpoint{0.860538in}{2.824112in}}%
\pgfpathclose%
\pgfusepath{stroke,fill}%
\end{pgfscope}%
\begin{pgfscope}%
\pgfsetrectcap%
\pgfsetroundjoin%
\pgfsetlinewidth{1.505625pt}%
\definecolor{currentstroke}{rgb}{0.498039,0.498039,0.498039}%
\pgfsetstrokecolor{currentstroke}%
\pgfsetdash{}{0pt}%
\pgfpathmoveto{\pgfqpoint{0.904982in}{3.966111in}}%
\pgfpathlineto{\pgfqpoint{1.349427in}{3.966111in}}%
\pgfusepath{stroke}%
\end{pgfscope}%
\begin{pgfscope}%
\definecolor{textcolor}{rgb}{0.000000,0.000000,0.000000}%
\pgfsetstrokecolor{textcolor}%
\pgfsetfillcolor{textcolor}%
\pgftext[x=1.527204in,y=3.888333in,left,base]{\color{textcolor}\fontsize{16.000000}{19.200000}\selectfont original encoding}%
\end{pgfscope}%
\begin{pgfscope}%
\pgfsetbuttcap%
\pgfsetroundjoin%
\pgfsetlinewidth{1.505625pt}%
\definecolor{currentstroke}{rgb}{1.000000,0.498039,0.054902}%
\pgfsetstrokecolor{currentstroke}%
\pgfsetdash{{5.550000pt}{2.400000pt}}{0.000000pt}%
\pgfpathmoveto{\pgfqpoint{0.904982in}{3.654111in}}%
\pgfpathlineto{\pgfqpoint{1.349427in}{3.654111in}}%
\pgfusepath{stroke}%
\end{pgfscope}%
\begin{pgfscope}%
\definecolor{textcolor}{rgb}{0.000000,0.000000,0.000000}%
\pgfsetstrokecolor{textcolor}%
\pgfsetfillcolor{textcolor}%
\pgftext[x=1.527204in,y=3.576333in,left,base]{\color{textcolor}\fontsize{16.000000}{19.200000}\selectfont learned at first UIP}%
\end{pgfscope}%
\begin{pgfscope}%
\pgfsetbuttcap%
\pgfsetroundjoin%
\pgfsetlinewidth{1.505625pt}%
\definecolor{currentstroke}{rgb}{0.839216,0.152941,0.156863}%
\pgfsetstrokecolor{currentstroke}%
\pgfsetdash{{1.500000pt}{2.475000pt}}{0.000000pt}%
\pgfpathmoveto{\pgfqpoint{0.904982in}{3.344111in}}%
\pgfpathlineto{\pgfqpoint{1.349427in}{3.344111in}}%
\pgfusepath{stroke}%
\end{pgfscope}%
\begin{pgfscope}%
\definecolor{textcolor}{rgb}{0.000000,0.000000,0.000000}%
\pgfsetstrokecolor{textcolor}%
\pgfsetfillcolor{textcolor}%
\pgftext[x=1.527204in,y=3.266333in,left,base]{\color{textcolor}\fontsize{16.000000}{19.200000}\selectfont learned at last UIP}%
\end{pgfscope}%
\begin{pgfscope}%
\pgfsetbuttcap%
\pgfsetroundjoin%
\pgfsetlinewidth{1.505625pt}%
\definecolor{currentstroke}{rgb}{0.172549,0.627451,0.172549}%
\pgfsetstrokecolor{currentstroke}%
\pgfsetdash{{9.600000pt}{2.400000pt}{1.500000pt}{2.400000pt}}{0.000000pt}%
\pgfpathmoveto{\pgfqpoint{0.904982in}{3.034111in}}%
\pgfpathlineto{\pgfqpoint{1.349427in}{3.034111in}}%
\pgfusepath{stroke}%
\end{pgfscope}%
\begin{pgfscope}%
\definecolor{textcolor}{rgb}{0.000000,0.000000,0.000000}%
\pgfsetstrokecolor{textcolor}%
\pgfsetfillcolor{textcolor}%
\pgftext[x=1.527204in,y=2.956333in,left,base]{\color{textcolor}\fontsize{16.000000}{19.200000}\selectfont reduced constraint}%
\end{pgfscope}%
\end{pgfpicture}%
\makeatother%
\endgroup%

%% file: figures/cactus_hrp_clingo.pgf
%% Creator: Matplotlib, PGF backend
%%
%% To include the figure in your LaTeX document, write
%%   \input{<filename>.pgf}
%%
%% Make sure the required packages are loaded in your preamble
%%   \usepackage{pgf}
%%
%% and, on pdftex
%%   \usepackage[utf8]{inputenc}\DeclareUnicodeCharacter{2212}{-}
%%
%% or, on luatex and xetex
%%   \usepackage{unicode-math}
%%
%% Figures using additional raster images can only be included by \input if
%% they are in the same directory as the main LaTeX file. For loading figures
%% from other directories you can use the `import` package
%%   \usepackage{import}
%%
%% and then include the figures with
%%   \import{<path to file>}{<filename>.pgf}
%%
%% Matplotlib used the following preamble
%%   \usepackage{fontspec}
%%
\begingroup%
\makeatletter%
\begin{pgfpicture}%
\pgfpathrectangle{\pgfpointorigin}{\pgfqpoint{5.554914in}{4.243888in}}%
\pgfusepath{use as bounding box, clip}%
\begin{pgfscope}%
\pgfsetbuttcap%
\pgfsetmiterjoin%
\definecolor{currentfill}{rgb}{1.000000,1.000000,1.000000}%
\pgfsetfillcolor{currentfill}%
\pgfsetlinewidth{0.000000pt}%
\definecolor{currentstroke}{rgb}{1.000000,1.000000,1.000000}%
\pgfsetstrokecolor{currentstroke}%
\pgfsetdash{}{0pt}%
\pgfpathmoveto{\pgfqpoint{0.000000in}{0.000000in}}%
\pgfpathlineto{\pgfqpoint{5.554914in}{0.000000in}}%
\pgfpathlineto{\pgfqpoint{5.554914in}{4.243888in}}%
\pgfpathlineto{\pgfqpoint{0.000000in}{4.243888in}}%
\pgfpathclose%
\pgfusepath{fill}%
\end{pgfscope}%
\begin{pgfscope}%
\pgfsetbuttcap%
\pgfsetmiterjoin%
\definecolor{currentfill}{rgb}{1.000000,1.000000,1.000000}%
\pgfsetfillcolor{currentfill}%
\pgfsetlinewidth{0.000000pt}%
\definecolor{currentstroke}{rgb}{0.000000,0.000000,0.000000}%
\pgfsetstrokecolor{currentstroke}%
\pgfsetstrokeopacity{0.000000}%
\pgfsetdash{}{0pt}%
\pgfpathmoveto{\pgfqpoint{0.594914in}{0.547888in}}%
\pgfpathlineto{\pgfqpoint{5.554914in}{0.547888in}}%
\pgfpathlineto{\pgfqpoint{5.554914in}{4.243888in}}%
\pgfpathlineto{\pgfqpoint{0.594914in}{4.243888in}}%
\pgfpathclose%
\pgfusepath{fill}%
\end{pgfscope}%
\begin{pgfscope}%
\pgfsetbuttcap%
\pgfsetroundjoin%
\definecolor{currentfill}{rgb}{0.000000,0.000000,0.000000}%
\pgfsetfillcolor{currentfill}%
\pgfsetlinewidth{0.803000pt}%
\definecolor{currentstroke}{rgb}{0.000000,0.000000,0.000000}%
\pgfsetstrokecolor{currentstroke}%
\pgfsetdash{}{0pt}%
\pgfsys@defobject{currentmarker}{\pgfqpoint{0.000000in}{-0.048611in}}{\pgfqpoint{0.000000in}{0.000000in}}{%
\pgfpathmoveto{\pgfqpoint{0.000000in}{0.000000in}}%
\pgfpathlineto{\pgfqpoint{0.000000in}{-0.048611in}}%
\pgfusepath{stroke,fill}%
}%
\begin{pgfscope}%
\pgfsys@transformshift{0.698501in}{0.547888in}%
\pgfsys@useobject{currentmarker}{}%
\end{pgfscope}%
\end{pgfscope}%
\begin{pgfscope}%
\definecolor{textcolor}{rgb}{0.000000,0.000000,0.000000}%
\pgfsetstrokecolor{textcolor}%
\pgfsetfillcolor{textcolor}%
\pgftext[x=0.698501in,y=0.450666in,,top]{\color{textcolor}\fontsize{16.000000}{19.200000}\selectfont \(\displaystyle 0\)}%
\end{pgfscope}%
\begin{pgfscope}%
\pgfsetbuttcap%
\pgfsetroundjoin%
\definecolor{currentfill}{rgb}{0.000000,0.000000,0.000000}%
\pgfsetfillcolor{currentfill}%
\pgfsetlinewidth{0.803000pt}%
\definecolor{currentstroke}{rgb}{0.000000,0.000000,0.000000}%
\pgfsetstrokecolor{currentstroke}%
\pgfsetdash{}{0pt}%
\pgfsys@defobject{currentmarker}{\pgfqpoint{0.000000in}{-0.048611in}}{\pgfqpoint{0.000000in}{0.000000in}}{%
\pgfpathmoveto{\pgfqpoint{0.000000in}{0.000000in}}%
\pgfpathlineto{\pgfqpoint{0.000000in}{-0.048611in}}%
\pgfusepath{stroke,fill}%
}%
\begin{pgfscope}%
\pgfsys@transformshift{1.307838in}{0.547888in}%
\pgfsys@useobject{currentmarker}{}%
\end{pgfscope}%
\end{pgfscope}%
\begin{pgfscope}%
\definecolor{textcolor}{rgb}{0.000000,0.000000,0.000000}%
\pgfsetstrokecolor{textcolor}%
\pgfsetfillcolor{textcolor}%
\pgftext[x=1.307838in,y=0.450666in,,top]{\color{textcolor}\fontsize{16.000000}{19.200000}\selectfont \(\displaystyle 5\)}%
\end{pgfscope}%
\begin{pgfscope}%
\pgfsetbuttcap%
\pgfsetroundjoin%
\definecolor{currentfill}{rgb}{0.000000,0.000000,0.000000}%
\pgfsetfillcolor{currentfill}%
\pgfsetlinewidth{0.803000pt}%
\definecolor{currentstroke}{rgb}{0.000000,0.000000,0.000000}%
\pgfsetstrokecolor{currentstroke}%
\pgfsetdash{}{0pt}%
\pgfsys@defobject{currentmarker}{\pgfqpoint{0.000000in}{-0.048611in}}{\pgfqpoint{0.000000in}{0.000000in}}{%
\pgfpathmoveto{\pgfqpoint{0.000000in}{0.000000in}}%
\pgfpathlineto{\pgfqpoint{0.000000in}{-0.048611in}}%
\pgfusepath{stroke,fill}%
}%
\begin{pgfscope}%
\pgfsys@transformshift{1.917174in}{0.547888in}%
\pgfsys@useobject{currentmarker}{}%
\end{pgfscope}%
\end{pgfscope}%
\begin{pgfscope}%
\definecolor{textcolor}{rgb}{0.000000,0.000000,0.000000}%
\pgfsetstrokecolor{textcolor}%
\pgfsetfillcolor{textcolor}%
\pgftext[x=1.917174in,y=0.450666in,,top]{\color{textcolor}\fontsize{16.000000}{19.200000}\selectfont \(\displaystyle 10\)}%
\end{pgfscope}%
\begin{pgfscope}%
\pgfsetbuttcap%
\pgfsetroundjoin%
\definecolor{currentfill}{rgb}{0.000000,0.000000,0.000000}%
\pgfsetfillcolor{currentfill}%
\pgfsetlinewidth{0.803000pt}%
\definecolor{currentstroke}{rgb}{0.000000,0.000000,0.000000}%
\pgfsetstrokecolor{currentstroke}%
\pgfsetdash{}{0pt}%
\pgfsys@defobject{currentmarker}{\pgfqpoint{0.000000in}{-0.048611in}}{\pgfqpoint{0.000000in}{0.000000in}}{%
\pgfpathmoveto{\pgfqpoint{0.000000in}{0.000000in}}%
\pgfpathlineto{\pgfqpoint{0.000000in}{-0.048611in}}%
\pgfusepath{stroke,fill}%
}%
\begin{pgfscope}%
\pgfsys@transformshift{2.526511in}{0.547888in}%
\pgfsys@useobject{currentmarker}{}%
\end{pgfscope}%
\end{pgfscope}%
\begin{pgfscope}%
\definecolor{textcolor}{rgb}{0.000000,0.000000,0.000000}%
\pgfsetstrokecolor{textcolor}%
\pgfsetfillcolor{textcolor}%
\pgftext[x=2.526511in,y=0.450666in,,top]{\color{textcolor}\fontsize{16.000000}{19.200000}\selectfont \(\displaystyle 15\)}%
\end{pgfscope}%
\begin{pgfscope}%
\pgfsetbuttcap%
\pgfsetroundjoin%
\definecolor{currentfill}{rgb}{0.000000,0.000000,0.000000}%
\pgfsetfillcolor{currentfill}%
\pgfsetlinewidth{0.803000pt}%
\definecolor{currentstroke}{rgb}{0.000000,0.000000,0.000000}%
\pgfsetstrokecolor{currentstroke}%
\pgfsetdash{}{0pt}%
\pgfsys@defobject{currentmarker}{\pgfqpoint{0.000000in}{-0.048611in}}{\pgfqpoint{0.000000in}{0.000000in}}{%
\pgfpathmoveto{\pgfqpoint{0.000000in}{0.000000in}}%
\pgfpathlineto{\pgfqpoint{0.000000in}{-0.048611in}}%
\pgfusepath{stroke,fill}%
}%
\begin{pgfscope}%
\pgfsys@transformshift{3.135848in}{0.547888in}%
\pgfsys@useobject{currentmarker}{}%
\end{pgfscope}%
\end{pgfscope}%
\begin{pgfscope}%
\definecolor{textcolor}{rgb}{0.000000,0.000000,0.000000}%
\pgfsetstrokecolor{textcolor}%
\pgfsetfillcolor{textcolor}%
\pgftext[x=3.135848in,y=0.450666in,,top]{\color{textcolor}\fontsize{16.000000}{19.200000}\selectfont \(\displaystyle 20\)}%
\end{pgfscope}%
\begin{pgfscope}%
\pgfsetbuttcap%
\pgfsetroundjoin%
\definecolor{currentfill}{rgb}{0.000000,0.000000,0.000000}%
\pgfsetfillcolor{currentfill}%
\pgfsetlinewidth{0.803000pt}%
\definecolor{currentstroke}{rgb}{0.000000,0.000000,0.000000}%
\pgfsetstrokecolor{currentstroke}%
\pgfsetdash{}{0pt}%
\pgfsys@defobject{currentmarker}{\pgfqpoint{0.000000in}{-0.048611in}}{\pgfqpoint{0.000000in}{0.000000in}}{%
\pgfpathmoveto{\pgfqpoint{0.000000in}{0.000000in}}%
\pgfpathlineto{\pgfqpoint{0.000000in}{-0.048611in}}%
\pgfusepath{stroke,fill}%
}%
\begin{pgfscope}%
\pgfsys@transformshift{3.745184in}{0.547888in}%
\pgfsys@useobject{currentmarker}{}%
\end{pgfscope}%
\end{pgfscope}%
\begin{pgfscope}%
\definecolor{textcolor}{rgb}{0.000000,0.000000,0.000000}%
\pgfsetstrokecolor{textcolor}%
\pgfsetfillcolor{textcolor}%
\pgftext[x=3.745184in,y=0.450666in,,top]{\color{textcolor}\fontsize{16.000000}{19.200000}\selectfont \(\displaystyle 25\)}%
\end{pgfscope}%
\begin{pgfscope}%
\pgfsetbuttcap%
\pgfsetroundjoin%
\definecolor{currentfill}{rgb}{0.000000,0.000000,0.000000}%
\pgfsetfillcolor{currentfill}%
\pgfsetlinewidth{0.803000pt}%
\definecolor{currentstroke}{rgb}{0.000000,0.000000,0.000000}%
\pgfsetstrokecolor{currentstroke}%
\pgfsetdash{}{0pt}%
\pgfsys@defobject{currentmarker}{\pgfqpoint{0.000000in}{-0.048611in}}{\pgfqpoint{0.000000in}{0.000000in}}{%
\pgfpathmoveto{\pgfqpoint{0.000000in}{0.000000in}}%
\pgfpathlineto{\pgfqpoint{0.000000in}{-0.048611in}}%
\pgfusepath{stroke,fill}%
}%
\begin{pgfscope}%
\pgfsys@transformshift{4.354521in}{0.547888in}%
\pgfsys@useobject{currentmarker}{}%
\end{pgfscope}%
\end{pgfscope}%
\begin{pgfscope}%
\definecolor{textcolor}{rgb}{0.000000,0.000000,0.000000}%
\pgfsetstrokecolor{textcolor}%
\pgfsetfillcolor{textcolor}%
\pgftext[x=4.354521in,y=0.450666in,,top]{\color{textcolor}\fontsize{16.000000}{19.200000}\selectfont \(\displaystyle 30\)}%
\end{pgfscope}%
\begin{pgfscope}%
\pgfsetbuttcap%
\pgfsetroundjoin%
\definecolor{currentfill}{rgb}{0.000000,0.000000,0.000000}%
\pgfsetfillcolor{currentfill}%
\pgfsetlinewidth{0.803000pt}%
\definecolor{currentstroke}{rgb}{0.000000,0.000000,0.000000}%
\pgfsetstrokecolor{currentstroke}%
\pgfsetdash{}{0pt}%
\pgfsys@defobject{currentmarker}{\pgfqpoint{0.000000in}{-0.048611in}}{\pgfqpoint{0.000000in}{0.000000in}}{%
\pgfpathmoveto{\pgfqpoint{0.000000in}{0.000000in}}%
\pgfpathlineto{\pgfqpoint{0.000000in}{-0.048611in}}%
\pgfusepath{stroke,fill}%
}%
\begin{pgfscope}%
\pgfsys@transformshift{4.963857in}{0.547888in}%
\pgfsys@useobject{currentmarker}{}%
\end{pgfscope}%
\end{pgfscope}%
\begin{pgfscope}%
\definecolor{textcolor}{rgb}{0.000000,0.000000,0.000000}%
\pgfsetstrokecolor{textcolor}%
\pgfsetfillcolor{textcolor}%
\pgftext[x=4.963857in,y=0.450666in,,top]{\color{textcolor}\fontsize{16.000000}{19.200000}\selectfont \(\displaystyle 35\)}%
\end{pgfscope}%
\begin{pgfscope}%
\definecolor{textcolor}{rgb}{0.000000,0.000000,0.000000}%
\pgfsetstrokecolor{textcolor}%
\pgfsetfillcolor{textcolor}%
\pgftext[x=3.074914in,y=0.197555in,,top]{\color{textcolor}\fontsize{16.000000}{19.200000}\selectfont Number of instances}%
\end{pgfscope}%
\begin{pgfscope}%
\pgfsetbuttcap%
\pgfsetroundjoin%
\definecolor{currentfill}{rgb}{0.000000,0.000000,0.000000}%
\pgfsetfillcolor{currentfill}%
\pgfsetlinewidth{0.803000pt}%
\definecolor{currentstroke}{rgb}{0.000000,0.000000,0.000000}%
\pgfsetstrokecolor{currentstroke}%
\pgfsetdash{}{0pt}%
\pgfsys@defobject{currentmarker}{\pgfqpoint{-0.048611in}{0.000000in}}{\pgfqpoint{0.000000in}{0.000000in}}{%
\pgfpathmoveto{\pgfqpoint{0.000000in}{0.000000in}}%
\pgfpathlineto{\pgfqpoint{-0.048611in}{0.000000in}}%
\pgfusepath{stroke,fill}%
}%
\begin{pgfscope}%
\pgfsys@transformshift{0.594914in}{0.715797in}%
\pgfsys@useobject{currentmarker}{}%
\end{pgfscope}%
\end{pgfscope}%
\begin{pgfscope}%
\definecolor{textcolor}{rgb}{0.000000,0.000000,0.000000}%
\pgfsetstrokecolor{textcolor}%
\pgfsetfillcolor{textcolor}%
\pgftext[x=0.387623in, y=0.638686in, left, base]{\color{textcolor}\fontsize{16.000000}{19.200000}\selectfont \(\displaystyle 0\)}%
\end{pgfscope}%
\begin{pgfscope}%
\pgfsetbuttcap%
\pgfsetroundjoin%
\definecolor{currentfill}{rgb}{0.000000,0.000000,0.000000}%
\pgfsetfillcolor{currentfill}%
\pgfsetlinewidth{0.803000pt}%
\definecolor{currentstroke}{rgb}{0.000000,0.000000,0.000000}%
\pgfsetstrokecolor{currentstroke}%
\pgfsetdash{}{0pt}%
\pgfsys@defobject{currentmarker}{\pgfqpoint{-0.048611in}{0.000000in}}{\pgfqpoint{0.000000in}{0.000000in}}{%
\pgfpathmoveto{\pgfqpoint{0.000000in}{0.000000in}}%
\pgfpathlineto{\pgfqpoint{-0.048611in}{0.000000in}}%
\pgfusepath{stroke,fill}%
}%
\begin{pgfscope}%
\pgfsys@transformshift{0.594914in}{1.137996in}%
\pgfsys@useobject{currentmarker}{}%
\end{pgfscope}%
\end{pgfscope}%
\begin{pgfscope}%
\definecolor{textcolor}{rgb}{0.000000,0.000000,0.000000}%
\pgfsetstrokecolor{textcolor}%
\pgfsetfillcolor{textcolor}%
\pgftext[x=0.387623in, y=1.060885in, left, base]{\color{textcolor}\fontsize{16.000000}{19.200000}\selectfont \(\displaystyle 8\)}%
\end{pgfscope}%
\begin{pgfscope}%
\pgfsetbuttcap%
\pgfsetroundjoin%
\definecolor{currentfill}{rgb}{0.000000,0.000000,0.000000}%
\pgfsetfillcolor{currentfill}%
\pgfsetlinewidth{0.803000pt}%
\definecolor{currentstroke}{rgb}{0.000000,0.000000,0.000000}%
\pgfsetstrokecolor{currentstroke}%
\pgfsetdash{}{0pt}%
\pgfsys@defobject{currentmarker}{\pgfqpoint{-0.048611in}{0.000000in}}{\pgfqpoint{0.000000in}{0.000000in}}{%
\pgfpathmoveto{\pgfqpoint{0.000000in}{0.000000in}}%
\pgfpathlineto{\pgfqpoint{-0.048611in}{0.000000in}}%
\pgfusepath{stroke,fill}%
}%
\begin{pgfscope}%
\pgfsys@transformshift{0.594914in}{1.560196in}%
\pgfsys@useobject{currentmarker}{}%
\end{pgfscope}%
\end{pgfscope}%
\begin{pgfscope}%
\definecolor{textcolor}{rgb}{0.000000,0.000000,0.000000}%
\pgfsetstrokecolor{textcolor}%
\pgfsetfillcolor{textcolor}%
\pgftext[x=0.277555in, y=1.483085in, left, base]{\color{textcolor}\fontsize{16.000000}{19.200000}\selectfont \(\displaystyle 16\)}%
\end{pgfscope}%
\begin{pgfscope}%
\pgfsetbuttcap%
\pgfsetroundjoin%
\definecolor{currentfill}{rgb}{0.000000,0.000000,0.000000}%
\pgfsetfillcolor{currentfill}%
\pgfsetlinewidth{0.803000pt}%
\definecolor{currentstroke}{rgb}{0.000000,0.000000,0.000000}%
\pgfsetstrokecolor{currentstroke}%
\pgfsetdash{}{0pt}%
\pgfsys@defobject{currentmarker}{\pgfqpoint{-0.048611in}{0.000000in}}{\pgfqpoint{0.000000in}{0.000000in}}{%
\pgfpathmoveto{\pgfqpoint{0.000000in}{0.000000in}}%
\pgfpathlineto{\pgfqpoint{-0.048611in}{0.000000in}}%
\pgfusepath{stroke,fill}%
}%
\begin{pgfscope}%
\pgfsys@transformshift{0.594914in}{1.982396in}%
\pgfsys@useobject{currentmarker}{}%
\end{pgfscope}%
\end{pgfscope}%
\begin{pgfscope}%
\definecolor{textcolor}{rgb}{0.000000,0.000000,0.000000}%
\pgfsetstrokecolor{textcolor}%
\pgfsetfillcolor{textcolor}%
\pgftext[x=0.277555in, y=1.905285in, left, base]{\color{textcolor}\fontsize{16.000000}{19.200000}\selectfont \(\displaystyle 24\)}%
\end{pgfscope}%
\begin{pgfscope}%
\pgfsetbuttcap%
\pgfsetroundjoin%
\definecolor{currentfill}{rgb}{0.000000,0.000000,0.000000}%
\pgfsetfillcolor{currentfill}%
\pgfsetlinewidth{0.803000pt}%
\definecolor{currentstroke}{rgb}{0.000000,0.000000,0.000000}%
\pgfsetstrokecolor{currentstroke}%
\pgfsetdash{}{0pt}%
\pgfsys@defobject{currentmarker}{\pgfqpoint{-0.048611in}{0.000000in}}{\pgfqpoint{0.000000in}{0.000000in}}{%
\pgfpathmoveto{\pgfqpoint{0.000000in}{0.000000in}}%
\pgfpathlineto{\pgfqpoint{-0.048611in}{0.000000in}}%
\pgfusepath{stroke,fill}%
}%
\begin{pgfscope}%
\pgfsys@transformshift{0.594914in}{2.404595in}%
\pgfsys@useobject{currentmarker}{}%
\end{pgfscope}%
\end{pgfscope}%
\begin{pgfscope}%
\definecolor{textcolor}{rgb}{0.000000,0.000000,0.000000}%
\pgfsetstrokecolor{textcolor}%
\pgfsetfillcolor{textcolor}%
\pgftext[x=0.277555in, y=2.327484in, left, base]{\color{textcolor}\fontsize{16.000000}{19.200000}\selectfont \(\displaystyle 32\)}%
\end{pgfscope}%
\begin{pgfscope}%
\pgfsetbuttcap%
\pgfsetroundjoin%
\definecolor{currentfill}{rgb}{0.000000,0.000000,0.000000}%
\pgfsetfillcolor{currentfill}%
\pgfsetlinewidth{0.803000pt}%
\definecolor{currentstroke}{rgb}{0.000000,0.000000,0.000000}%
\pgfsetstrokecolor{currentstroke}%
\pgfsetdash{}{0pt}%
\pgfsys@defobject{currentmarker}{\pgfqpoint{-0.048611in}{0.000000in}}{\pgfqpoint{0.000000in}{0.000000in}}{%
\pgfpathmoveto{\pgfqpoint{0.000000in}{0.000000in}}%
\pgfpathlineto{\pgfqpoint{-0.048611in}{0.000000in}}%
\pgfusepath{stroke,fill}%
}%
\begin{pgfscope}%
\pgfsys@transformshift{0.594914in}{2.826795in}%
\pgfsys@useobject{currentmarker}{}%
\end{pgfscope}%
\end{pgfscope}%
\begin{pgfscope}%
\definecolor{textcolor}{rgb}{0.000000,0.000000,0.000000}%
\pgfsetstrokecolor{textcolor}%
\pgfsetfillcolor{textcolor}%
\pgftext[x=0.277555in, y=2.749684in, left, base]{\color{textcolor}\fontsize{16.000000}{19.200000}\selectfont \(\displaystyle 40\)}%
\end{pgfscope}%
\begin{pgfscope}%
\pgfsetbuttcap%
\pgfsetroundjoin%
\definecolor{currentfill}{rgb}{0.000000,0.000000,0.000000}%
\pgfsetfillcolor{currentfill}%
\pgfsetlinewidth{0.803000pt}%
\definecolor{currentstroke}{rgb}{0.000000,0.000000,0.000000}%
\pgfsetstrokecolor{currentstroke}%
\pgfsetdash{}{0pt}%
\pgfsys@defobject{currentmarker}{\pgfqpoint{-0.048611in}{0.000000in}}{\pgfqpoint{0.000000in}{0.000000in}}{%
\pgfpathmoveto{\pgfqpoint{0.000000in}{0.000000in}}%
\pgfpathlineto{\pgfqpoint{-0.048611in}{0.000000in}}%
\pgfusepath{stroke,fill}%
}%
\begin{pgfscope}%
\pgfsys@transformshift{0.594914in}{3.248995in}%
\pgfsys@useobject{currentmarker}{}%
\end{pgfscope}%
\end{pgfscope}%
\begin{pgfscope}%
\definecolor{textcolor}{rgb}{0.000000,0.000000,0.000000}%
\pgfsetstrokecolor{textcolor}%
\pgfsetfillcolor{textcolor}%
\pgftext[x=0.277555in, y=3.171883in, left, base]{\color{textcolor}\fontsize{16.000000}{19.200000}\selectfont \(\displaystyle 48\)}%
\end{pgfscope}%
\begin{pgfscope}%
\pgfsetbuttcap%
\pgfsetroundjoin%
\definecolor{currentfill}{rgb}{0.000000,0.000000,0.000000}%
\pgfsetfillcolor{currentfill}%
\pgfsetlinewidth{0.803000pt}%
\definecolor{currentstroke}{rgb}{0.000000,0.000000,0.000000}%
\pgfsetstrokecolor{currentstroke}%
\pgfsetdash{}{0pt}%
\pgfsys@defobject{currentmarker}{\pgfqpoint{-0.048611in}{0.000000in}}{\pgfqpoint{0.000000in}{0.000000in}}{%
\pgfpathmoveto{\pgfqpoint{0.000000in}{0.000000in}}%
\pgfpathlineto{\pgfqpoint{-0.048611in}{0.000000in}}%
\pgfusepath{stroke,fill}%
}%
\begin{pgfscope}%
\pgfsys@transformshift{0.594914in}{3.671194in}%
\pgfsys@useobject{currentmarker}{}%
\end{pgfscope}%
\end{pgfscope}%
\begin{pgfscope}%
\definecolor{textcolor}{rgb}{0.000000,0.000000,0.000000}%
\pgfsetstrokecolor{textcolor}%
\pgfsetfillcolor{textcolor}%
\pgftext[x=0.277555in, y=3.594083in, left, base]{\color{textcolor}\fontsize{16.000000}{19.200000}\selectfont \(\displaystyle 56\)}%
\end{pgfscope}%
\begin{pgfscope}%
\pgfsetbuttcap%
\pgfsetroundjoin%
\definecolor{currentfill}{rgb}{0.000000,0.000000,0.000000}%
\pgfsetfillcolor{currentfill}%
\pgfsetlinewidth{0.803000pt}%
\definecolor{currentstroke}{rgb}{0.000000,0.000000,0.000000}%
\pgfsetstrokecolor{currentstroke}%
\pgfsetdash{}{0pt}%
\pgfsys@defobject{currentmarker}{\pgfqpoint{-0.048611in}{0.000000in}}{\pgfqpoint{0.000000in}{0.000000in}}{%
\pgfpathmoveto{\pgfqpoint{0.000000in}{0.000000in}}%
\pgfpathlineto{\pgfqpoint{-0.048611in}{0.000000in}}%
\pgfusepath{stroke,fill}%
}%
\begin{pgfscope}%
\pgfsys@transformshift{0.594914in}{4.093394in}%
\pgfsys@useobject{currentmarker}{}%
\end{pgfscope}%
\end{pgfscope}%
\begin{pgfscope}%
\definecolor{textcolor}{rgb}{0.000000,0.000000,0.000000}%
\pgfsetstrokecolor{textcolor}%
\pgfsetfillcolor{textcolor}%
\pgftext[x=0.277555in, y=4.016283in, left, base]{\color{textcolor}\fontsize{16.000000}{19.200000}\selectfont \(\displaystyle 64\)}%
\end{pgfscope}%
\begin{pgfscope}%
\definecolor{textcolor}{rgb}{0.000000,0.000000,0.000000}%
\pgfsetstrokecolor{textcolor}%
\pgfsetfillcolor{textcolor}%
\pgftext[x=0.222000in,y=2.395888in,,bottom,rotate=90.000000]{\color{textcolor}\fontsize{16.000000}{19.200000}\selectfont Real time consumption (minutes)}%
\end{pgfscope}%
\begin{pgfscope}%
\pgfpathrectangle{\pgfqpoint{0.594914in}{0.547888in}}{\pgfqpoint{4.960000in}{3.696000in}}%
\pgfusepath{clip}%
\pgfsetrectcap%
\pgfsetroundjoin%
\pgfsetlinewidth{1.505625pt}%
\definecolor{currentstroke}{rgb}{0.498039,0.498039,0.498039}%
\pgfsetstrokecolor{currentstroke}%
\pgfsetdash{}{0pt}%
\pgfpathmoveto{\pgfqpoint{0.820368in}{0.715888in}}%
\pgfpathlineto{\pgfqpoint{0.942236in}{0.715982in}}%
\pgfpathlineto{\pgfqpoint{1.064103in}{0.716173in}}%
\pgfpathlineto{\pgfqpoint{1.185970in}{0.716462in}}%
\pgfpathlineto{\pgfqpoint{1.307838in}{0.717348in}}%
\pgfpathlineto{\pgfqpoint{1.429705in}{0.718237in}}%
\pgfpathlineto{\pgfqpoint{1.551572in}{0.720638in}}%
\pgfpathlineto{\pgfqpoint{1.673440in}{0.723980in}}%
\pgfpathlineto{\pgfqpoint{1.795307in}{0.728096in}}%
\pgfpathlineto{\pgfqpoint{1.917174in}{0.733230in}}%
\pgfpathlineto{\pgfqpoint{2.039042in}{0.743418in}}%
\pgfpathlineto{\pgfqpoint{2.160909in}{0.754127in}}%
\pgfpathlineto{\pgfqpoint{2.282776in}{0.767072in}}%
\pgfpathlineto{\pgfqpoint{2.404644in}{0.784174in}}%
\pgfpathlineto{\pgfqpoint{2.526511in}{0.806354in}}%
\pgfpathlineto{\pgfqpoint{2.648378in}{0.832079in}}%
\pgfpathlineto{\pgfqpoint{2.770246in}{0.863263in}}%
\pgfpathlineto{\pgfqpoint{2.892113in}{0.901939in}}%
\pgfpathlineto{\pgfqpoint{3.013980in}{0.950443in}}%
\pgfpathlineto{\pgfqpoint{3.135848in}{1.000114in}}%
\pgfpathlineto{\pgfqpoint{3.257715in}{1.052680in}}%
\pgfpathlineto{\pgfqpoint{3.379582in}{1.117383in}}%
\pgfpathlineto{\pgfqpoint{3.501450in}{1.228148in}}%
\pgfpathlineto{\pgfqpoint{3.623317in}{1.350928in}}%
\pgfpathlineto{\pgfqpoint{3.745184in}{1.482940in}}%
\pgfpathlineto{\pgfqpoint{3.867051in}{1.646042in}}%
\pgfpathlineto{\pgfqpoint{3.988919in}{1.831083in}}%
\pgfpathlineto{\pgfqpoint{4.110786in}{2.076473in}}%
\pgfpathlineto{\pgfqpoint{4.232653in}{2.349395in}}%
\pgfpathlineto{\pgfqpoint{4.354521in}{2.637104in}}%
\pgfpathlineto{\pgfqpoint{4.476388in}{2.927090in}}%
\pgfpathlineto{\pgfqpoint{4.598255in}{3.262829in}}%
\pgfpathlineto{\pgfqpoint{4.720123in}{3.611777in}}%
\pgfpathlineto{\pgfqpoint{4.841990in}{4.072898in}}%
\pgfusepath{stroke}%
\end{pgfscope}%
\begin{pgfscope}%
\pgfpathrectangle{\pgfqpoint{0.594914in}{0.547888in}}{\pgfqpoint{4.960000in}{3.696000in}}%
\pgfusepath{clip}%
\pgfsetbuttcap%
\pgfsetroundjoin%
\pgfsetlinewidth{1.505625pt}%
\definecolor{currentstroke}{rgb}{1.000000,0.498039,0.054902}%
\pgfsetstrokecolor{currentstroke}%
\pgfsetdash{{5.550000pt}{2.400000pt}}{0.000000pt}%
\pgfpathmoveto{\pgfqpoint{0.820368in}{0.716005in}}%
\pgfpathlineto{\pgfqpoint{0.942236in}{0.716303in}}%
\pgfpathlineto{\pgfqpoint{1.064103in}{0.716703in}}%
\pgfpathlineto{\pgfqpoint{1.185970in}{0.717209in}}%
\pgfpathlineto{\pgfqpoint{1.307838in}{0.718904in}}%
\pgfpathlineto{\pgfqpoint{1.429705in}{0.723474in}}%
\pgfpathlineto{\pgfqpoint{1.551572in}{0.729649in}}%
\pgfpathlineto{\pgfqpoint{1.673440in}{0.737710in}}%
\pgfpathlineto{\pgfqpoint{1.795307in}{0.748501in}}%
\pgfpathlineto{\pgfqpoint{1.917174in}{0.768058in}}%
\pgfpathlineto{\pgfqpoint{2.039042in}{0.792271in}}%
\pgfpathlineto{\pgfqpoint{2.160909in}{0.822827in}}%
\pgfpathlineto{\pgfqpoint{2.282776in}{0.862761in}}%
\pgfpathlineto{\pgfqpoint{2.404644in}{0.922387in}}%
\pgfpathlineto{\pgfqpoint{2.526511in}{1.008189in}}%
\pgfpathlineto{\pgfqpoint{2.648378in}{1.102237in}}%
\pgfpathlineto{\pgfqpoint{2.770246in}{1.210992in}}%
\pgfpathlineto{\pgfqpoint{2.892113in}{1.364962in}}%
\pgfpathlineto{\pgfqpoint{3.013980in}{1.592097in}}%
\pgfpathlineto{\pgfqpoint{3.135848in}{1.835326in}}%
\pgfpathlineto{\pgfqpoint{3.257715in}{2.090998in}}%
\pgfpathlineto{\pgfqpoint{3.379582in}{2.457166in}}%
\pgfusepath{stroke}%
\end{pgfscope}%
\begin{pgfscope}%
\pgfpathrectangle{\pgfqpoint{0.594914in}{0.547888in}}{\pgfqpoint{4.960000in}{3.696000in}}%
\pgfusepath{clip}%
\pgfsetbuttcap%
\pgfsetroundjoin%
\pgfsetlinewidth{1.505625pt}%
\definecolor{currentstroke}{rgb}{0.839216,0.152941,0.156863}%
\pgfsetstrokecolor{currentstroke}%
\pgfsetdash{{1.500000pt}{2.475000pt}}{0.000000pt}%
\pgfpathmoveto{\pgfqpoint{0.820368in}{0.716093in}}%
\pgfpathlineto{\pgfqpoint{0.942236in}{0.716391in}}%
\pgfpathlineto{\pgfqpoint{1.064103in}{0.716794in}}%
\pgfpathlineto{\pgfqpoint{1.185970in}{0.717297in}}%
\pgfpathlineto{\pgfqpoint{1.307838in}{0.719549in}}%
\pgfpathlineto{\pgfqpoint{1.429705in}{0.724357in}}%
\pgfpathlineto{\pgfqpoint{1.551572in}{0.730932in}}%
\pgfpathlineto{\pgfqpoint{1.673440in}{0.738512in}}%
\pgfpathlineto{\pgfqpoint{1.795307in}{0.750228in}}%
\pgfpathlineto{\pgfqpoint{1.917174in}{0.770720in}}%
\pgfpathlineto{\pgfqpoint{2.039042in}{0.798565in}}%
\pgfpathlineto{\pgfqpoint{2.160909in}{0.833806in}}%
\pgfpathlineto{\pgfqpoint{2.282776in}{0.874724in}}%
\pgfpathlineto{\pgfqpoint{2.404644in}{0.941873in}}%
\pgfpathlineto{\pgfqpoint{2.526511in}{1.027451in}}%
\pgfpathlineto{\pgfqpoint{2.648378in}{1.125163in}}%
\pgfpathlineto{\pgfqpoint{2.770246in}{1.244872in}}%
\pgfpathlineto{\pgfqpoint{2.892113in}{1.366587in}}%
\pgfpathlineto{\pgfqpoint{3.013980in}{1.619254in}}%
\pgfpathlineto{\pgfqpoint{3.135848in}{1.878364in}}%
\pgfpathlineto{\pgfqpoint{3.257715in}{2.151672in}}%
\pgfpathlineto{\pgfqpoint{3.379582in}{2.472186in}}%
\pgfusepath{stroke}%
\end{pgfscope}%
\begin{pgfscope}%
\pgfpathrectangle{\pgfqpoint{0.594914in}{0.547888in}}{\pgfqpoint{4.960000in}{3.696000in}}%
\pgfusepath{clip}%
\pgfsetbuttcap%
\pgfsetroundjoin%
\pgfsetlinewidth{1.505625pt}%
\definecolor{currentstroke}{rgb}{0.172549,0.627451,0.172549}%
\pgfsetstrokecolor{currentstroke}%
\pgfsetdash{{9.600000pt}{2.400000pt}{1.500000pt}{2.400000pt}}{0.000000pt}%
\pgfpathmoveto{\pgfqpoint{0.820368in}{0.715890in}}%
\pgfpathlineto{\pgfqpoint{0.942236in}{0.716081in}}%
\pgfpathlineto{\pgfqpoint{1.064103in}{0.716276in}}%
\pgfpathlineto{\pgfqpoint{1.185970in}{0.716566in}}%
\pgfpathlineto{\pgfqpoint{1.307838in}{0.717280in}}%
\pgfpathlineto{\pgfqpoint{1.429705in}{0.718166in}}%
\pgfpathlineto{\pgfqpoint{1.551572in}{0.719809in}}%
\pgfpathlineto{\pgfqpoint{1.673440in}{0.722212in}}%
\pgfpathlineto{\pgfqpoint{1.795307in}{0.724873in}}%
\pgfpathlineto{\pgfqpoint{1.917174in}{0.728611in}}%
\pgfpathlineto{\pgfqpoint{2.039042in}{0.736046in}}%
\pgfpathlineto{\pgfqpoint{2.160909in}{0.743968in}}%
\pgfpathlineto{\pgfqpoint{2.282776in}{0.752297in}}%
\pgfpathlineto{\pgfqpoint{2.404644in}{0.765212in}}%
\pgfpathlineto{\pgfqpoint{2.526511in}{0.778209in}}%
\pgfpathlineto{\pgfqpoint{2.648378in}{0.792539in}}%
\pgfpathlineto{\pgfqpoint{2.770246in}{0.809609in}}%
\pgfpathlineto{\pgfqpoint{2.892113in}{0.834430in}}%
\pgfpathlineto{\pgfqpoint{3.013980in}{0.860346in}}%
\pgfpathlineto{\pgfqpoint{3.135848in}{0.888772in}}%
\pgfpathlineto{\pgfqpoint{3.257715in}{0.919051in}}%
\pgfpathlineto{\pgfqpoint{3.379582in}{0.970470in}}%
\pgfpathlineto{\pgfqpoint{3.501450in}{1.025568in}}%
\pgfpathlineto{\pgfqpoint{3.623317in}{1.098224in}}%
\pgfpathlineto{\pgfqpoint{3.745184in}{1.172773in}}%
\pgfpathlineto{\pgfqpoint{3.867051in}{1.256555in}}%
\pgfpathlineto{\pgfqpoint{3.988919in}{1.344387in}}%
\pgfpathlineto{\pgfqpoint{4.110786in}{1.454224in}}%
\pgfpathlineto{\pgfqpoint{4.232653in}{1.592337in}}%
\pgfpathlineto{\pgfqpoint{4.354521in}{1.738533in}}%
\pgfpathlineto{\pgfqpoint{4.476388in}{1.914103in}}%
\pgfpathlineto{\pgfqpoint{4.598255in}{2.126943in}}%
\pgfpathlineto{\pgfqpoint{4.720123in}{2.361938in}}%
\pgfpathlineto{\pgfqpoint{4.841990in}{2.631349in}}%
\pgfpathlineto{\pgfqpoint{4.963857in}{2.911538in}}%
\pgfpathlineto{\pgfqpoint{5.085725in}{3.215205in}}%
\pgfpathlineto{\pgfqpoint{5.207592in}{3.580801in}}%
\pgfpathlineto{\pgfqpoint{5.329459in}{4.075888in}}%
\pgfusepath{stroke}%
\end{pgfscope}%
\begin{pgfscope}%
\pgfsetrectcap%
\pgfsetmiterjoin%
\pgfsetlinewidth{0.803000pt}%
\definecolor{currentstroke}{rgb}{0.000000,0.000000,0.000000}%
\pgfsetstrokecolor{currentstroke}%
\pgfsetdash{}{0pt}%
\pgfpathmoveto{\pgfqpoint{0.594914in}{0.547888in}}%
\pgfpathlineto{\pgfqpoint{0.594914in}{4.243888in}}%
\pgfusepath{stroke}%
\end{pgfscope}%
\begin{pgfscope}%
\pgfsetrectcap%
\pgfsetmiterjoin%
\pgfsetlinewidth{0.803000pt}%
\definecolor{currentstroke}{rgb}{0.000000,0.000000,0.000000}%
\pgfsetstrokecolor{currentstroke}%
\pgfsetdash{}{0pt}%
\pgfpathmoveto{\pgfqpoint{5.554914in}{0.547888in}}%
\pgfpathlineto{\pgfqpoint{5.554914in}{4.243888in}}%
\pgfusepath{stroke}%
\end{pgfscope}%
\begin{pgfscope}%
\pgfsetrectcap%
\pgfsetmiterjoin%
\pgfsetlinewidth{0.803000pt}%
\definecolor{currentstroke}{rgb}{0.000000,0.000000,0.000000}%
\pgfsetstrokecolor{currentstroke}%
\pgfsetdash{}{0pt}%
\pgfpathmoveto{\pgfqpoint{0.594914in}{0.547888in}}%
\pgfpathlineto{\pgfqpoint{5.554914in}{0.547888in}}%
\pgfusepath{stroke}%
\end{pgfscope}%
\begin{pgfscope}%
\pgfsetrectcap%
\pgfsetmiterjoin%
\pgfsetlinewidth{0.803000pt}%
\definecolor{currentstroke}{rgb}{0.000000,0.000000,0.000000}%
\pgfsetstrokecolor{currentstroke}%
\pgfsetdash{}{0pt}%
\pgfpathmoveto{\pgfqpoint{0.594914in}{4.243888in}}%
\pgfpathlineto{\pgfqpoint{5.554914in}{4.243888in}}%
\pgfusepath{stroke}%
\end{pgfscope}%
\begin{pgfscope}%
\pgfsetbuttcap%
\pgfsetmiterjoin%
\definecolor{currentfill}{rgb}{1.000000,1.000000,1.000000}%
\pgfsetfillcolor{currentfill}%
\pgfsetfillopacity{0.800000}%
\pgfsetlinewidth{1.003750pt}%
\definecolor{currentstroke}{rgb}{0.800000,0.800000,0.800000}%
\pgfsetstrokecolor{currentstroke}%
\pgfsetstrokeopacity{0.800000}%
\pgfsetdash{}{0pt}%
\pgfpathmoveto{\pgfqpoint{0.750469in}{2.824112in}}%
\pgfpathlineto{\pgfqpoint{3.201803in}{2.824112in}}%
\pgfpathquadraticcurveto{\pgfqpoint{3.246247in}{2.824112in}}{\pgfqpoint{3.246247in}{2.868556in}}%
\pgfpathlineto{\pgfqpoint{3.246247in}{4.088333in}}%
\pgfpathquadraticcurveto{\pgfqpoint{3.246247in}{4.132777in}}{\pgfqpoint{3.201803in}{4.132777in}}%
\pgfpathlineto{\pgfqpoint{0.750469in}{4.132777in}}%
\pgfpathquadraticcurveto{\pgfqpoint{0.706025in}{4.132777in}}{\pgfqpoint{0.706025in}{4.088333in}}%
\pgfpathlineto{\pgfqpoint{0.706025in}{2.868556in}}%
\pgfpathquadraticcurveto{\pgfqpoint{0.706025in}{2.824112in}}{\pgfqpoint{0.750469in}{2.824112in}}%
\pgfpathclose%
\pgfusepath{stroke,fill}%
\end{pgfscope}%
\begin{pgfscope}%
\pgfsetrectcap%
\pgfsetroundjoin%
\pgfsetlinewidth{1.505625pt}%
\definecolor{currentstroke}{rgb}{0.498039,0.498039,0.498039}%
\pgfsetstrokecolor{currentstroke}%
\pgfsetdash{}{0pt}%
\pgfpathmoveto{\pgfqpoint{0.794914in}{3.966111in}}%
\pgfpathlineto{\pgfqpoint{1.239358in}{3.966111in}}%
\pgfusepath{stroke}%
\end{pgfscope}%
\begin{pgfscope}%
\definecolor{textcolor}{rgb}{0.000000,0.000000,0.000000}%
\pgfsetstrokecolor{textcolor}%
\pgfsetfillcolor{textcolor}%
\pgftext[x=1.417136in,y=3.888333in,left,base]{\color{textcolor}\fontsize{16.000000}{19.200000}\selectfont original encoding}%
\end{pgfscope}%
\begin{pgfscope}%
\pgfsetbuttcap%
\pgfsetroundjoin%
\pgfsetlinewidth{1.505625pt}%
\definecolor{currentstroke}{rgb}{1.000000,0.498039,0.054902}%
\pgfsetstrokecolor{currentstroke}%
\pgfsetdash{{5.550000pt}{2.400000pt}}{0.000000pt}%
\pgfpathmoveto{\pgfqpoint{0.794914in}{3.654111in}}%
\pgfpathlineto{\pgfqpoint{1.239358in}{3.654111in}}%
\pgfusepath{stroke}%
\end{pgfscope}%
\begin{pgfscope}%
\definecolor{textcolor}{rgb}{0.000000,0.000000,0.000000}%
\pgfsetstrokecolor{textcolor}%
\pgfsetfillcolor{textcolor}%
\pgftext[x=1.417136in,y=3.576333in,left,base]{\color{textcolor}\fontsize{16.000000}{19.200000}\selectfont learned at first UIP}%
\end{pgfscope}%
\begin{pgfscope}%
\pgfsetbuttcap%
\pgfsetroundjoin%
\pgfsetlinewidth{1.505625pt}%
\definecolor{currentstroke}{rgb}{0.839216,0.152941,0.156863}%
\pgfsetstrokecolor{currentstroke}%
\pgfsetdash{{1.500000pt}{2.475000pt}}{0.000000pt}%
\pgfpathmoveto{\pgfqpoint{0.794914in}{3.344111in}}%
\pgfpathlineto{\pgfqpoint{1.239358in}{3.344111in}}%
\pgfusepath{stroke}%
\end{pgfscope}%
\begin{pgfscope}%
\definecolor{textcolor}{rgb}{0.000000,0.000000,0.000000}%
\pgfsetstrokecolor{textcolor}%
\pgfsetfillcolor{textcolor}%
\pgftext[x=1.417136in,y=3.266333in,left,base]{\color{textcolor}\fontsize{16.000000}{19.200000}\selectfont learned at last UIP}%
\end{pgfscope}%
\begin{pgfscope}%
\pgfsetbuttcap%
\pgfsetroundjoin%
\pgfsetlinewidth{1.505625pt}%
\definecolor{currentstroke}{rgb}{0.172549,0.627451,0.172549}%
\pgfsetstrokecolor{currentstroke}%
\pgfsetdash{{9.600000pt}{2.400000pt}{1.500000pt}{2.400000pt}}{0.000000pt}%
\pgfpathmoveto{\pgfqpoint{0.794914in}{3.034111in}}%
\pgfpathlineto{\pgfqpoint{1.239358in}{3.034111in}}%
\pgfusepath{stroke}%
\end{pgfscope}%
\begin{pgfscope}%
\definecolor{textcolor}{rgb}{0.000000,0.000000,0.000000}%
\pgfsetstrokecolor{textcolor}%
\pgfsetfillcolor{textcolor}%
\pgftext[x=1.417136in,y=2.956333in,left,base]{\color{textcolor}\fontsize{16.000000}{19.200000}\selectfont reduced constraint}%
\end{pgfscope}%
\end{pgfpicture}%
\makeatother%
\endgroup%

%% file: figures/cactus_3col_crafted_clingo.pgf
%% Creator: Matplotlib, PGF backend
%%
%% To include the figure in your LaTeX document, write
%%   \input{<filename>.pgf}
%%
%% Make sure the required packages are loaded in your preamble
%%   \usepackage{pgf}
%%
%% and, on pdftex
%%   \usepackage[utf8]{inputenc}\DeclareUnicodeCharacter{2212}{-}
%%
%% or, on luatex and xetex
%%   \usepackage{unicode-math}
%%
%% Figures using additional raster images can only be included by \input if
%% they are in the same directory as the main LaTeX file. For loading figures
%% from other directories you can use the `import` package
%%   \usepackage{import}
%%
%% and then include the figures with
%%   \import{<path to file>}{<filename>.pgf}
%%
%% Matplotlib used the following preamble
%%   \usepackage{fontspec}
%%
\begingroup%
\makeatletter%
\begin{pgfpicture}%
\pgfpathrectangle{\pgfpointorigin}{\pgfqpoint{5.664982in}{4.243888in}}%
\pgfusepath{use as bounding box, clip}%
\begin{pgfscope}%
\pgfsetbuttcap%
\pgfsetmiterjoin%
\definecolor{currentfill}{rgb}{1.000000,1.000000,1.000000}%
\pgfsetfillcolor{currentfill}%
\pgfsetlinewidth{0.000000pt}%
\definecolor{currentstroke}{rgb}{1.000000,1.000000,1.000000}%
\pgfsetstrokecolor{currentstroke}%
\pgfsetdash{}{0pt}%
\pgfpathmoveto{\pgfqpoint{0.000000in}{0.000000in}}%
\pgfpathlineto{\pgfqpoint{5.664982in}{0.000000in}}%
\pgfpathlineto{\pgfqpoint{5.664982in}{4.243888in}}%
\pgfpathlineto{\pgfqpoint{0.000000in}{4.243888in}}%
\pgfpathclose%
\pgfusepath{fill}%
\end{pgfscope}%
\begin{pgfscope}%
\pgfsetbuttcap%
\pgfsetmiterjoin%
\definecolor{currentfill}{rgb}{1.000000,1.000000,1.000000}%
\pgfsetfillcolor{currentfill}%
\pgfsetlinewidth{0.000000pt}%
\definecolor{currentstroke}{rgb}{0.000000,0.000000,0.000000}%
\pgfsetstrokecolor{currentstroke}%
\pgfsetstrokeopacity{0.000000}%
\pgfsetdash{}{0pt}%
\pgfpathmoveto{\pgfqpoint{0.704982in}{0.547888in}}%
\pgfpathlineto{\pgfqpoint{5.664982in}{0.547888in}}%
\pgfpathlineto{\pgfqpoint{5.664982in}{4.243888in}}%
\pgfpathlineto{\pgfqpoint{0.704982in}{4.243888in}}%
\pgfpathclose%
\pgfusepath{fill}%
\end{pgfscope}%
\begin{pgfscope}%
\pgfsetbuttcap%
\pgfsetroundjoin%
\definecolor{currentfill}{rgb}{0.000000,0.000000,0.000000}%
\pgfsetfillcolor{currentfill}%
\pgfsetlinewidth{0.803000pt}%
\definecolor{currentstroke}{rgb}{0.000000,0.000000,0.000000}%
\pgfsetstrokecolor{currentstroke}%
\pgfsetdash{}{0pt}%
\pgfsys@defobject{currentmarker}{\pgfqpoint{0.000000in}{-0.048611in}}{\pgfqpoint{0.000000in}{0.000000in}}{%
\pgfpathmoveto{\pgfqpoint{0.000000in}{0.000000in}}%
\pgfpathlineto{\pgfqpoint{0.000000in}{-0.048611in}}%
\pgfusepath{stroke,fill}%
}%
\begin{pgfscope}%
\pgfsys@transformshift{0.907778in}{0.547888in}%
\pgfsys@useobject{currentmarker}{}%
\end{pgfscope}%
\end{pgfscope}%
\begin{pgfscope}%
\definecolor{textcolor}{rgb}{0.000000,0.000000,0.000000}%
\pgfsetstrokecolor{textcolor}%
\pgfsetfillcolor{textcolor}%
\pgftext[x=0.907778in,y=0.450666in,,top]{\color{textcolor}\fontsize{16.000000}{19.200000}\selectfont \(\displaystyle 0\)}%
\end{pgfscope}%
\begin{pgfscope}%
\pgfsetbuttcap%
\pgfsetroundjoin%
\definecolor{currentfill}{rgb}{0.000000,0.000000,0.000000}%
\pgfsetfillcolor{currentfill}%
\pgfsetlinewidth{0.803000pt}%
\definecolor{currentstroke}{rgb}{0.000000,0.000000,0.000000}%
\pgfsetstrokecolor{currentstroke}%
\pgfsetdash{}{0pt}%
\pgfsys@defobject{currentmarker}{\pgfqpoint{0.000000in}{-0.048611in}}{\pgfqpoint{0.000000in}{0.000000in}}{%
\pgfpathmoveto{\pgfqpoint{0.000000in}{0.000000in}}%
\pgfpathlineto{\pgfqpoint{0.000000in}{-0.048611in}}%
\pgfusepath{stroke,fill}%
}%
\begin{pgfscope}%
\pgfsys@transformshift{1.474247in}{0.547888in}%
\pgfsys@useobject{currentmarker}{}%
\end{pgfscope}%
\end{pgfscope}%
\begin{pgfscope}%
\definecolor{textcolor}{rgb}{0.000000,0.000000,0.000000}%
\pgfsetstrokecolor{textcolor}%
\pgfsetfillcolor{textcolor}%
\pgftext[x=1.474247in,y=0.450666in,,top]{\color{textcolor}\fontsize{16.000000}{19.200000}\selectfont \(\displaystyle 25\)}%
\end{pgfscope}%
\begin{pgfscope}%
\pgfsetbuttcap%
\pgfsetroundjoin%
\definecolor{currentfill}{rgb}{0.000000,0.000000,0.000000}%
\pgfsetfillcolor{currentfill}%
\pgfsetlinewidth{0.803000pt}%
\definecolor{currentstroke}{rgb}{0.000000,0.000000,0.000000}%
\pgfsetstrokecolor{currentstroke}%
\pgfsetdash{}{0pt}%
\pgfsys@defobject{currentmarker}{\pgfqpoint{0.000000in}{-0.048611in}}{\pgfqpoint{0.000000in}{0.000000in}}{%
\pgfpathmoveto{\pgfqpoint{0.000000in}{0.000000in}}%
\pgfpathlineto{\pgfqpoint{0.000000in}{-0.048611in}}%
\pgfusepath{stroke,fill}%
}%
\begin{pgfscope}%
\pgfsys@transformshift{2.040715in}{0.547888in}%
\pgfsys@useobject{currentmarker}{}%
\end{pgfscope}%
\end{pgfscope}%
\begin{pgfscope}%
\definecolor{textcolor}{rgb}{0.000000,0.000000,0.000000}%
\pgfsetstrokecolor{textcolor}%
\pgfsetfillcolor{textcolor}%
\pgftext[x=2.040715in,y=0.450666in,,top]{\color{textcolor}\fontsize{16.000000}{19.200000}\selectfont \(\displaystyle 50\)}%
\end{pgfscope}%
\begin{pgfscope}%
\pgfsetbuttcap%
\pgfsetroundjoin%
\definecolor{currentfill}{rgb}{0.000000,0.000000,0.000000}%
\pgfsetfillcolor{currentfill}%
\pgfsetlinewidth{0.803000pt}%
\definecolor{currentstroke}{rgb}{0.000000,0.000000,0.000000}%
\pgfsetstrokecolor{currentstroke}%
\pgfsetdash{}{0pt}%
\pgfsys@defobject{currentmarker}{\pgfqpoint{0.000000in}{-0.048611in}}{\pgfqpoint{0.000000in}{0.000000in}}{%
\pgfpathmoveto{\pgfqpoint{0.000000in}{0.000000in}}%
\pgfpathlineto{\pgfqpoint{0.000000in}{-0.048611in}}%
\pgfusepath{stroke,fill}%
}%
\begin{pgfscope}%
\pgfsys@transformshift{2.607184in}{0.547888in}%
\pgfsys@useobject{currentmarker}{}%
\end{pgfscope}%
\end{pgfscope}%
\begin{pgfscope}%
\definecolor{textcolor}{rgb}{0.000000,0.000000,0.000000}%
\pgfsetstrokecolor{textcolor}%
\pgfsetfillcolor{textcolor}%
\pgftext[x=2.607184in,y=0.450666in,,top]{\color{textcolor}\fontsize{16.000000}{19.200000}\selectfont \(\displaystyle 75\)}%
\end{pgfscope}%
\begin{pgfscope}%
\pgfsetbuttcap%
\pgfsetroundjoin%
\definecolor{currentfill}{rgb}{0.000000,0.000000,0.000000}%
\pgfsetfillcolor{currentfill}%
\pgfsetlinewidth{0.803000pt}%
\definecolor{currentstroke}{rgb}{0.000000,0.000000,0.000000}%
\pgfsetstrokecolor{currentstroke}%
\pgfsetdash{}{0pt}%
\pgfsys@defobject{currentmarker}{\pgfqpoint{0.000000in}{-0.048611in}}{\pgfqpoint{0.000000in}{0.000000in}}{%
\pgfpathmoveto{\pgfqpoint{0.000000in}{0.000000in}}%
\pgfpathlineto{\pgfqpoint{0.000000in}{-0.048611in}}%
\pgfusepath{stroke,fill}%
}%
\begin{pgfscope}%
\pgfsys@transformshift{3.173653in}{0.547888in}%
\pgfsys@useobject{currentmarker}{}%
\end{pgfscope}%
\end{pgfscope}%
\begin{pgfscope}%
\definecolor{textcolor}{rgb}{0.000000,0.000000,0.000000}%
\pgfsetstrokecolor{textcolor}%
\pgfsetfillcolor{textcolor}%
\pgftext[x=3.173653in,y=0.450666in,,top]{\color{textcolor}\fontsize{16.000000}{19.200000}\selectfont \(\displaystyle 100\)}%
\end{pgfscope}%
\begin{pgfscope}%
\pgfsetbuttcap%
\pgfsetroundjoin%
\definecolor{currentfill}{rgb}{0.000000,0.000000,0.000000}%
\pgfsetfillcolor{currentfill}%
\pgfsetlinewidth{0.803000pt}%
\definecolor{currentstroke}{rgb}{0.000000,0.000000,0.000000}%
\pgfsetstrokecolor{currentstroke}%
\pgfsetdash{}{0pt}%
\pgfsys@defobject{currentmarker}{\pgfqpoint{0.000000in}{-0.048611in}}{\pgfqpoint{0.000000in}{0.000000in}}{%
\pgfpathmoveto{\pgfqpoint{0.000000in}{0.000000in}}%
\pgfpathlineto{\pgfqpoint{0.000000in}{-0.048611in}}%
\pgfusepath{stroke,fill}%
}%
\begin{pgfscope}%
\pgfsys@transformshift{3.740121in}{0.547888in}%
\pgfsys@useobject{currentmarker}{}%
\end{pgfscope}%
\end{pgfscope}%
\begin{pgfscope}%
\definecolor{textcolor}{rgb}{0.000000,0.000000,0.000000}%
\pgfsetstrokecolor{textcolor}%
\pgfsetfillcolor{textcolor}%
\pgftext[x=3.740121in,y=0.450666in,,top]{\color{textcolor}\fontsize{16.000000}{19.200000}\selectfont \(\displaystyle 125\)}%
\end{pgfscope}%
\begin{pgfscope}%
\pgfsetbuttcap%
\pgfsetroundjoin%
\definecolor{currentfill}{rgb}{0.000000,0.000000,0.000000}%
\pgfsetfillcolor{currentfill}%
\pgfsetlinewidth{0.803000pt}%
\definecolor{currentstroke}{rgb}{0.000000,0.000000,0.000000}%
\pgfsetstrokecolor{currentstroke}%
\pgfsetdash{}{0pt}%
\pgfsys@defobject{currentmarker}{\pgfqpoint{0.000000in}{-0.048611in}}{\pgfqpoint{0.000000in}{0.000000in}}{%
\pgfpathmoveto{\pgfqpoint{0.000000in}{0.000000in}}%
\pgfpathlineto{\pgfqpoint{0.000000in}{-0.048611in}}%
\pgfusepath{stroke,fill}%
}%
\begin{pgfscope}%
\pgfsys@transformshift{4.306590in}{0.547888in}%
\pgfsys@useobject{currentmarker}{}%
\end{pgfscope}%
\end{pgfscope}%
\begin{pgfscope}%
\definecolor{textcolor}{rgb}{0.000000,0.000000,0.000000}%
\pgfsetstrokecolor{textcolor}%
\pgfsetfillcolor{textcolor}%
\pgftext[x=4.306590in,y=0.450666in,,top]{\color{textcolor}\fontsize{16.000000}{19.200000}\selectfont \(\displaystyle 150\)}%
\end{pgfscope}%
\begin{pgfscope}%
\pgfsetbuttcap%
\pgfsetroundjoin%
\definecolor{currentfill}{rgb}{0.000000,0.000000,0.000000}%
\pgfsetfillcolor{currentfill}%
\pgfsetlinewidth{0.803000pt}%
\definecolor{currentstroke}{rgb}{0.000000,0.000000,0.000000}%
\pgfsetstrokecolor{currentstroke}%
\pgfsetdash{}{0pt}%
\pgfsys@defobject{currentmarker}{\pgfqpoint{0.000000in}{-0.048611in}}{\pgfqpoint{0.000000in}{0.000000in}}{%
\pgfpathmoveto{\pgfqpoint{0.000000in}{0.000000in}}%
\pgfpathlineto{\pgfqpoint{0.000000in}{-0.048611in}}%
\pgfusepath{stroke,fill}%
}%
\begin{pgfscope}%
\pgfsys@transformshift{4.873059in}{0.547888in}%
\pgfsys@useobject{currentmarker}{}%
\end{pgfscope}%
\end{pgfscope}%
\begin{pgfscope}%
\definecolor{textcolor}{rgb}{0.000000,0.000000,0.000000}%
\pgfsetstrokecolor{textcolor}%
\pgfsetfillcolor{textcolor}%
\pgftext[x=4.873059in,y=0.450666in,,top]{\color{textcolor}\fontsize{16.000000}{19.200000}\selectfont \(\displaystyle 175\)}%
\end{pgfscope}%
\begin{pgfscope}%
\pgfsetbuttcap%
\pgfsetroundjoin%
\definecolor{currentfill}{rgb}{0.000000,0.000000,0.000000}%
\pgfsetfillcolor{currentfill}%
\pgfsetlinewidth{0.803000pt}%
\definecolor{currentstroke}{rgb}{0.000000,0.000000,0.000000}%
\pgfsetstrokecolor{currentstroke}%
\pgfsetdash{}{0pt}%
\pgfsys@defobject{currentmarker}{\pgfqpoint{0.000000in}{-0.048611in}}{\pgfqpoint{0.000000in}{0.000000in}}{%
\pgfpathmoveto{\pgfqpoint{0.000000in}{0.000000in}}%
\pgfpathlineto{\pgfqpoint{0.000000in}{-0.048611in}}%
\pgfusepath{stroke,fill}%
}%
\begin{pgfscope}%
\pgfsys@transformshift{5.439528in}{0.547888in}%
\pgfsys@useobject{currentmarker}{}%
\end{pgfscope}%
\end{pgfscope}%
\begin{pgfscope}%
\definecolor{textcolor}{rgb}{0.000000,0.000000,0.000000}%
\pgfsetstrokecolor{textcolor}%
\pgfsetfillcolor{textcolor}%
\pgftext[x=5.439528in,y=0.450666in,,top]{\color{textcolor}\fontsize{16.000000}{19.200000}\selectfont \(\displaystyle 200\)}%
\end{pgfscope}%
\begin{pgfscope}%
\definecolor{textcolor}{rgb}{0.000000,0.000000,0.000000}%
\pgfsetstrokecolor{textcolor}%
\pgfsetfillcolor{textcolor}%
\pgftext[x=3.184982in,y=0.197555in,,top]{\color{textcolor}\fontsize{16.000000}{19.200000}\selectfont Number of instances}%
\end{pgfscope}%
\begin{pgfscope}%
\pgfsetbuttcap%
\pgfsetroundjoin%
\definecolor{currentfill}{rgb}{0.000000,0.000000,0.000000}%
\pgfsetfillcolor{currentfill}%
\pgfsetlinewidth{0.803000pt}%
\definecolor{currentstroke}{rgb}{0.000000,0.000000,0.000000}%
\pgfsetstrokecolor{currentstroke}%
\pgfsetdash{}{0pt}%
\pgfsys@defobject{currentmarker}{\pgfqpoint{-0.048611in}{0.000000in}}{\pgfqpoint{0.000000in}{0.000000in}}{%
\pgfpathmoveto{\pgfqpoint{0.000000in}{0.000000in}}%
\pgfpathlineto{\pgfqpoint{-0.048611in}{0.000000in}}%
\pgfusepath{stroke,fill}%
}%
\begin{pgfscope}%
\pgfsys@transformshift{0.704982in}{0.715866in}%
\pgfsys@useobject{currentmarker}{}%
\end{pgfscope}%
\end{pgfscope}%
\begin{pgfscope}%
\definecolor{textcolor}{rgb}{0.000000,0.000000,0.000000}%
\pgfsetstrokecolor{textcolor}%
\pgfsetfillcolor{textcolor}%
\pgftext[x=0.497692in, y=0.638755in, left, base]{\color{textcolor}\fontsize{16.000000}{19.200000}\selectfont \(\displaystyle 0\)}%
\end{pgfscope}%
\begin{pgfscope}%
\pgfsetbuttcap%
\pgfsetroundjoin%
\definecolor{currentfill}{rgb}{0.000000,0.000000,0.000000}%
\pgfsetfillcolor{currentfill}%
\pgfsetlinewidth{0.803000pt}%
\definecolor{currentstroke}{rgb}{0.000000,0.000000,0.000000}%
\pgfsetstrokecolor{currentstroke}%
\pgfsetdash{}{0pt}%
\pgfsys@defobject{currentmarker}{\pgfqpoint{-0.048611in}{0.000000in}}{\pgfqpoint{0.000000in}{0.000000in}}{%
\pgfpathmoveto{\pgfqpoint{0.000000in}{0.000000in}}%
\pgfpathlineto{\pgfqpoint{-0.048611in}{0.000000in}}%
\pgfusepath{stroke,fill}%
}%
\begin{pgfscope}%
\pgfsys@transformshift{0.704982in}{1.110212in}%
\pgfsys@useobject{currentmarker}{}%
\end{pgfscope}%
\end{pgfscope}%
\begin{pgfscope}%
\definecolor{textcolor}{rgb}{0.000000,0.000000,0.000000}%
\pgfsetstrokecolor{textcolor}%
\pgfsetfillcolor{textcolor}%
\pgftext[x=0.387623in, y=1.033101in, left, base]{\color{textcolor}\fontsize{16.000000}{19.200000}\selectfont \(\displaystyle 30\)}%
\end{pgfscope}%
\begin{pgfscope}%
\pgfsetbuttcap%
\pgfsetroundjoin%
\definecolor{currentfill}{rgb}{0.000000,0.000000,0.000000}%
\pgfsetfillcolor{currentfill}%
\pgfsetlinewidth{0.803000pt}%
\definecolor{currentstroke}{rgb}{0.000000,0.000000,0.000000}%
\pgfsetstrokecolor{currentstroke}%
\pgfsetdash{}{0pt}%
\pgfsys@defobject{currentmarker}{\pgfqpoint{-0.048611in}{0.000000in}}{\pgfqpoint{0.000000in}{0.000000in}}{%
\pgfpathmoveto{\pgfqpoint{0.000000in}{0.000000in}}%
\pgfpathlineto{\pgfqpoint{-0.048611in}{0.000000in}}%
\pgfusepath{stroke,fill}%
}%
\begin{pgfscope}%
\pgfsys@transformshift{0.704982in}{1.504558in}%
\pgfsys@useobject{currentmarker}{}%
\end{pgfscope}%
\end{pgfscope}%
\begin{pgfscope}%
\definecolor{textcolor}{rgb}{0.000000,0.000000,0.000000}%
\pgfsetstrokecolor{textcolor}%
\pgfsetfillcolor{textcolor}%
\pgftext[x=0.387623in, y=1.427447in, left, base]{\color{textcolor}\fontsize{16.000000}{19.200000}\selectfont \(\displaystyle 60\)}%
\end{pgfscope}%
\begin{pgfscope}%
\pgfsetbuttcap%
\pgfsetroundjoin%
\definecolor{currentfill}{rgb}{0.000000,0.000000,0.000000}%
\pgfsetfillcolor{currentfill}%
\pgfsetlinewidth{0.803000pt}%
\definecolor{currentstroke}{rgb}{0.000000,0.000000,0.000000}%
\pgfsetstrokecolor{currentstroke}%
\pgfsetdash{}{0pt}%
\pgfsys@defobject{currentmarker}{\pgfqpoint{-0.048611in}{0.000000in}}{\pgfqpoint{0.000000in}{0.000000in}}{%
\pgfpathmoveto{\pgfqpoint{0.000000in}{0.000000in}}%
\pgfpathlineto{\pgfqpoint{-0.048611in}{0.000000in}}%
\pgfusepath{stroke,fill}%
}%
\begin{pgfscope}%
\pgfsys@transformshift{0.704982in}{1.898904in}%
\pgfsys@useobject{currentmarker}{}%
\end{pgfscope}%
\end{pgfscope}%
\begin{pgfscope}%
\definecolor{textcolor}{rgb}{0.000000,0.000000,0.000000}%
\pgfsetstrokecolor{textcolor}%
\pgfsetfillcolor{textcolor}%
\pgftext[x=0.387623in, y=1.821793in, left, base]{\color{textcolor}\fontsize{16.000000}{19.200000}\selectfont \(\displaystyle 90\)}%
\end{pgfscope}%
\begin{pgfscope}%
\pgfsetbuttcap%
\pgfsetroundjoin%
\definecolor{currentfill}{rgb}{0.000000,0.000000,0.000000}%
\pgfsetfillcolor{currentfill}%
\pgfsetlinewidth{0.803000pt}%
\definecolor{currentstroke}{rgb}{0.000000,0.000000,0.000000}%
\pgfsetstrokecolor{currentstroke}%
\pgfsetdash{}{0pt}%
\pgfsys@defobject{currentmarker}{\pgfqpoint{-0.048611in}{0.000000in}}{\pgfqpoint{0.000000in}{0.000000in}}{%
\pgfpathmoveto{\pgfqpoint{0.000000in}{0.000000in}}%
\pgfpathlineto{\pgfqpoint{-0.048611in}{0.000000in}}%
\pgfusepath{stroke,fill}%
}%
\begin{pgfscope}%
\pgfsys@transformshift{0.704982in}{2.293250in}%
\pgfsys@useobject{currentmarker}{}%
\end{pgfscope}%
\end{pgfscope}%
\begin{pgfscope}%
\definecolor{textcolor}{rgb}{0.000000,0.000000,0.000000}%
\pgfsetstrokecolor{textcolor}%
\pgfsetfillcolor{textcolor}%
\pgftext[x=0.277555in, y=2.216139in, left, base]{\color{textcolor}\fontsize{16.000000}{19.200000}\selectfont \(\displaystyle 120\)}%
\end{pgfscope}%
\begin{pgfscope}%
\pgfsetbuttcap%
\pgfsetroundjoin%
\definecolor{currentfill}{rgb}{0.000000,0.000000,0.000000}%
\pgfsetfillcolor{currentfill}%
\pgfsetlinewidth{0.803000pt}%
\definecolor{currentstroke}{rgb}{0.000000,0.000000,0.000000}%
\pgfsetstrokecolor{currentstroke}%
\pgfsetdash{}{0pt}%
\pgfsys@defobject{currentmarker}{\pgfqpoint{-0.048611in}{0.000000in}}{\pgfqpoint{0.000000in}{0.000000in}}{%
\pgfpathmoveto{\pgfqpoint{0.000000in}{0.000000in}}%
\pgfpathlineto{\pgfqpoint{-0.048611in}{0.000000in}}%
\pgfusepath{stroke,fill}%
}%
\begin{pgfscope}%
\pgfsys@transformshift{0.704982in}{2.687596in}%
\pgfsys@useobject{currentmarker}{}%
\end{pgfscope}%
\end{pgfscope}%
\begin{pgfscope}%
\definecolor{textcolor}{rgb}{0.000000,0.000000,0.000000}%
\pgfsetstrokecolor{textcolor}%
\pgfsetfillcolor{textcolor}%
\pgftext[x=0.277555in, y=2.610485in, left, base]{\color{textcolor}\fontsize{16.000000}{19.200000}\selectfont \(\displaystyle 150\)}%
\end{pgfscope}%
\begin{pgfscope}%
\pgfsetbuttcap%
\pgfsetroundjoin%
\definecolor{currentfill}{rgb}{0.000000,0.000000,0.000000}%
\pgfsetfillcolor{currentfill}%
\pgfsetlinewidth{0.803000pt}%
\definecolor{currentstroke}{rgb}{0.000000,0.000000,0.000000}%
\pgfsetstrokecolor{currentstroke}%
\pgfsetdash{}{0pt}%
\pgfsys@defobject{currentmarker}{\pgfqpoint{-0.048611in}{0.000000in}}{\pgfqpoint{0.000000in}{0.000000in}}{%
\pgfpathmoveto{\pgfqpoint{0.000000in}{0.000000in}}%
\pgfpathlineto{\pgfqpoint{-0.048611in}{0.000000in}}%
\pgfusepath{stroke,fill}%
}%
\begin{pgfscope}%
\pgfsys@transformshift{0.704982in}{3.081942in}%
\pgfsys@useobject{currentmarker}{}%
\end{pgfscope}%
\end{pgfscope}%
\begin{pgfscope}%
\definecolor{textcolor}{rgb}{0.000000,0.000000,0.000000}%
\pgfsetstrokecolor{textcolor}%
\pgfsetfillcolor{textcolor}%
\pgftext[x=0.277555in, y=3.004831in, left, base]{\color{textcolor}\fontsize{16.000000}{19.200000}\selectfont \(\displaystyle 180\)}%
\end{pgfscope}%
\begin{pgfscope}%
\pgfsetbuttcap%
\pgfsetroundjoin%
\definecolor{currentfill}{rgb}{0.000000,0.000000,0.000000}%
\pgfsetfillcolor{currentfill}%
\pgfsetlinewidth{0.803000pt}%
\definecolor{currentstroke}{rgb}{0.000000,0.000000,0.000000}%
\pgfsetstrokecolor{currentstroke}%
\pgfsetdash{}{0pt}%
\pgfsys@defobject{currentmarker}{\pgfqpoint{-0.048611in}{0.000000in}}{\pgfqpoint{0.000000in}{0.000000in}}{%
\pgfpathmoveto{\pgfqpoint{0.000000in}{0.000000in}}%
\pgfpathlineto{\pgfqpoint{-0.048611in}{0.000000in}}%
\pgfusepath{stroke,fill}%
}%
\begin{pgfscope}%
\pgfsys@transformshift{0.704982in}{3.476289in}%
\pgfsys@useobject{currentmarker}{}%
\end{pgfscope}%
\end{pgfscope}%
\begin{pgfscope}%
\definecolor{textcolor}{rgb}{0.000000,0.000000,0.000000}%
\pgfsetstrokecolor{textcolor}%
\pgfsetfillcolor{textcolor}%
\pgftext[x=0.277555in, y=3.399177in, left, base]{\color{textcolor}\fontsize{16.000000}{19.200000}\selectfont \(\displaystyle 210\)}%
\end{pgfscope}%
\begin{pgfscope}%
\pgfsetbuttcap%
\pgfsetroundjoin%
\definecolor{currentfill}{rgb}{0.000000,0.000000,0.000000}%
\pgfsetfillcolor{currentfill}%
\pgfsetlinewidth{0.803000pt}%
\definecolor{currentstroke}{rgb}{0.000000,0.000000,0.000000}%
\pgfsetstrokecolor{currentstroke}%
\pgfsetdash{}{0pt}%
\pgfsys@defobject{currentmarker}{\pgfqpoint{-0.048611in}{0.000000in}}{\pgfqpoint{0.000000in}{0.000000in}}{%
\pgfpathmoveto{\pgfqpoint{0.000000in}{0.000000in}}%
\pgfpathlineto{\pgfqpoint{-0.048611in}{0.000000in}}%
\pgfusepath{stroke,fill}%
}%
\begin{pgfscope}%
\pgfsys@transformshift{0.704982in}{3.870635in}%
\pgfsys@useobject{currentmarker}{}%
\end{pgfscope}%
\end{pgfscope}%
\begin{pgfscope}%
\definecolor{textcolor}{rgb}{0.000000,0.000000,0.000000}%
\pgfsetstrokecolor{textcolor}%
\pgfsetfillcolor{textcolor}%
\pgftext[x=0.277555in, y=3.793524in, left, base]{\color{textcolor}\fontsize{16.000000}{19.200000}\selectfont \(\displaystyle 240\)}%
\end{pgfscope}%
\begin{pgfscope}%
\definecolor{textcolor}{rgb}{0.000000,0.000000,0.000000}%
\pgfsetstrokecolor{textcolor}%
\pgfsetfillcolor{textcolor}%
\pgftext[x=0.222000in,y=2.395888in,,bottom,rotate=90.000000]{\color{textcolor}\fontsize{16.000000}{19.200000}\selectfont Real time consumption (minutes)}%
\end{pgfscope}%
\begin{pgfscope}%
\pgfpathrectangle{\pgfqpoint{0.704982in}{0.547888in}}{\pgfqpoint{4.960000in}{3.696000in}}%
\pgfusepath{clip}%
\pgfsetrectcap%
\pgfsetroundjoin%
\pgfsetlinewidth{1.505625pt}%
\definecolor{currentstroke}{rgb}{0.498039,0.498039,0.498039}%
\pgfsetstrokecolor{currentstroke}%
\pgfsetdash{}{0pt}%
\pgfpathmoveto{\pgfqpoint{0.930437in}{0.715889in}}%
\pgfpathlineto{\pgfqpoint{1.179683in}{0.716767in}}%
\pgfpathlineto{\pgfqpoint{1.315635in}{0.718211in}}%
\pgfpathlineto{\pgfqpoint{1.451588in}{0.720790in}}%
\pgfpathlineto{\pgfqpoint{1.542223in}{0.723590in}}%
\pgfpathlineto{\pgfqpoint{1.632858in}{0.727977in}}%
\pgfpathlineto{\pgfqpoint{1.768810in}{0.736229in}}%
\pgfpathlineto{\pgfqpoint{1.859445in}{0.742783in}}%
\pgfpathlineto{\pgfqpoint{1.950080in}{0.750477in}}%
\pgfpathlineto{\pgfqpoint{2.063374in}{0.762077in}}%
\pgfpathlineto{\pgfqpoint{2.131350in}{0.770190in}}%
\pgfpathlineto{\pgfqpoint{2.244644in}{0.786420in}}%
\pgfpathlineto{\pgfqpoint{2.312620in}{0.797153in}}%
\pgfpathlineto{\pgfqpoint{2.403255in}{0.812692in}}%
\pgfpathlineto{\pgfqpoint{2.493890in}{0.831046in}}%
\pgfpathlineto{\pgfqpoint{2.584525in}{0.851769in}}%
\pgfpathlineto{\pgfqpoint{2.652501in}{0.868554in}}%
\pgfpathlineto{\pgfqpoint{2.743136in}{0.892824in}}%
\pgfpathlineto{\pgfqpoint{2.833771in}{0.919874in}}%
\pgfpathlineto{\pgfqpoint{2.924406in}{0.947834in}}%
\pgfpathlineto{\pgfqpoint{2.992383in}{0.969865in}}%
\pgfpathlineto{\pgfqpoint{3.060359in}{0.994109in}}%
\pgfpathlineto{\pgfqpoint{3.173653in}{1.036007in}}%
\pgfpathlineto{\pgfqpoint{3.218970in}{1.053627in}}%
\pgfpathlineto{\pgfqpoint{3.309605in}{1.090471in}}%
\pgfpathlineto{\pgfqpoint{3.400240in}{1.128922in}}%
\pgfpathlineto{\pgfqpoint{3.468216in}{1.159741in}}%
\pgfpathlineto{\pgfqpoint{3.558851in}{1.202961in}}%
\pgfpathlineto{\pgfqpoint{3.581510in}{1.214236in}}%
\pgfpathlineto{\pgfqpoint{3.649486in}{1.252093in}}%
\pgfpathlineto{\pgfqpoint{3.740121in}{1.303956in}}%
\pgfpathlineto{\pgfqpoint{3.808098in}{1.345904in}}%
\pgfpathlineto{\pgfqpoint{3.830756in}{1.360529in}}%
\pgfpathlineto{\pgfqpoint{3.898733in}{1.407717in}}%
\pgfpathlineto{\pgfqpoint{3.966709in}{1.457649in}}%
\pgfpathlineto{\pgfqpoint{4.012026in}{1.492253in}}%
\pgfpathlineto{\pgfqpoint{4.034685in}{1.510014in}}%
\pgfpathlineto{\pgfqpoint{4.057344in}{1.529333in}}%
\pgfpathlineto{\pgfqpoint{4.125320in}{1.593276in}}%
\pgfpathlineto{\pgfqpoint{4.193296in}{1.660118in}}%
\pgfpathlineto{\pgfqpoint{4.261273in}{1.730981in}}%
\pgfpathlineto{\pgfqpoint{4.306590in}{1.782339in}}%
\pgfpathlineto{\pgfqpoint{4.374566in}{1.861512in}}%
\pgfpathlineto{\pgfqpoint{4.419884in}{1.915999in}}%
\pgfpathlineto{\pgfqpoint{4.465201in}{1.972525in}}%
\pgfpathlineto{\pgfqpoint{4.533178in}{2.061100in}}%
\pgfpathlineto{\pgfqpoint{4.646471in}{2.214592in}}%
\pgfpathlineto{\pgfqpoint{4.669130in}{2.245960in}}%
\pgfpathlineto{\pgfqpoint{4.714448in}{2.314420in}}%
\pgfpathlineto{\pgfqpoint{4.759765in}{2.384271in}}%
\pgfpathlineto{\pgfqpoint{4.827741in}{2.490851in}}%
\pgfpathlineto{\pgfqpoint{4.873059in}{2.566111in}}%
\pgfpathlineto{\pgfqpoint{4.918376in}{2.645966in}}%
\pgfpathlineto{\pgfqpoint{4.963694in}{2.728870in}}%
\pgfpathlineto{\pgfqpoint{5.009011in}{2.824751in}}%
\pgfpathlineto{\pgfqpoint{5.031670in}{2.874294in}}%
\pgfpathlineto{\pgfqpoint{5.076988in}{2.977985in}}%
\pgfpathlineto{\pgfqpoint{5.122305in}{3.087797in}}%
\pgfpathlineto{\pgfqpoint{5.167623in}{3.202382in}}%
\pgfpathlineto{\pgfqpoint{5.212940in}{3.326938in}}%
\pgfpathlineto{\pgfqpoint{5.258258in}{3.458212in}}%
\pgfpathlineto{\pgfqpoint{5.303575in}{3.595639in}}%
\pgfpathlineto{\pgfqpoint{5.326234in}{3.667659in}}%
\pgfpathlineto{\pgfqpoint{5.371551in}{3.819759in}}%
\pgfpathlineto{\pgfqpoint{5.394210in}{3.896674in}}%
\pgfpathlineto{\pgfqpoint{5.416869in}{3.979207in}}%
\pgfpathlineto{\pgfqpoint{5.439528in}{4.075888in}}%
\pgfpathlineto{\pgfqpoint{5.439528in}{4.075888in}}%
\pgfusepath{stroke}%
\end{pgfscope}%
\begin{pgfscope}%
\pgfpathrectangle{\pgfqpoint{0.704982in}{0.547888in}}{\pgfqpoint{4.960000in}{3.696000in}}%
\pgfusepath{clip}%
\pgfsetbuttcap%
\pgfsetroundjoin%
\pgfsetlinewidth{1.505625pt}%
\definecolor{currentstroke}{rgb}{1.000000,0.498039,0.054902}%
\pgfsetstrokecolor{currentstroke}%
\pgfsetdash{{5.550000pt}{2.400000pt}}{0.000000pt}%
\pgfpathmoveto{\pgfqpoint{0.930437in}{0.715888in}}%
\pgfpathlineto{\pgfqpoint{1.428929in}{0.717045in}}%
\pgfpathlineto{\pgfqpoint{1.904763in}{0.719347in}}%
\pgfpathlineto{\pgfqpoint{2.335279in}{0.722442in}}%
\pgfpathlineto{\pgfqpoint{2.743136in}{0.726551in}}%
\pgfpathlineto{\pgfqpoint{3.241629in}{0.733298in}}%
\pgfpathlineto{\pgfqpoint{3.626828in}{0.739618in}}%
\pgfpathlineto{\pgfqpoint{4.102661in}{0.748720in}}%
\pgfpathlineto{\pgfqpoint{4.442543in}{0.756407in}}%
\pgfpathlineto{\pgfqpoint{4.759765in}{0.764937in}}%
\pgfpathlineto{\pgfqpoint{5.031670in}{0.774051in}}%
\pgfpathlineto{\pgfqpoint{5.326234in}{0.785343in}}%
\pgfpathlineto{\pgfqpoint{5.439528in}{0.790126in}}%
\pgfpathlineto{\pgfqpoint{5.439528in}{0.790126in}}%
\pgfusepath{stroke}%
\end{pgfscope}%
\begin{pgfscope}%
\pgfpathrectangle{\pgfqpoint{0.704982in}{0.547888in}}{\pgfqpoint{4.960000in}{3.696000in}}%
\pgfusepath{clip}%
\pgfsetbuttcap%
\pgfsetroundjoin%
\pgfsetlinewidth{1.505625pt}%
\definecolor{currentstroke}{rgb}{0.839216,0.152941,0.156863}%
\pgfsetstrokecolor{currentstroke}%
\pgfsetdash{{1.500000pt}{2.475000pt}}{0.000000pt}%
\pgfpathmoveto{\pgfqpoint{0.930437in}{0.715889in}}%
\pgfpathlineto{\pgfqpoint{1.202342in}{0.716809in}}%
\pgfpathlineto{\pgfqpoint{1.383612in}{0.718576in}}%
\pgfpathlineto{\pgfqpoint{1.564882in}{0.721797in}}%
\pgfpathlineto{\pgfqpoint{1.700834in}{0.725848in}}%
\pgfpathlineto{\pgfqpoint{1.836787in}{0.731345in}}%
\pgfpathlineto{\pgfqpoint{1.950080in}{0.737050in}}%
\pgfpathlineto{\pgfqpoint{2.040715in}{0.742820in}}%
\pgfpathlineto{\pgfqpoint{2.131350in}{0.750071in}}%
\pgfpathlineto{\pgfqpoint{2.267303in}{0.763199in}}%
\pgfpathlineto{\pgfqpoint{2.403255in}{0.777671in}}%
\pgfpathlineto{\pgfqpoint{2.471232in}{0.785918in}}%
\pgfpathlineto{\pgfqpoint{2.561867in}{0.798947in}}%
\pgfpathlineto{\pgfqpoint{2.629843in}{0.810088in}}%
\pgfpathlineto{\pgfqpoint{2.697819in}{0.822469in}}%
\pgfpathlineto{\pgfqpoint{2.811113in}{0.844387in}}%
\pgfpathlineto{\pgfqpoint{2.901748in}{0.862867in}}%
\pgfpathlineto{\pgfqpoint{2.992383in}{0.882592in}}%
\pgfpathlineto{\pgfqpoint{3.083018in}{0.904055in}}%
\pgfpathlineto{\pgfqpoint{3.218970in}{0.938857in}}%
\pgfpathlineto{\pgfqpoint{3.309605in}{0.963257in}}%
\pgfpathlineto{\pgfqpoint{3.400240in}{0.989210in}}%
\pgfpathlineto{\pgfqpoint{3.513534in}{1.024013in}}%
\pgfpathlineto{\pgfqpoint{3.604169in}{1.053201in}}%
\pgfpathlineto{\pgfqpoint{3.762780in}{1.107926in}}%
\pgfpathlineto{\pgfqpoint{3.853415in}{1.141116in}}%
\pgfpathlineto{\pgfqpoint{3.876074in}{1.149516in}}%
\pgfpathlineto{\pgfqpoint{3.898733in}{1.158792in}}%
\pgfpathlineto{\pgfqpoint{3.966709in}{1.189542in}}%
\pgfpathlineto{\pgfqpoint{4.057344in}{1.234325in}}%
\pgfpathlineto{\pgfqpoint{4.147979in}{1.281047in}}%
\pgfpathlineto{\pgfqpoint{4.193296in}{1.305407in}}%
\pgfpathlineto{\pgfqpoint{4.215955in}{1.318995in}}%
\pgfpathlineto{\pgfqpoint{4.261273in}{1.348333in}}%
\pgfpathlineto{\pgfqpoint{4.329249in}{1.396962in}}%
\pgfpathlineto{\pgfqpoint{4.374566in}{1.432310in}}%
\pgfpathlineto{\pgfqpoint{4.419884in}{1.469227in}}%
\pgfpathlineto{\pgfqpoint{4.465201in}{1.508800in}}%
\pgfpathlineto{\pgfqpoint{4.533178in}{1.570115in}}%
\pgfpathlineto{\pgfqpoint{4.623813in}{1.655295in}}%
\pgfpathlineto{\pgfqpoint{4.646471in}{1.677120in}}%
\pgfpathlineto{\pgfqpoint{4.691789in}{1.728854in}}%
\pgfpathlineto{\pgfqpoint{4.759765in}{1.814509in}}%
\pgfpathlineto{\pgfqpoint{4.805083in}{1.873273in}}%
\pgfpathlineto{\pgfqpoint{4.827741in}{1.903606in}}%
\pgfpathlineto{\pgfqpoint{4.873059in}{1.966740in}}%
\pgfpathlineto{\pgfqpoint{4.941035in}{2.072029in}}%
\pgfpathlineto{\pgfqpoint{4.963694in}{2.109447in}}%
\pgfpathlineto{\pgfqpoint{5.009011in}{2.188292in}}%
\pgfpathlineto{\pgfqpoint{5.054329in}{2.272025in}}%
\pgfpathlineto{\pgfqpoint{5.099646in}{2.358935in}}%
\pgfpathlineto{\pgfqpoint{5.167623in}{2.492149in}}%
\pgfpathlineto{\pgfqpoint{5.212940in}{2.585527in}}%
\pgfpathlineto{\pgfqpoint{5.280916in}{2.732250in}}%
\pgfpathlineto{\pgfqpoint{5.326234in}{2.836355in}}%
\pgfpathlineto{\pgfqpoint{5.371551in}{2.944369in}}%
\pgfpathlineto{\pgfqpoint{5.394210in}{3.001236in}}%
\pgfpathlineto{\pgfqpoint{5.416869in}{3.063126in}}%
\pgfpathlineto{\pgfqpoint{5.439528in}{3.127393in}}%
\pgfpathlineto{\pgfqpoint{5.439528in}{3.127393in}}%
\pgfusepath{stroke}%
\end{pgfscope}%
\begin{pgfscope}%
\pgfpathrectangle{\pgfqpoint{0.704982in}{0.547888in}}{\pgfqpoint{4.960000in}{3.696000in}}%
\pgfusepath{clip}%
\pgfsetbuttcap%
\pgfsetroundjoin%
\pgfsetlinewidth{1.505625pt}%
\definecolor{currentstroke}{rgb}{0.172549,0.627451,0.172549}%
\pgfsetstrokecolor{currentstroke}%
\pgfsetdash{{9.600000pt}{2.400000pt}{1.500000pt}{2.400000pt}}{0.000000pt}%
\pgfpathmoveto{\pgfqpoint{0.930437in}{0.715888in}}%
\pgfpathlineto{\pgfqpoint{1.519564in}{0.717162in}}%
\pgfpathlineto{\pgfqpoint{2.018057in}{0.719552in}}%
\pgfpathlineto{\pgfqpoint{2.652501in}{0.724205in}}%
\pgfpathlineto{\pgfqpoint{3.218970in}{0.729951in}}%
\pgfpathlineto{\pgfqpoint{3.830756in}{0.737807in}}%
\pgfpathlineto{\pgfqpoint{4.397225in}{0.746589in}}%
\pgfpathlineto{\pgfqpoint{4.805083in}{0.754811in}}%
\pgfpathlineto{\pgfqpoint{5.122305in}{0.762728in}}%
\pgfpathlineto{\pgfqpoint{5.416869in}{0.771438in}}%
\pgfpathlineto{\pgfqpoint{5.439528in}{0.772153in}}%
\pgfpathlineto{\pgfqpoint{5.439528in}{0.772153in}}%
\pgfusepath{stroke}%
\end{pgfscope}%
\begin{pgfscope}%
\pgfsetrectcap%
\pgfsetmiterjoin%
\pgfsetlinewidth{0.803000pt}%
\definecolor{currentstroke}{rgb}{0.000000,0.000000,0.000000}%
\pgfsetstrokecolor{currentstroke}%
\pgfsetdash{}{0pt}%
\pgfpathmoveto{\pgfqpoint{0.704982in}{0.547888in}}%
\pgfpathlineto{\pgfqpoint{0.704982in}{4.243888in}}%
\pgfusepath{stroke}%
\end{pgfscope}%
\begin{pgfscope}%
\pgfsetrectcap%
\pgfsetmiterjoin%
\pgfsetlinewidth{0.803000pt}%
\definecolor{currentstroke}{rgb}{0.000000,0.000000,0.000000}%
\pgfsetstrokecolor{currentstroke}%
\pgfsetdash{}{0pt}%
\pgfpathmoveto{\pgfqpoint{5.664982in}{0.547888in}}%
\pgfpathlineto{\pgfqpoint{5.664982in}{4.243888in}}%
\pgfusepath{stroke}%
\end{pgfscope}%
\begin{pgfscope}%
\pgfsetrectcap%
\pgfsetmiterjoin%
\pgfsetlinewidth{0.803000pt}%
\definecolor{currentstroke}{rgb}{0.000000,0.000000,0.000000}%
\pgfsetstrokecolor{currentstroke}%
\pgfsetdash{}{0pt}%
\pgfpathmoveto{\pgfqpoint{0.704982in}{0.547888in}}%
\pgfpathlineto{\pgfqpoint{5.664982in}{0.547888in}}%
\pgfusepath{stroke}%
\end{pgfscope}%
\begin{pgfscope}%
\pgfsetrectcap%
\pgfsetmiterjoin%
\pgfsetlinewidth{0.803000pt}%
\definecolor{currentstroke}{rgb}{0.000000,0.000000,0.000000}%
\pgfsetstrokecolor{currentstroke}%
\pgfsetdash{}{0pt}%
\pgfpathmoveto{\pgfqpoint{0.704982in}{4.243888in}}%
\pgfpathlineto{\pgfqpoint{5.664982in}{4.243888in}}%
\pgfusepath{stroke}%
\end{pgfscope}%
\begin{pgfscope}%
\pgfsetbuttcap%
\pgfsetmiterjoin%
\definecolor{currentfill}{rgb}{1.000000,1.000000,1.000000}%
\pgfsetfillcolor{currentfill}%
\pgfsetfillopacity{0.800000}%
\pgfsetlinewidth{1.003750pt}%
\definecolor{currentstroke}{rgb}{0.800000,0.800000,0.800000}%
\pgfsetstrokecolor{currentstroke}%
\pgfsetstrokeopacity{0.800000}%
\pgfsetdash{}{0pt}%
\pgfpathmoveto{\pgfqpoint{0.860538in}{2.824112in}}%
\pgfpathlineto{\pgfqpoint{3.311871in}{2.824112in}}%
\pgfpathquadraticcurveto{\pgfqpoint{3.356315in}{2.824112in}}{\pgfqpoint{3.356315in}{2.868556in}}%
\pgfpathlineto{\pgfqpoint{3.356315in}{4.088333in}}%
\pgfpathquadraticcurveto{\pgfqpoint{3.356315in}{4.132777in}}{\pgfqpoint{3.311871in}{4.132777in}}%
\pgfpathlineto{\pgfqpoint{0.860538in}{4.132777in}}%
\pgfpathquadraticcurveto{\pgfqpoint{0.816093in}{4.132777in}}{\pgfqpoint{0.816093in}{4.088333in}}%
\pgfpathlineto{\pgfqpoint{0.816093in}{2.868556in}}%
\pgfpathquadraticcurveto{\pgfqpoint{0.816093in}{2.824112in}}{\pgfqpoint{0.860538in}{2.824112in}}%
\pgfpathclose%
\pgfusepath{stroke,fill}%
\end{pgfscope}%
\begin{pgfscope}%
\pgfsetrectcap%
\pgfsetroundjoin%
\pgfsetlinewidth{1.505625pt}%
\definecolor{currentstroke}{rgb}{0.498039,0.498039,0.498039}%
\pgfsetstrokecolor{currentstroke}%
\pgfsetdash{}{0pt}%
\pgfpathmoveto{\pgfqpoint{0.904982in}{3.966111in}}%
\pgfpathlineto{\pgfqpoint{1.349427in}{3.966111in}}%
\pgfusepath{stroke}%
\end{pgfscope}%
\begin{pgfscope}%
\definecolor{textcolor}{rgb}{0.000000,0.000000,0.000000}%
\pgfsetstrokecolor{textcolor}%
\pgfsetfillcolor{textcolor}%
\pgftext[x=1.527204in,y=3.888333in,left,base]{\color{textcolor}\fontsize{16.000000}{19.200000}\selectfont original encoding}%
\end{pgfscope}%
\begin{pgfscope}%
\pgfsetbuttcap%
\pgfsetroundjoin%
\pgfsetlinewidth{1.505625pt}%
\definecolor{currentstroke}{rgb}{1.000000,0.498039,0.054902}%
\pgfsetstrokecolor{currentstroke}%
\pgfsetdash{{5.550000pt}{2.400000pt}}{0.000000pt}%
\pgfpathmoveto{\pgfqpoint{0.904982in}{3.654111in}}%
\pgfpathlineto{\pgfqpoint{1.349427in}{3.654111in}}%
\pgfusepath{stroke}%
\end{pgfscope}%
\begin{pgfscope}%
\definecolor{textcolor}{rgb}{0.000000,0.000000,0.000000}%
\pgfsetstrokecolor{textcolor}%
\pgfsetfillcolor{textcolor}%
\pgftext[x=1.527204in,y=3.576333in,left,base]{\color{textcolor}\fontsize{16.000000}{19.200000}\selectfont learned at first UIP}%
\end{pgfscope}%
\begin{pgfscope}%
\pgfsetbuttcap%
\pgfsetroundjoin%
\pgfsetlinewidth{1.505625pt}%
\definecolor{currentstroke}{rgb}{0.839216,0.152941,0.156863}%
\pgfsetstrokecolor{currentstroke}%
\pgfsetdash{{1.500000pt}{2.475000pt}}{0.000000pt}%
\pgfpathmoveto{\pgfqpoint{0.904982in}{3.344111in}}%
\pgfpathlineto{\pgfqpoint{1.349427in}{3.344111in}}%
\pgfusepath{stroke}%
\end{pgfscope}%
\begin{pgfscope}%
\definecolor{textcolor}{rgb}{0.000000,0.000000,0.000000}%
\pgfsetstrokecolor{textcolor}%
\pgfsetfillcolor{textcolor}%
\pgftext[x=1.527204in,y=3.266333in,left,base]{\color{textcolor}\fontsize{16.000000}{19.200000}\selectfont learned at last UIP}%
\end{pgfscope}%
\begin{pgfscope}%
\pgfsetbuttcap%
\pgfsetroundjoin%
\pgfsetlinewidth{1.505625pt}%
\definecolor{currentstroke}{rgb}{0.172549,0.627451,0.172549}%
\pgfsetstrokecolor{currentstroke}%
\pgfsetdash{{9.600000pt}{2.400000pt}{1.500000pt}{2.400000pt}}{0.000000pt}%
\pgfpathmoveto{\pgfqpoint{0.904982in}{3.034111in}}%
\pgfpathlineto{\pgfqpoint{1.349427in}{3.034111in}}%
\pgfusepath{stroke}%
\end{pgfscope}%
\begin{pgfscope}%
\definecolor{textcolor}{rgb}{0.000000,0.000000,0.000000}%
\pgfsetstrokecolor{textcolor}%
\pgfsetfillcolor{textcolor}%
\pgftext[x=1.527204in,y=2.956333in,left,base]{\color{textcolor}\fontsize{16.000000}{19.200000}\selectfont reduced constraint}%
\end{pgfscope}%
\end{pgfpicture}%
\makeatother%
\endgroup%